\documentclass[journal,10pt,twocolumn,oneside]{IEEEtran}

\usepackage{booktabs}
\usepackage{wrapfig}
\usepackage{lipsum}
\usepackage{verbatim}
\usepackage{algorithm}
\usepackage{algorithmicx}
\usepackage{algpseudocode}
\usepackage{graphicx}
\usepackage{subfigure}
\usepackage{cite}
\usepackage[cmex10]{amsmath}
\usepackage{amsthm}
\usepackage{epstopdf}
\usepackage{diagbox}
\usepackage{array}
\usepackage{enumerate}

\usepackage{makecell}
\usepackage{arydshln}
\usepackage{multirow}
\usepackage{multicol}

\usepackage{nicefrac}
\usepackage{bm}
\def\0{{\mathbf 0}}
\def\1{{\mathbf 1}}

\def\a{{\mathbf a}}

\def\c{{\mathbf c}}

\def\e{{\mathbf e}}
\def\f{{\mathbf f}}
\def\g{{\mathbf g}}
\def\h{{\mathbf h}}

\def\l{{\mathbf l}}

\def\n{{\mathbf n}}

\def\x{{\mathbf x}}
\def\y{{\mathbf y}}

\def\H{{\mathbf H}}

\def\L{{\mathbf L}}
\def\M{{\mathbf M}}

\def\W{{\mathbf W}}

\def\rPr{{\text{Pr}}}

\def\ie{{\textit{i.e.}}}
\def\eg{{\textit{e.g.}}}

\def\cH{{\mathcal H}}

\def\cL{{\mathcal L}}

\def\cN{{\mathcal N}}
\def\cO{{\mathcal O}}
\def\cR{{\mathcal R}}

\def\cY{{\mathcal Y}}

\def\bsPsi{{\boldsymbol \Psi}}

\usepackage{amsfonts}

\usepackage{color}

\usepackage{mdwmath}
\usepackage{url}
\usepackage{amssymb}
\usepackage{soul}

\newtheorem{definition}{\textbf{Definition}}

\begin{document}

%
\title{Graph-Based Depth Denoising \& Dequantization for Point Cloud Enhancement}
\author{
\IEEEauthorblockN{Xue Zhang, \emph{Member, IEEE}, Gene Cheung, \emph{Fellow, IEEE}, Jiahao Pang, \emph{Member, IEEE}, Yash Sanghvi, Abhiram Gnanasambandam, Stanley H. Chan, \emph{Senior Member, IEEE}}
\renewcommand{\baselinestretch}{1.0}
\thanks{The work of G. Cheung was supported in part by the Natural Sciences and Engineering Research
Council of Canada (NSERC) RGPIN-2019-06271, RGPAS-2019-00110, and InterDigital. The work of S. H. Chan was supported in part by the US National Science Foundation CCF-1763896, CCSS-2030570. \emph{(Corresponding author: Gene Cheung.)}}
\thanks{X. Zhang is with the College of Computer Science and Engineering, Shandong University of Science and Technology, Qingdao, 266590, China. (e-mail: xuezhang@sdust.edu.cn)} 
\thanks{G. Cheung is with the department of EECS, York University, 4700 Keele Street, Toronto, M3J 1P3, Canada. (e-mail: genec@yorku.ca)}
\thanks{J. Pang is with InterDigital Communications Inc., New York, NY 10120, USA. (e-mail: jiahao.pang@interdigital.com)}
\thanks{Y. Sanghvi, A. Gnanasambandam and S. H. Chan are with the School of ECE, Purdue University, West Lafayette, IN 47907, USA. (e-mail:\{ysanghvi, agnanasa, stanchan\}@purdue.edu)}
}
\maketitle
%
\begin{abstract}
A 3D point cloud is typically constructed from depth measurements acquired by sensors at one or more viewpoints. 
The measurements suffer from both quantization and noise corruption. 
To improve quality, previous works denoise a point cloud \textit{a posteriori} after projecting the imperfect depth data onto 3D space.
Instead, we enhance depth measurements directly on the sensed images \textit{a priori}, before synthesizing a 3D point cloud.
By enhancing near the physical sensing process, we tailor our optimization to our depth formation model before subsequent processing steps that obscure measurement errors. 

Specifically, we model depth formation as a combined process of signal-dependent noise addition and non-uniform log-based quantization.
The designed model is validated (with parameters fitted) using collected empirical data from a representative depth sensor.
To enhance each pixel row in a depth image, we first encode intra-view similarities between available row pixels as edge weights via feature graph learning. 
We next establish inter-view similarities with another rectified depth image via viewpoint mapping and sparse linear interpolation.
This leads to a maximum a posteriori (MAP) graph filtering objective that is convex and differentiable.
We minimize the objective efficiently using accelerated gradient descent (AGD), where the optimal step size is approximated via Gershgorin circle theorem (GCT). 
Experiments show that our method significantly outperformed recent point cloud denoising schemes and state-of-the-art image denoising schemes in two established point cloud quality metrics.
\end{abstract}
%


%
\maketitle

\section{Introduction}
\begin{figure}[h]
\centering
\includegraphics[width=\linewidth]{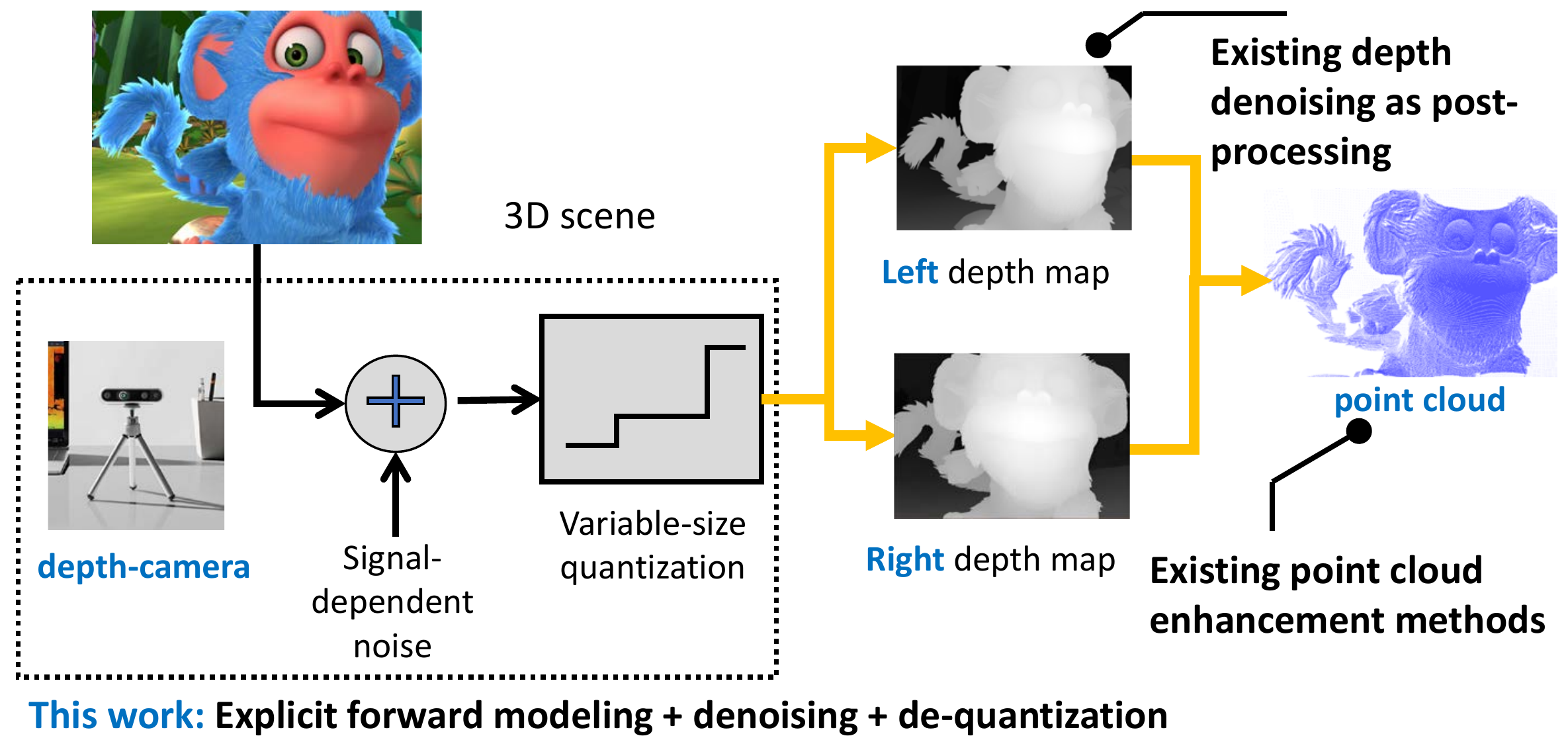}
\vspace{-0.25in}
\caption{Depth measurements from a consumer sensor suffer from signal-dependent noise and variable-size quantization. 
We enhance multi-view depth measurements as close to the physical sensing process as possible, resulting in higher-quality subsequent 3D point cloud construction.}
\label{fig:main_contribution}
\vspace{-0.1in}
\end{figure}

\textit{Point Cloud} (PC) is a collection of discrete geometric samples of the surface of a physical object in 3D space, useful for a range of imaging applications such as immersive communication and virtual / augmented reality (AR/VR) \cite{mekuria2016design, gunkel2018virtual, yuan2020sampling,zhang2021graph}. 
With the ubiquity of inexpensive active sensors like Microsoft Kinect and Intel RealSense, one common method to generate a PC is to deploy one or more sensors at multiple viewpoints to capture depth measurements (in the form of images) of an object, then project these measurements to 3D space to synthesize a PC \cite{hartley2003multiple, huang2018apolloscape}.
However, limitations in the depth acquisition process mean that the acquired depth measurements suffer from both imprecision (due to quantization) and additive noise.
This results in a noisy synthesized PC, and previous works focus on denoising PCs using a variety of methods: low-rank prior, low-dimensional manifold model (LDMM), surface smoothness priors expressed as graph total variation (GTV), graph Laplacian regularizer (GLR) and feature graph Laplacian regularizer (FGLR), Moving Robust Principal Components Analysis (MRPCA), data-driven learning, etc \cite{zeng20193d, dinesh2020point, hu2020feature, mattei2017point, rakotosaona2020pointcleannet,luo2020differentiable}.

However, all the aforementioned denoising methods enhance a PC \textit{a posteriori}, \ie, after a PC is synthesized from corrupted depth measurements. 
Recent works in image denoising \cite{punnappurath2019learning, nguyen2016raw} have shown that by denoising raw sensed RGB measurements directly on the Bayer-patterned grid \textit{before} demosaicking, contrast boosting and other steps typical in an imaging pipeline \cite{farsiu2005multiframe, sahu2017contrast} that obscure acquisition noise, one can dramatically improve performance compared to denoising the constructed image \textit{after} the pipeline (up to 15dB in PSNR).

Inspired by these works, we propose to \textit{enhance} depth measurements in sensed images \textit{a priori}, before projecting to the 3D space to synthesize a PC.
By ``enhancement'', here we mean performing joint denoising and dequantization based on our proposed depth formation model.
In our case, by enhancing near the physical sensing process before steps in a PC synthesis pipeline including projection, registration, stitching, and filtering \cite{alexiadis2012real}, we can tailor our optimization by parameterizing our proposed depth formation model to a specific sensor.
As illustrated in Fig.\;\ref{fig:main_contribution}, this means fitting model parameters for \textit{signal-dependent noise} and \textit{variable-size quantization} that are unique for individual sensors. 
This precise modeling of sensed depth measurements is in stark contrast to the grossly inaccurate i.i.d. Gaussian noise\footnote{For example, a corrupted depth pixel is erred \textit{perpendicular} to the image plane, which is very different from a projected 3D point corrupted by i.i.d. Gaussian noise along its three coordinates as assumed in existing works.} typically assumed in the PC denoising literature \cite{zeng20193d, dinesh2020point, hu2020feature, mattei2017point, rakotosaona2020pointcleannet,luo2020differentiable}.  

Mathematically, we first model depth pixel acquisition as a combined process of signal-dependent noise addition and non-uniform log-based quantization.
We verify the validity of this model and fit model parameters through an empirical study using the \textit{Intel RealSense\textsuperscript{TM} D435} camera---a representative depth sensor that employs stereo correspondence technology on a pair of IR images to improve depth quality. 
We parameterize the signal-dependent noise variance as a function of depth from empirical data using the golden-section (GS) search \cite{press2007numerical} in a \textit{maximum likelihood} (ML) formulation.

\begin{figure}
\begin{center}
\begin{tabular}{cc}
\includegraphics[width=1.5in]{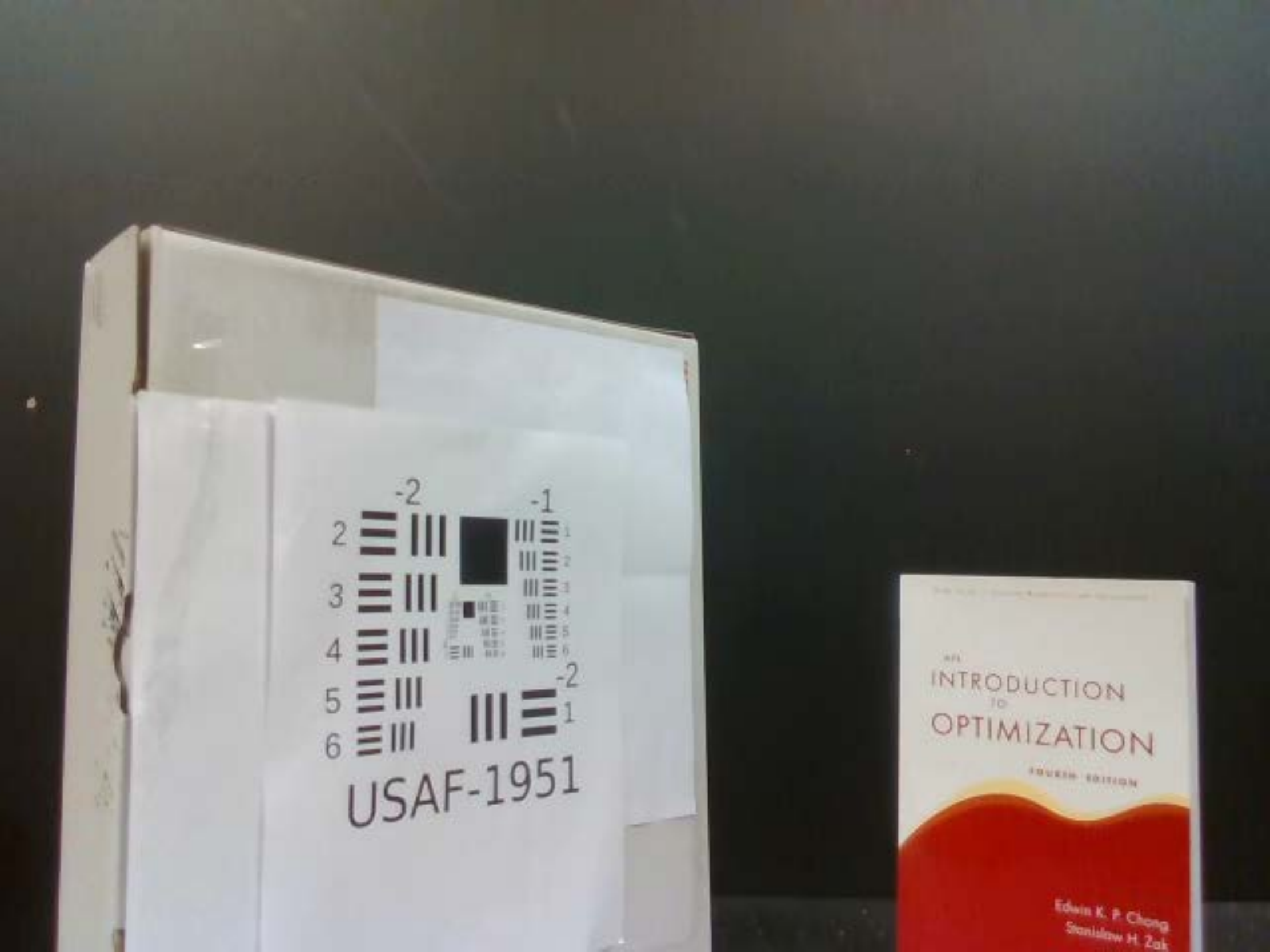}&\includegraphics[width=1.48in]{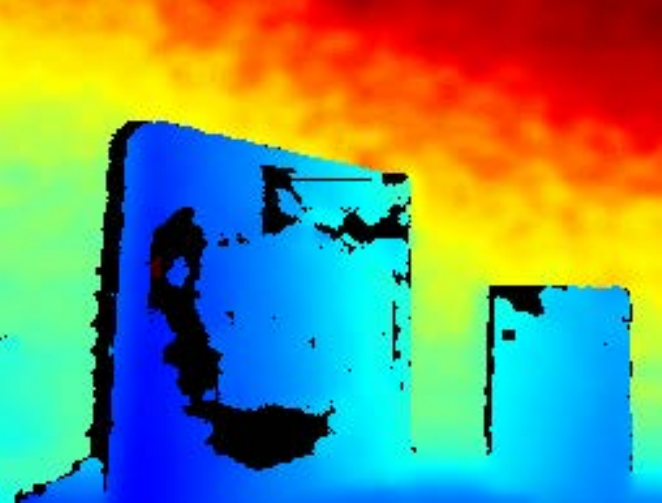}\\
\parbox{.45\linewidth}{\centering\small (a)} & \parbox{.45\linewidth}{\centering\small (b)}\\
\end{tabular}
\end{center}
\vspace{-0.15in}
\caption{An example of collected images using the \textit{Intel RealSense\textsuperscript{TM} D435} : (a) the color image, (b) the corresponding depth map with missing pixels.}
\label{fig:depth}
\vspace{-0.1in}
\end{figure}

To enhance a row of available pixels in a left depth image---often interleaved with missing pixels\footnote{A depth pixel from a structured light sensor is missing if an occlusion or optical interference occurs \cite{kadambi20143d}. A depth pixel from a time-of-flight sensor is missing if the emitted laser arrives at an incident angle such that it is not reflected and returned to the sensor \cite{kadambi20143d}.} as shown in Fig.\;\ref{fig:depth}---we first encode \textit{intra-view} similarities between neighboring available pixels as edge weights in a graph via \textit{feature graph learning} \cite{hu2020feature}. 
Specifically, estimated 3D features per depth pixel---\ie, 3D coordinate and surface normal---are used to compute \textit{feature distance} $d_{ij}$ between pixel pair $(i,j)$, so that edge weight $w_{ij} = \exp (-d_{ij})$ can be computed using rich 3D structure information to construct a similarity graph.

We next establish \textit{inter-view} similarities with a rectified right depth image via viewpoint mapping \cite{jin2016region} and sparse linear interpolation.
Appropriate approximations via Taylor series expansion lead to a \textit{maximum a posteriori} (MAP) graph filtering objective that is convex and differentiable.
We minimize the objective efficiently using \textit{accelerated gradient descent} (AGD) \cite{nesterov1983method, bubeck2014convex}, where the optimal step size is efficiently approximated via \textit{Gershgorin circle theorem} (GCT) \cite{horn2012matrix}.
Experimental results show that by enhancing depth measurements in the image domain prior to PC synthesis, our method significantly outperformed several recent PC denoising algorithms \cite{zeng20193d, mattei2017point,guennebaud2007algebraic, oztireli2009feature, rakotosaona2020pointcleannet,luo2020differentiable} and representative image denoising schemes\footnote{Our paper is a non-trivial journal extension of our earlier work \cite{zhang20203d} that assumed signal-independent noise and uniform quantization. We compare against \cite{zhang20203d} experimentally in Section\;\ref{sec:results}.} \cite{tomasi98, dabov2007image,zhang20203d} in two commonly used PC quality metrics \cite{girardeau2005change,tian2017geometric} for three different types of depth datasets.

We summarize our technical contributions as follows:
\begin{enumerate}[(i)]
\item To improve PC construction quality near the physical sensing process, we enhance depth measurements in the image domain, where we can parameterize our depth formation model for a specific sensor. 
\item We design the depth formation model by combining \textit{both} signal-dependent noise addition \textit{and} non-uniform log-based quantization process, validated with collected empirical data from a representative depth sensor.
\item We encode intra-view similarities for \textit{available} pixel pairs in a depth image row as edge weights via feature graph learning, and inter-view similarities via viewpoint mapping and sparse linear interpolation.
\item We minimize the resulting convex and differentiable MAP graph filtering objective via AGD, where the important step size is approximated speedily via GCT.
\end{enumerate}
We stress that, while the general approach to denoise / restore observed data as close to the physical layer as possible is not new, our specialization for 3D point cloud is novel.
In particular, we are the first in the 3D point cloud literature to propose a joint non-uniform quantization / signal-dependent additive noise model, and a corresponding restoration algorithm designed for the proposed model.

The outline of the paper is as follows. 
We first overview related works in Section\;\ref{sec:related}.
We then describe our depth capturing system and the depth formulation model in Section\;\ref{sec:system}. 
We also present our empirical study based on collected depth data.
The formulation of our optimization problem and the corresponding algorithm to solve it are discussed in Section\;\ref{sec:formulate}.
We discuss the graph learning procedure to capture inter-pixel similarities in Section\;\ref{sec:learn}.
Experimental results and conclusion are presented in Section\;\ref{sec:results} and \ref{sec:conclude}, respectively. 

\graphicspath{{fig/}}

\section{Related Works}
\label{sec:related}
We review two categories of related works: depth image denoising / enhancement and 3D point cloud denoising.

\subsection{Depth Image Denoising / Enhancement}

\subsubsection{Depth image denoising}
%
Unlike natural image denoising \cite{dabov2007image, Nagahama2021_unrolling, Chan2017_pnp}, existing literature in depth image denoising is dominated by model-based schemes \cite{zhang20203d, hu2013depth, lu2014depth, xie2014single, ham2015robust, pang17, gu2017learning, yang2014color, Liu2015_sparse}.  
Specifically, it is common to perform depth denoising using an assumed signal prior or filter \cite{hu2013depth, lu2014depth, xie2014single, ham2015robust, pang17, gu2017learning}. For example, \cite{hu2013depth} used a sparsity prior for each pixel patch in the graph Fourier domain, while \cite{xie2014single} employed a regularized shock filter.
Both methods exploited the known piecewise smooth (PWS) characteristic of depth images \cite{hu16spl} for better performance.
When the color image associated to the depth map is available, denoising performance can be further boosted.
For example, \cite{yang2014color} built neighborhood graphs from color images for depth denoising.

However, these works modeled the depth degradation process with \textit{signal-independent} noise---\eg, i.i.d. additive Gaussian noise \cite{hu2013depth,pang17,gu2017learning}---for simplicity.
As discussed in \cite{nguyen2012modeling,mallick2014characterizations} and also verified in our work, depth measurements are noisier for larger depth values; \ie, the noise variance is \textit{signal-dependent}.
Moreover, existing depth image denoising algorithms offer no explicit procedures to handle missing depth values that are common in real-world sensed depth images, as shown in Fig.\;\ref{fig:depth}.
In contrast, our proposed depth formation model accounts for noise variance's signal dependency, which we parameterize accurately using collected empirical data of an actual depth sensor in Section\;\ref{subsubsec:noise}. 
Further, our graph-based optimization enhances available pixels around missing ones by capturing inter-pixel similarity via feature graph learning \cite{hu2020feature}.
We compare against several representative image-denoising schemes \cite{tomasi98, dabov2007image,zhang20203d} in Section\;\ref{sec:results}.

\subsubsection{Depth image dequantization}
Depth images are quantized by depth sensors for storage or transmission purposes~\cite{ideses2007depth}.
To enhance depth precision, previous works adopted different strategies.
Given several depth maps from different views, \cite{wan15} used the \textit{Projection On Convex Set} (POCS) procedure to enhance precision of two depth images simultaneously, while \cite{hu16spl,wang2014graph,zhang20203d} employed a graph-signal smoothness prior to enhance a single depth image.

Previous research all assumed that depth images are quantized uniformly.
However, in practical sensors, larger quantization bin are used as depth values increase, as confirmed by our empirical data in Section\;\ref{subsec:exp_data}.
Thus, we propose a non-uniform log-based quantization process, so that larger depth values have coarser quantization.

\vspace{-0.1in}
\subsection{3D Point Cloud Denoising}
PC denoising~\cite{javaheri2017subjective} is necessary prior to 2D viewpoint image rendering or another down-stream task.
Related model-based methods can be roughly categorized into four types: moving least square (MLS)-based methods \cite{alexa2003computing, fleishman2005robust, oztireli2009feature}, locally optimal projection (LOP)-based methods \cite{lipman2007parameterization, huang2009consolidation}, sparsity-based methods \cite{sun2015denoising,avron2010l1} and nonlocal-based methods \cite{zeng20193d,rosman2013patch,wang2008similarity}.
Among these works, with the exception of \cite{avron2010l1} that related the noise variance to surface normal, they all assumed independent additive noise.
For instance, \cite{sun2015denoising,zeng20193d,wang2008similarity} adopted the additive i.i.d. Gaussian noise model to simplify analysis.
As discussed earlier, we model noise on each measurement in a depth image as signal-dependent.

Beyond the signal-dependent nature of acquisition noise, another drawback of existing methods is the assumption that the 3D coordinates of each PC point are omnidirectionally corrupted by additive noise.
This is a gross oversimplification: depth measurements are first corrupted by additive noise and quantization perpendicular to the image plane, before 3D projection to synthesize a PC---the resulting distortion in the projected 3D space is far from omnidirectional.
In contrast, we enhance depth measurements on an image before 3D projection. 

%
Recent developments of \textit{deep neural networks} (DNN) have revolutionized different areas in computer vision and image processing, including PC denoising~\cite{rakotosaona2020pointcleannet,luo2020differentiable}.
Beyond the inaccurate assumption of each PC corrupted by i.i.d. Gaussian noise, these learning-based approaches are purely data-driven, and a large amount of labeled data is required for training~\cite{lecun2015deep}.
However, it is difficult to collect noiseless ground-truth data for supervised learning, since the PCs acquired from depth sensors are always noise-corrupted.
One option to circumvent this issue is to synthesize training data via computer graphics \cite{mayer2016large}.
However, it still suffers from inaccurate modeling of noise and missing pixels of actual depth sensors.
Different from the learning-based methods, by optimizing only a small set of parameters, our model-based method can easily adapt to new depth sensors with different specifications.
Moreover, compared to complex deep models acting like black boxes, the derived filters of our methods are interpretable \cite{zeng2019deep}.

\section{Forward Model and Validation}
\label{sec:system}
\begin{figure}
\begin{center}
\includegraphics[width=3in]{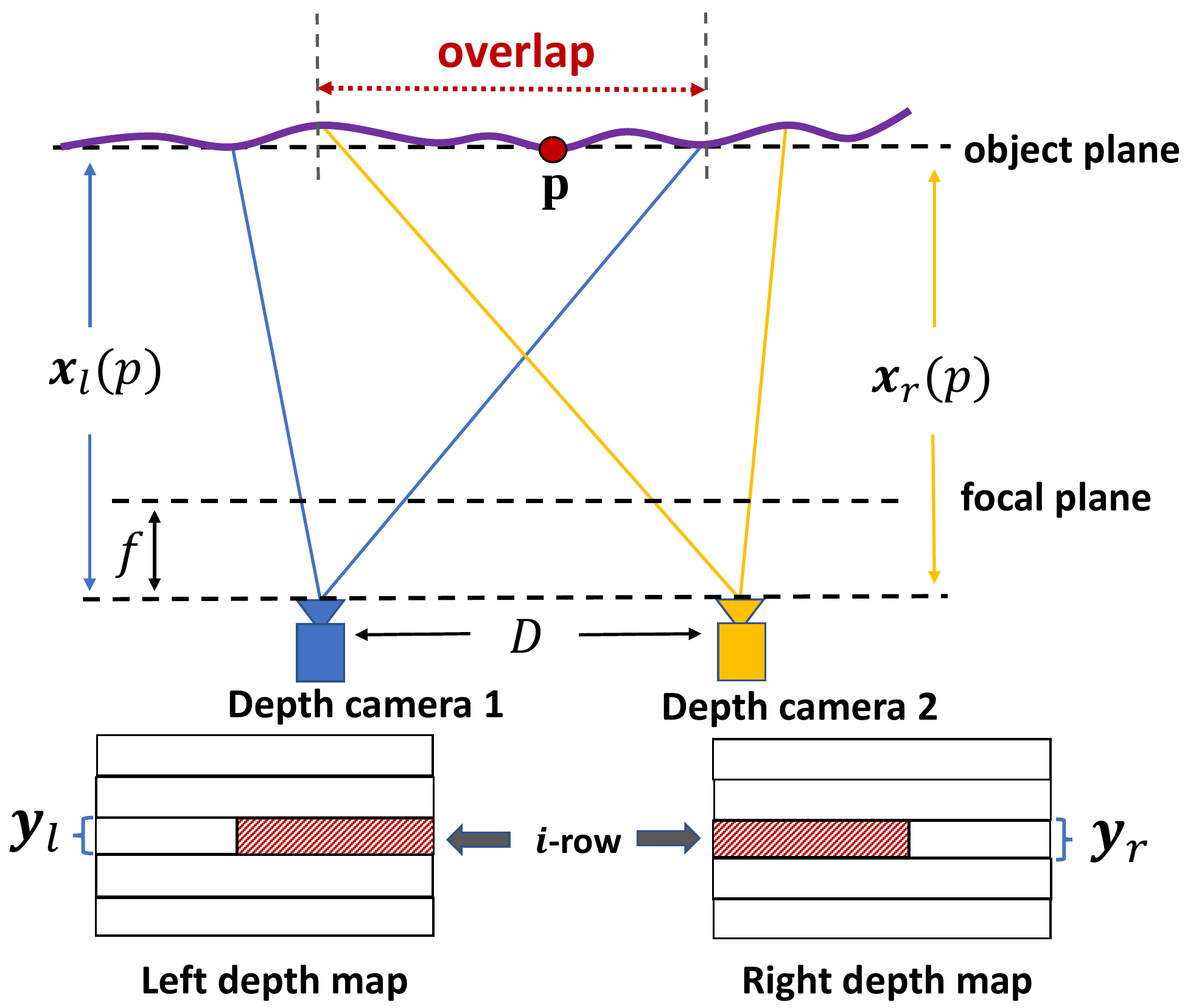}
\vspace{-0.1in}
\caption{\small An example of the depth camera system. 
Two depth cameras with focal length $f$ are placed distance $D$ apart capturing the same 3D scene. Assuming that the corresponding depth images are rectified, a left pixel row $\x_l$ maps to a right pixel row $\x_r$ via a view-to-view mapping described in Section\;\ref{subsec:V2VMapping},
$\y_l$ and $\y_r$ are corrupted observations.} 
\label{fig:model}
\end{center}
\vspace{-0.3cm}
\end{figure}

\subsection{System Overview}

We consider a depth-sensing system consisting of two depth sensors from different viewpoints, which are separated by a distance $D$\footnote{The extension to the more general multiple-camera (more than two) case is straightforward. We omitted this discussion for brevity.}. For systems with two sensors (or one sensor placed at two locations at different time instants), there exists an overlapping \textit{field of view} (FoV), where the same object surface is observed twice from two different viewpoints as illustrated in Fig.\;\ref{fig:model}. 
The output of each depth sensor is a depth map of resolution $H \times N$. Each pixel is a noise-corrupted observation of the physical distance between the camera and the object, and is non-uniformly quantized to a finite $B$-bit representation. Without access to the underlying hardware directly, we assume that the depth map is the ``rawest" signal we can acquire from the sensor.
To simplify later discussion, we assume that two viewpoint depth maps are \textit{rectified}; \ie, pixel row $h$ in the left view is capturing the same slice of the object as pixel row $h$ in the right view. Rectification is a well-studied problem, and a known procedure \cite{loop1999computing} can be executed prior to our depth enhancement algorithm. 

We now present the image formation model of a depth pixel. Our model consists of two parts: non-uniform quantization and signal-dependent noise.

\subsection{Non-Uniform Quantization}
\label{subsec:formation}

\begin{figure}
\begin{center}
\begin{tabular}{cc}
\includegraphics[width=1.65in]{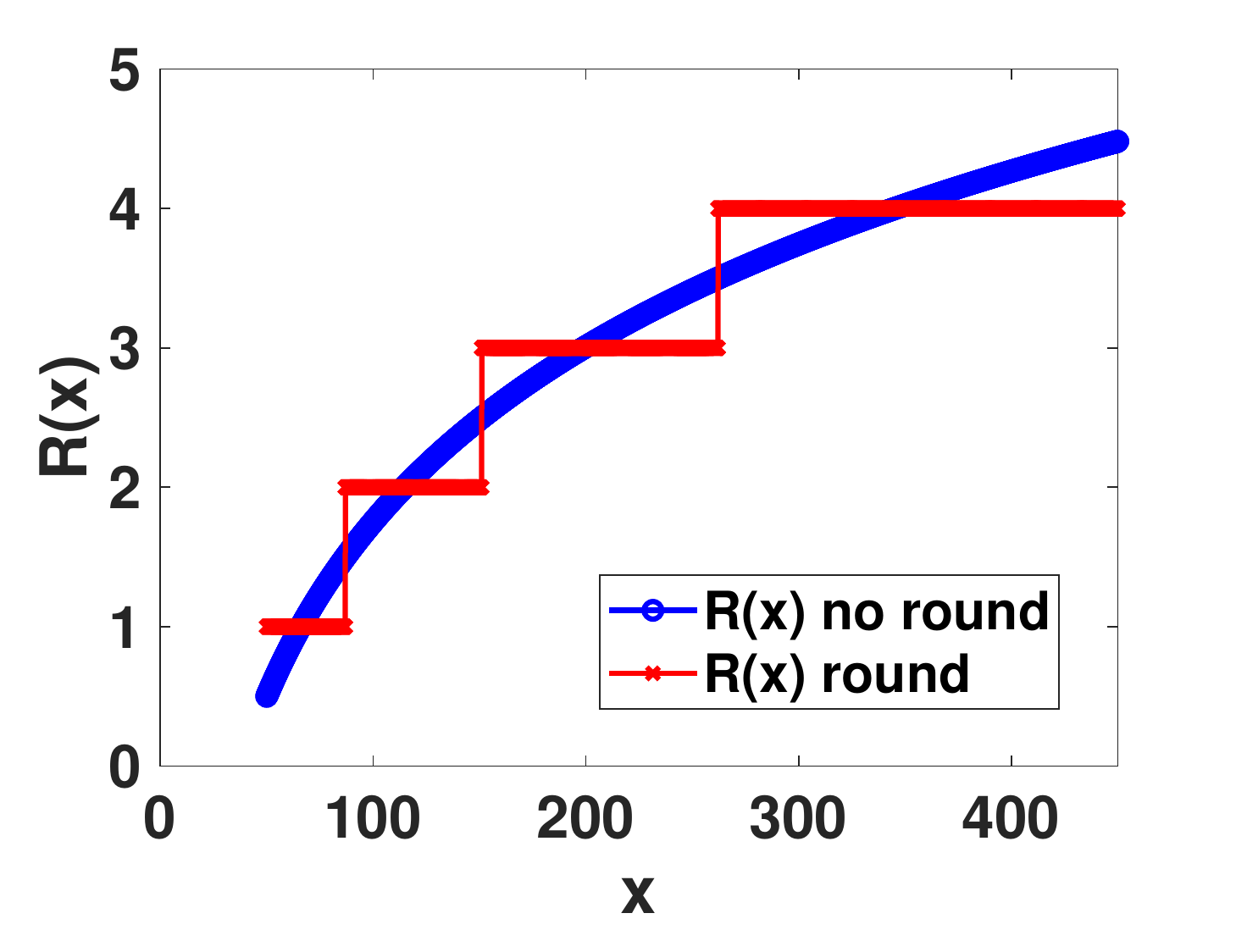} & \includegraphics[width=1.65in]{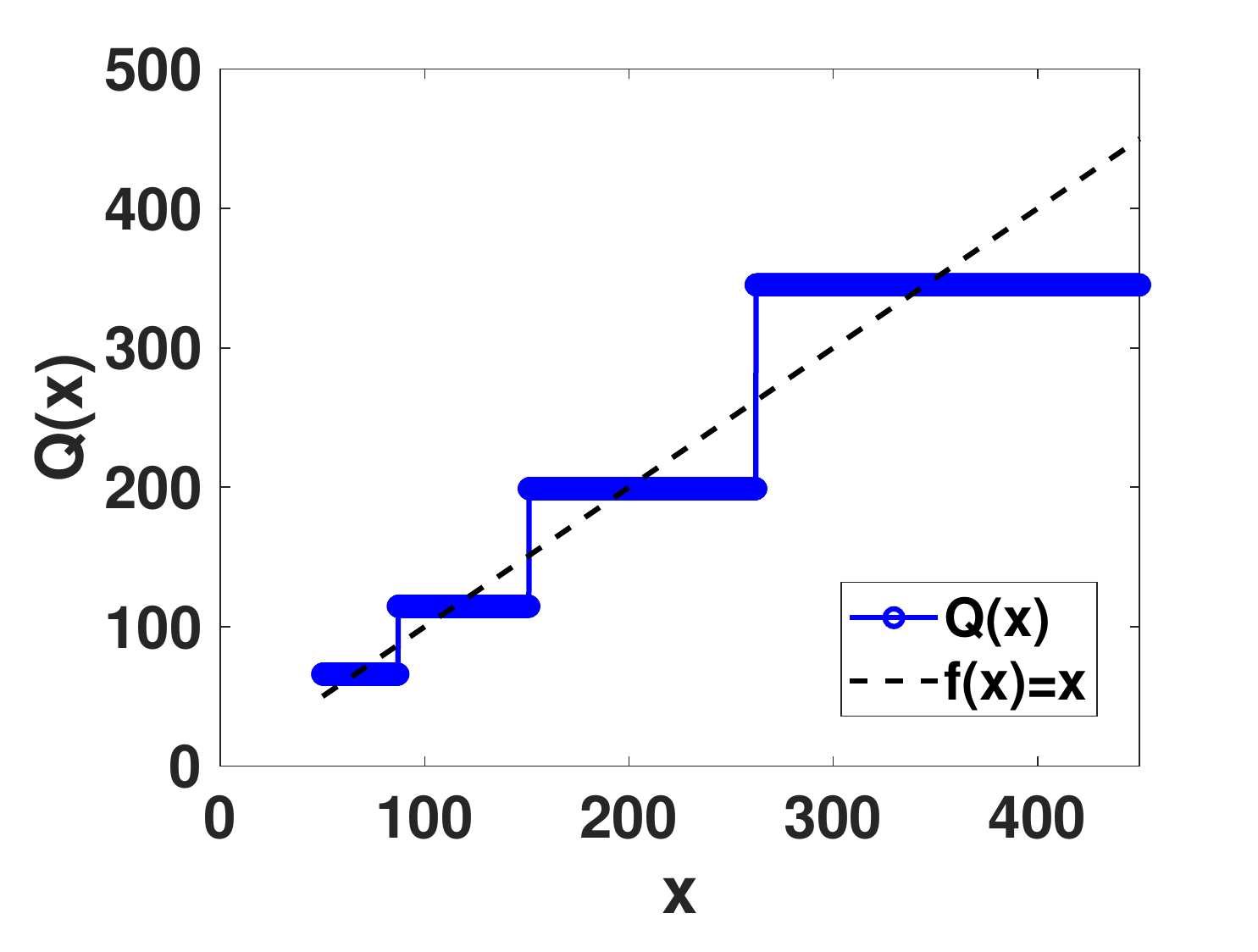}\\
\parbox{.45\linewidth}{\centering\small (a)} & \parbox{.45\linewidth}{\centering\small (b)}
\end{tabular}
\end{center}
\vspace{-0.15in}
\caption{Illustration of (a) quantization mapping $R(x)$ in \eqref{eq:quanPara} and (b) non-uniform quantization function $Q(x)$ in \eqref{eq:nonUniform}, with $\theta = 2$, $\rho = 1$, $x_{\min}=50$, $x_{\max}=450$.}
\label{fig:quan}
\end{figure}


Typically, a depth camera quantizes an input depth value into a finite $B$-bit representation. 
For example, Microsoft Kinect version 1.0 has a 11-bit per-pixel representation, while version 2.0 has a 13-bit representation. Such a quantization is common for depth sensors in the consumer market, where limited precision is allocated more for objects closer to the camera than ones further away. 

To model non-uniform quantization, we define our non-uniform quantizer with wider quantization bins for larger input depth values.

\begin{definition}[Quantization Mapping]
Let $x \in [x_{\min}, x_{\max})$ be the input value. 
We define the quantization mapping function $R(x)$ using the natural log and rounding functions as
\begin{equation}
R(x) \triangleq \text{round} \bigg[ \phi \ln{(\theta x + \rho)} - R_{0} + 0.5 \bigg],
\label{eq:quanPara}
\end{equation}
where $\phi$, $\theta$ and $\rho$ are parameters, and $R_{0} \triangleq \phi \ln (\theta x_{\min} + \rho)$. The rounding operation in \eqref{eq:quanPara} implies that the quantization mapping (QM) $R(x) \in \mathbb{Z}^+$ is \textit{piecewise constant}.
\end{definition}
The quantization mapping $R(x)$ maps the input depth value $x$ with range $[x_{\min}, x_{\max})$ to one of $2^B$ discrete values\footnote{A depth sensor typically specifies a range of depth sensitivity. For example, the range for MS Kinect 2.0 is $0.5$m to $4.5$m.}: $\{1, \ldots, 2^B\}$. 
To make sense of this definition, we note that $\theta > 0$ and $\theta x_{\min} + \rho \geq  1$. Since $\theta x + \rho \geq 1$, we use only the non-negative part of the log function in \eqref{eq:quanPara}. The rounding operation also means that a smaller slope of function $\ln (\theta x + \rho)$ would lead to a coarser quantization of $x$; \ie, a larger range of $x$ would map to the same rounded integer $R(x)$. 
Thus, concavity of the log function means that larger $x$ has coarser quantization, as desired. 
See Fig.\;\ref{fig:quan}(a) for an illustration. 

We note that the use of log function in \eqref{eq:quanPara} is similar to the $\mu$-law companding algorithm in ITU-T's G.711 standard for PCM digital communication\footnote{https://www.itu.int/rec/T-REC-G.711}, where non-uniform quantization is used for audio encoding.

The parameter $\phi$ is chosen so that $R(x) \in \{1, \ldots, 2^B\}$ can represent $2^B$ quantization bins given available $B$-bit representation per pixel.
Specifically, $\phi$ is computed as
\begin{align}
\phi = \frac{2^B}{\ln (\theta x_{\max} + \rho) - \ln (\theta x_{\min} + \rho)} .
\end{align}
In contrast, $\theta$ and $\rho$ are fitting parameters to fit characteristics of a particular depth sensor (to be discussed in Section\;\ref{subsec:exp_data}). 
Fig.\;\ref{fig:quan}(a) shows $R(x)$ when $B=2$, and there are $2^B = 4$ quantization bins, \ie, $R(x) \in \{1, 2, 3, 4\}$.

Given quantization mapping $R(x)$ in \eqref{eq:quanPara}, we can now define \textit{quantization function} $Q(x)$.
\begin{definition}[Quantization Function]
We define the quantization function $Q(x)$ as
\begin{align}
Q(x) = \frac{1}{\theta}\left( \exp\left\{\frac{R(x) + R_{0} - 0.5}{\phi} \right\} - \rho\right).
\label{eq:nonUniform}
\end{align}
\end{definition}
The quantization function in \eqref{eq:nonUniform} essentially reverses the operations in quantization mapping $R(\cdot)$ in (1) to recover input $x$. Given that $R(x)$ is piecewise constant, $Q(x)$ is also piecewise constant, as shown in Fig.\;\ref{fig:quan}(b). 
We verify that $Q(x)$ is indeed a quantization function---the centers of each quantization step coincide with the identity function $f(x) = x$---while the quantization bin size is increasing with $x$ as intended.

\subsection{Signal-Dependent Noise Model}
\label{subsubsec:noise}

Before quantization $Q(\cdot)$, we assume that each measurement $x$ is first corrupted by \textit{signal-dependent} noise $n \sim \mathcal{N}(0, \sigma^2)$ following a zero-mean Gaussian distribution. 
Noise variance $\sigma^2 \in \mathbb{R}^+$ is dependent on signal $x$. 
Specifically, as done in \cite{nguyen2012modeling, mallick2014characterizations}, we assume that the \textit{standard deviation} (SD) $\sigma$ is a quadratic function of signal $\x$, \ie, 
\begin{align}
\sigma = \alpha(x + \mu)^2 + \kappa, ~~~ \mbox{for}~ x \geq -\mu
\label{eq:sigDependent}
\end{align}
where parameters $\alpha>0$, $\mu<0$ and $\kappa>0$. 
It means that $\sigma$ increases quadratically with signal magnitude.

\subsection{Validation of the Model}
\label{subsec:exp_data}

To provide a concrete example of the quantization and noise model we just presented, we use the \textit{Intel RealSense\textsuperscript{TM} D435} camera to validate this model and fit model parameters through an empirical study.
Our chosen Intel sensor is representative of popular depth sensors in the market and shares similar characteristics.
Although the working principles of various types of 3D sensors, \textit{stereo, structured light}, and \textit{time of flight (ToF)}, are slightly different, existing works \cite{fehn04,mehrotra2011low,tontini2020numerical} commonly assumed non-uniform quantization and \cite{dermanis2000geomatic,mallick2014characterizations,kim2012parametric} assumed signal-dependent noise in their depth sensor models.
We stress, however, that we are the first to combine \textit{both} non-uniform quantization \textit{and} signal-dependent noise into the same forward model. 

\begin{figure}
\centering
\includegraphics[width=0.9\linewidth]{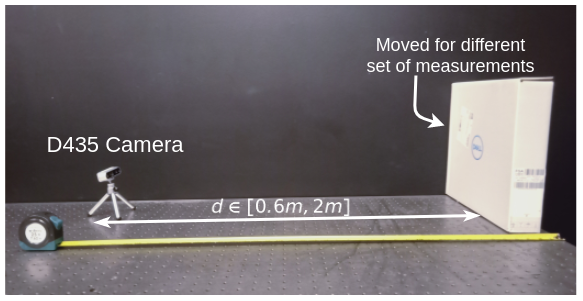}
\vspace{-0.1in}
\caption{Experimental Setup for obtaining set of measurements for different depths. The \textit{Intel RealSense\textsuperscript{TM} D435} camera is placed in front of the cardboard box and the distance between them is varied in steps of 100mm from 615mm to 1525mm.}
\label{fig:setup-expt}
\end{figure}

\begin{figure}[ht]
\centering
\begin{tabular}{c}
\includegraphics[trim={0cm 0.2cm 0cm 0.0cm},clip, width=\linewidth]{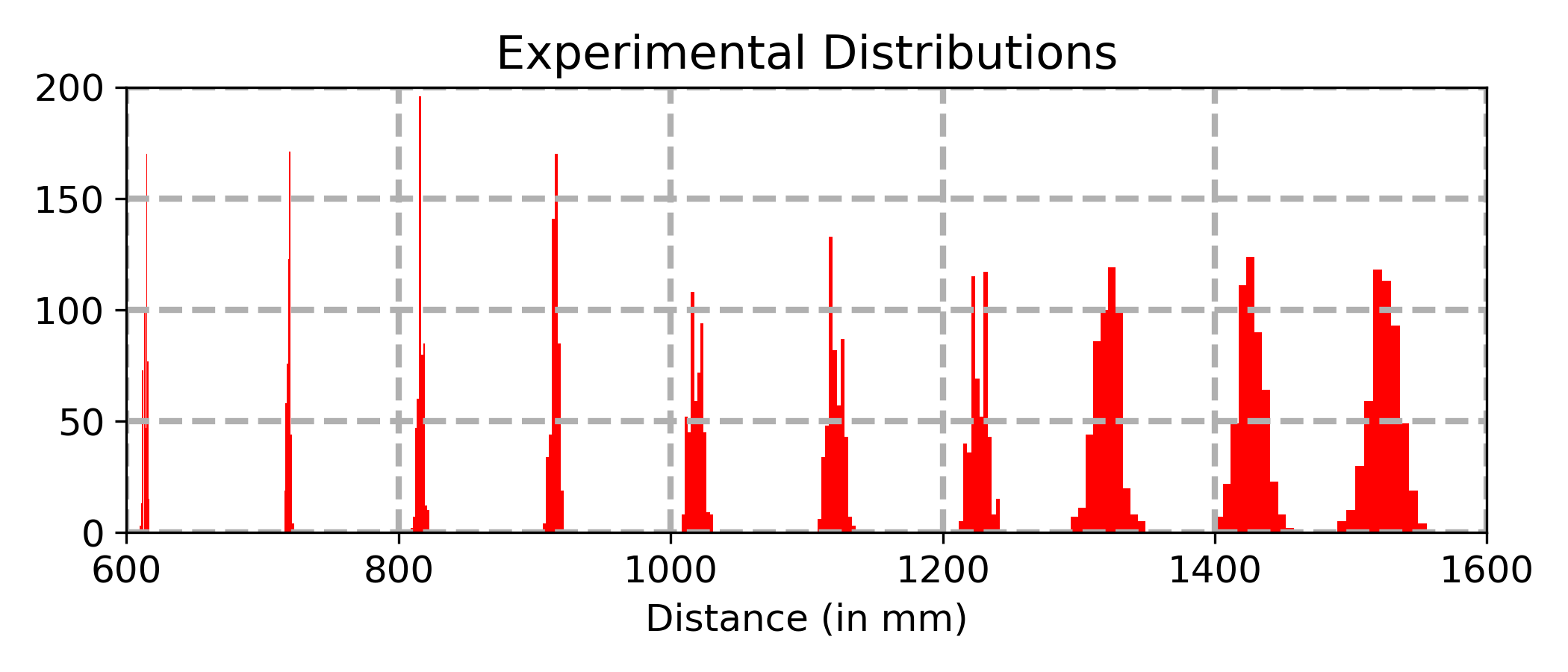}\\
\includegraphics[trim={0cm 0.2cm 0.0cm 0.0cm},clip, width=\linewidth]{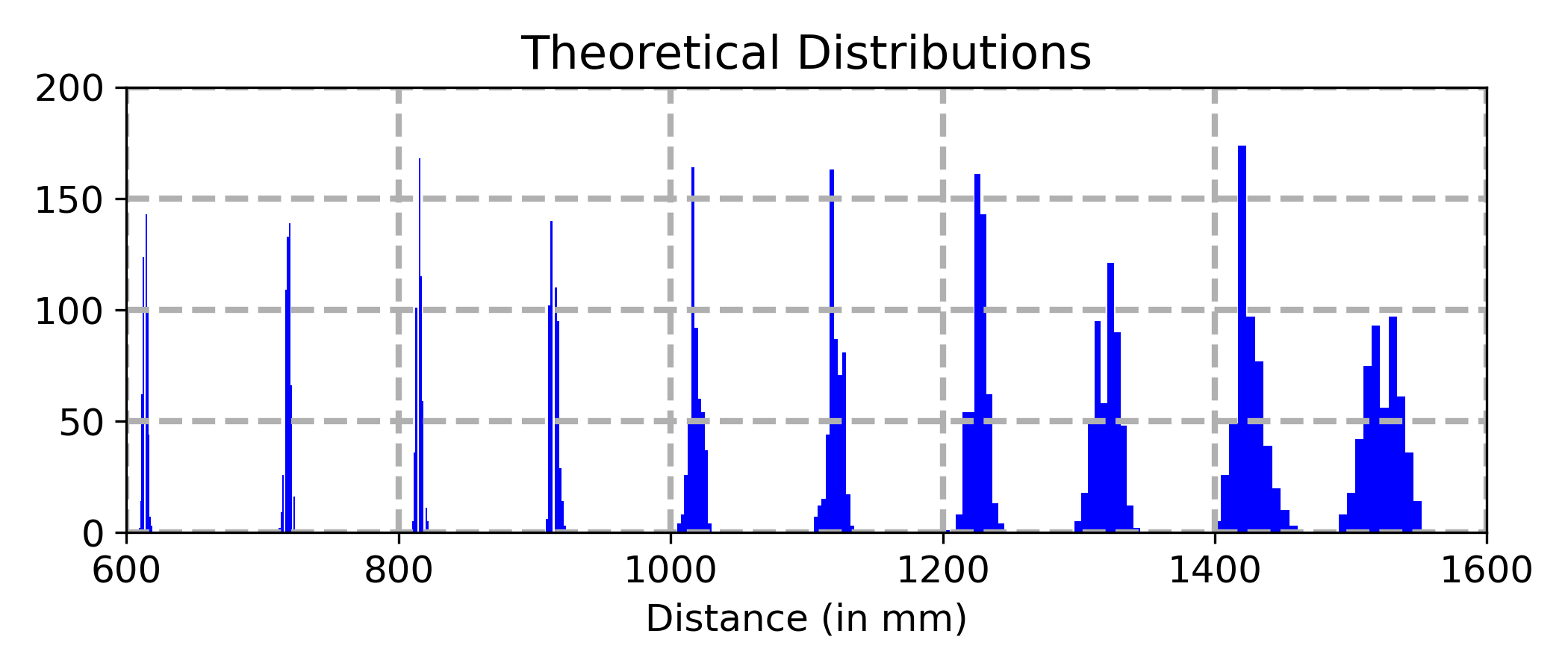}
\end{tabular}
\caption{Histogram of experimental data \textit{(top row)} and corresponding histogram generated using image formation model described in Section III B) \textit{(bottom row)}. For each set of experimental measurements obtained at a fixed distance (differentiated by colour) in the top row, the corresponding theoretical distribution is obtained by generating samples from equation (\ref{eq:nonUniform}) with mean $x_{l,i}^o$ and standard deviation $\sigma_{l,i}$ are chosen as the mean and variance of the corresponding samples from actual data. The variable quantization parameter $R(x)$, which ultimately determines the quantization output $Q(x)$, was set using trial and error to match the quantization bins for each set.}
    \label{fig:histograms}
\end{figure}

Our experimental setup is shown in Fig.\;\ref{fig:setup-expt}, where a cardboard box was placed at a measurable distance from the camera, while ensuring that both the camera and the box axes were aligned to each other. From each frame of the depth camera stream, five randomly selected measurements were collected from a $5\times5$ window around the center of the frame until $100$ measurements for a single distance were obtained. This process was repeated for multiple clusters of measurements, with distances between the camera and the box ranging from $615$mm to $1525$mm in steps of approximately $100$mm. 
The obtained measurements are shown as a histogram in the top half of Fig.\;\ref{fig:histograms}.

\noindent
\textbf{Empirical distribution}. From the top half of Fig.\;\ref{fig:histograms}, two observations can be made. 
First, the variance of measurements for a measurement cluster \textit{increases} as distance increases.  
This suggests that \eqref{eq:sigDependent} is a reasonable assumption, since the noise SD $\sigma$ increases\footnote{Note that the variance of measurements and noise variance $\sigma^2$ are fundamentally two different quantities, though the former is determined by the latter, and thus are positively correlated.} with true distance $x$ of a cluster.
Second, the quantization step size \textit{increases} as distance increases. 
Specifically, at $0.6$m, measurements varied in step of $1$mm, while at $1.5$m, measurements varied in step of $3$mm. 
This justifies the use of a non-uniform quantization function. 

\begin{figure}
\centering
\includegraphics[trim={0 0 40 20},clip,width=0.9\linewidth]{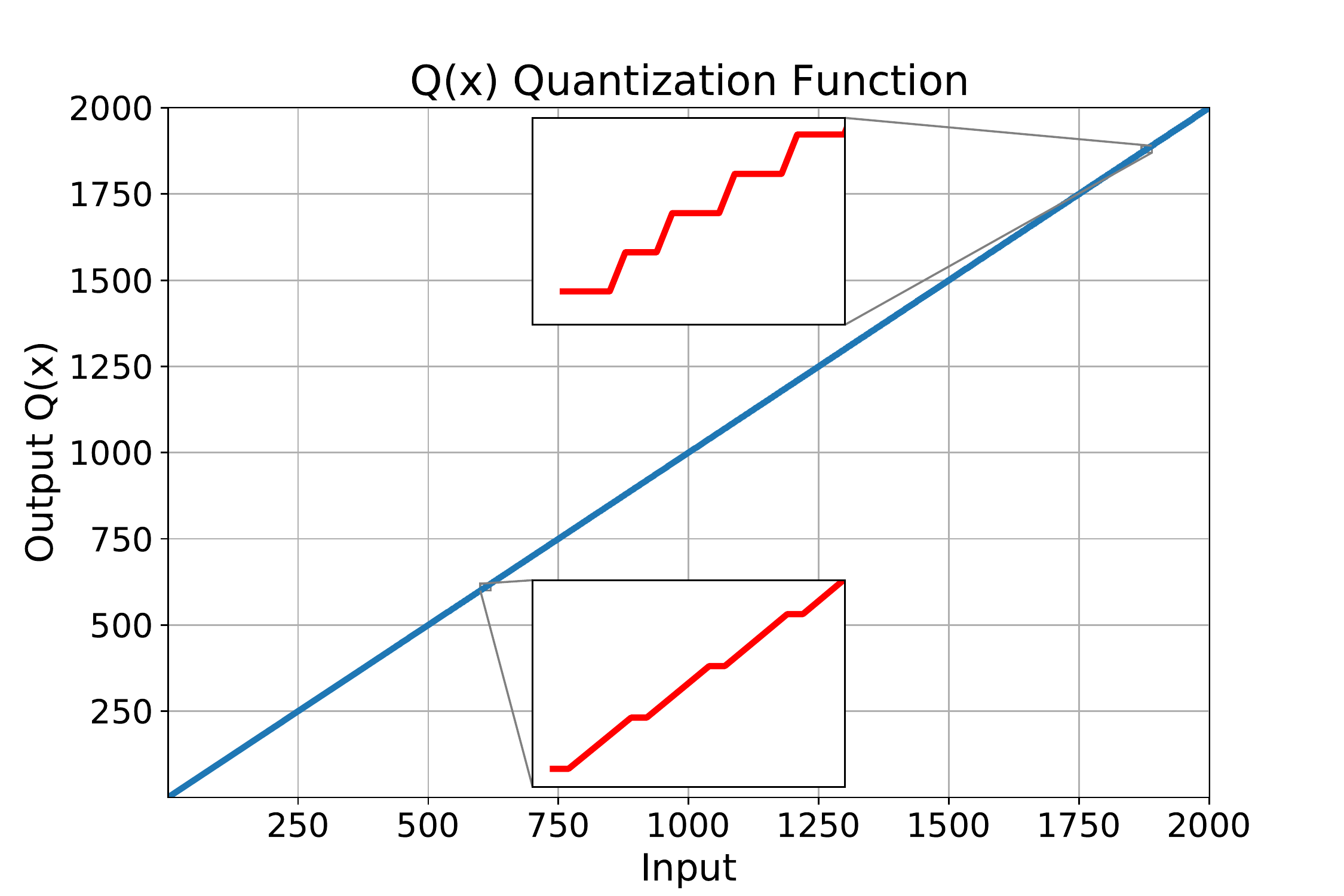}
\vspace{-0.18in}
\caption{Quantization Function $Q(x)$ used to generate second row of histograms in Figure \ref{fig:histograms}. As expected, the quantization function stays close to the $y = x$ line but on zooming in, we notice it is a staircase function with size of steps of the steps increasing as the argument of the function increases.}
    \label{fig:q_x} 
\end{figure}

\noindent
\textbf{Theoretical distribution}. 
We synthesized the theoretical statistics using the image formation model described in \eqref{eq:nonUniform} to match the corresponding experimental measurements. 
To generate these synthetic measurements, we estimated the ground truth to be equal to the mean of the corresponding experimental distribution.
Then, zero-mean Gaussian noise of variance $\sigma^2$ was added to the estimated ground truth followed by quantization using function $Q(x)$. 
The SD of the noise $\sigma$ varied quadratically as described in \eqref{eq:sigDependent}, and the parameters were set as $\alpha=1\times10^{-5}$, $\mu=-528$ and $\kappa=1.4$. 
Quantization mapping $R(x)$ as defined in \eqref{eq:quanPara} was determined similarly to qualitatively match the bins of experimental distributions. 
After data fitting, the parameters were found to be $\phi=500, \theta=500, \rho=200, x_{\min}=10$. Given $R(x)$, the corresponding $Q(x)$ was determined and was plotted in Fig.\;\ref{fig:q_x}. 
The figure shows that the function remains close to the $y=x$ line but on closer inspection, it is a staircase function with quantization bins of increasing size.  

The lower half of Fig.\;\ref{fig:histograms} shows the histogram of the synthetic measurements. 
Based on the similarity between the top and bottom halves of Fig.\; \ref{fig:histograms}, we can conclude that the image formation model described in previous subsection provides a good fit for measurements taken from a real depth sensor.

\subsection{Parameter Estimation for Noise Variance}

We next discuss how we compute noise variance $\sigma^2$ given observed depth measurements in a cluster in Fig.\;\ref{fig:histograms}. Denote by $\cY = \{y_1, y_2, \ldots,  y_M \}$ a set of noise-corrupted and quantized observations of ground-truth depth $x^*$.
For each observation $y_i \in \mathbb{R}^+$, $R(y_i) \in \mathbb{Z}^+$ is the corresponding quantization mapping value.
Thus, $x^* + n_i$ must reside in the quantization bin indexed by $R(y_i)$, \ie, 

\vspace{-0.05in}
\begin{small}
\begin{align}
R(y_i) - \frac{1}{2} \leq \phi \ln (\theta (x^* + n_i) + \rho) - R_{0} + \frac{1}{2} < R(y_i) + \frac{1}{2} .
\end{align}
\end{small}
Thus, we can derive that additive noise $n_i$ is lower- and upper-bounded as follows:

\vspace{-0.05in}
\begin{small}
\begin{align}
n^-(y_i, x^*) &\leq n_i < n^+(y_i, x^*), \nonumber \\
n^-(y_i, x^*) &\triangleq \underbrace{\theta^{-1} \left(\exp\left(\frac{R(y_i) + R_{0} - 1}{\phi}\right) - \rho \right)}_{z^-_i} - x^*,
\nonumber \\
n^+(y_i, x^*) &\triangleq \underbrace{\theta^{-1} \left(\exp\left(\frac{R(y_i) + R_{0}}{\phi}\right) - \rho \right)}_{z^+_i} - x^* .
\label{eq:noiseBound}
\end{align}
\end{small}\noindent
Note that $z^-_i$ and $z^+_i$ are both function of $y_i$. 

Since the $y_i$'s in $\mathcal{Y}$ are independent, we formulate a \textit{maximum likelihood estimation} (MLE) problem for variance $\sigma^2$:

\vspace{-0.05in}
\begin{small}
\begin{align}
\max_{\sigma^2} \rPr( \mathcal{Y} | \sigma^2) = & \prod_{i=1}^M \rPr( y_i | \sigma^2) 
\nonumber \\
=&\prod_{i=1}^M \rPr \left( \left. n^-(y_i,x^*) \leq n_i < n^+(y_i,x^*) \right| \sigma^2 \right).
\label{eq:ML}
\end{align}
\end{small}\noindent
Since $n_i \sim \cN(0,\sigma^2)$, we can write

\vspace{-0.05in}
\begin{small}
\begin{align}
\rPr( \mathcal{Y} | \sigma^2)
&= \prod_{i=1}^M \int_{\cR_i} \frac{1}{ \sqrt{2 \pi}\sigma} \exp \left(-\frac{n_i^2}{2 \sigma^2} \right)  \;  d\, n_i \triangleq p(\sigma)
\label{eq:ML2}
\end{align}
\end{small}\noindent
where the region of integration $\cR_i$ is
\begin{align}
\cR_i= \left\{ n_i ~|~ n^-(y_i,x^*) \leq n_i < n^+(y_i,x^*) \right\}. 
\end{align}

The probability $p(\sigma)$ in \eqref{eq:ML2} is difficult to maximize over $\sigma$ for two reasons.
First, each term in the product is of the form $\int_{\cR_i} e^{-n_i^2/2\sigma^2} d n_i$, which has no closed-form expression.
Second, \eqref{eq:ML2} involves a product of $M$ terms, each integrating over a different region $\cR_i$.  
Thus, we propose the following fast search procedure to find a near-optimal $\sigma$. 

For simplicity, first we assume there is only one observation, \ie, $M=1$. 
Then the optimization problem becomes

\begin{figure}
\begin{center}
\begin{tabular}{cc}
\includegraphics[width=1.65in]{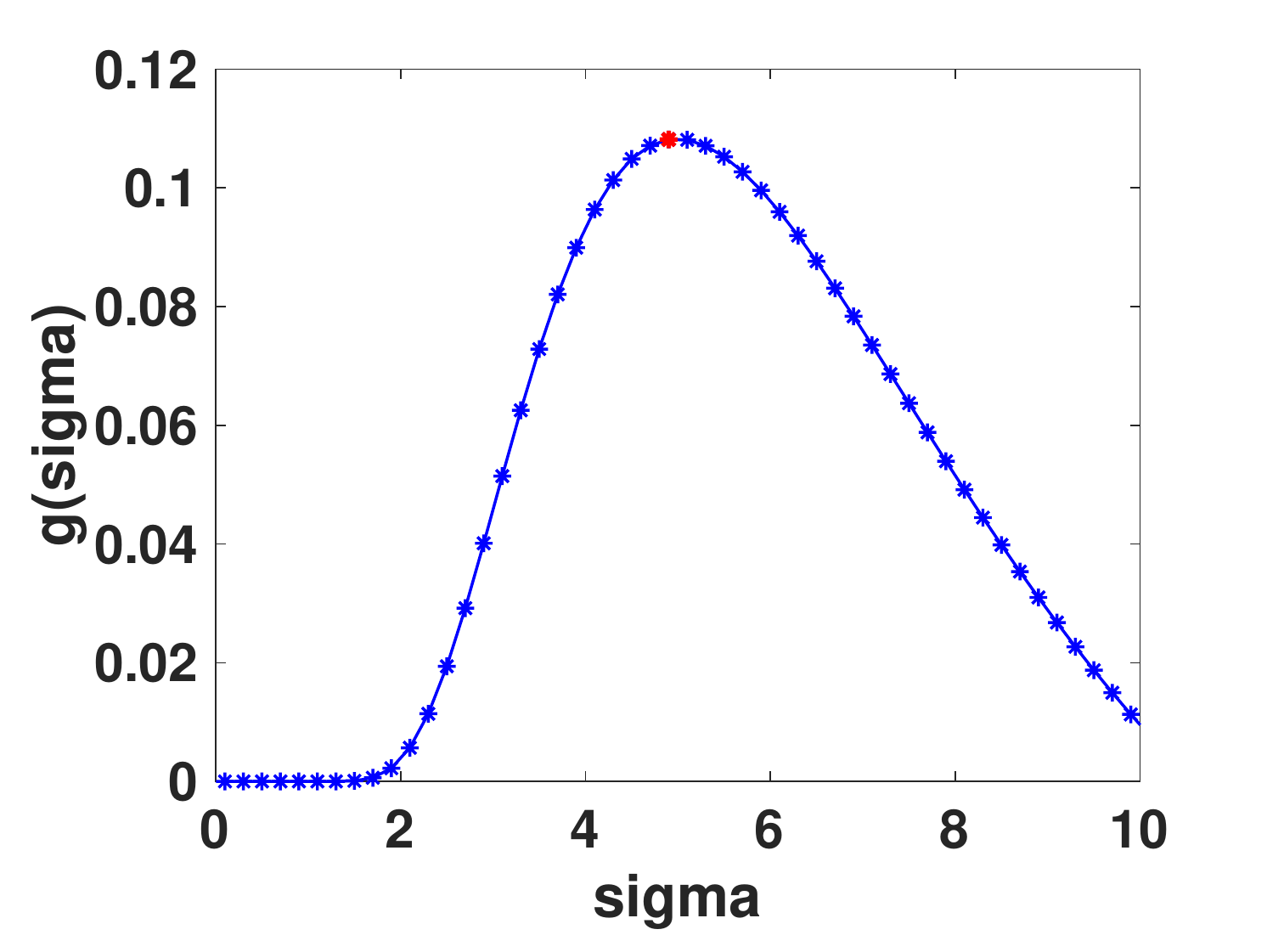} & \includegraphics[width=1.65in]{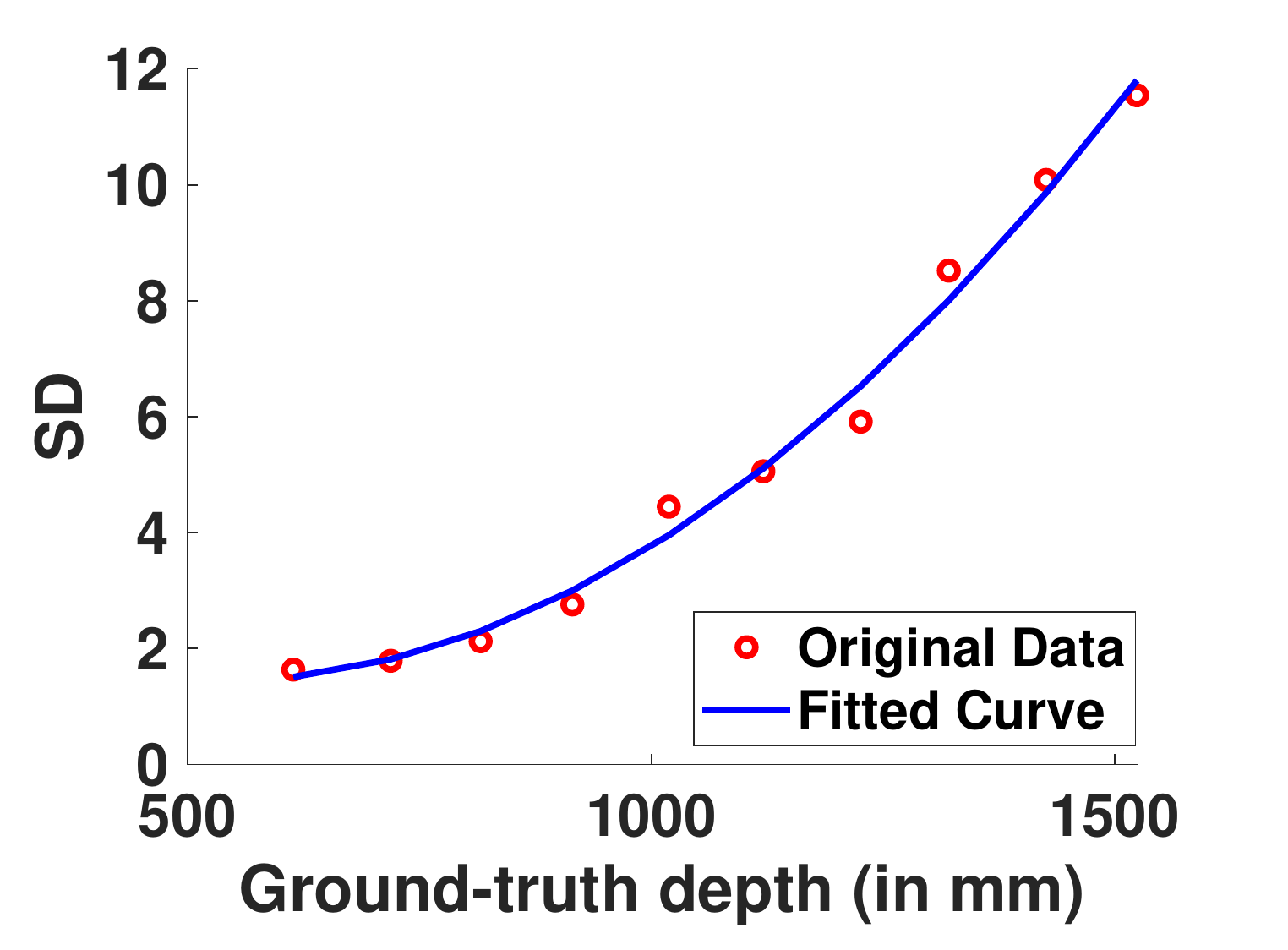}\\
\parbox{.45\linewidth}{\centering\small (a)} & \parbox{.45\linewidth}{\centering\small (b)}
\end{tabular}
\end{center}
\vspace{-0.15in}
\caption{Illustration of (a) unimodal objective $\tilde{g}(\sigma)$ in \eqref{eq:cdf_appr} with a single local maximum at $\sigma_1^*$ and (b) parameterization of $\sigma$ in \eqref{eq:sigDependent}.}
\label{fig:sigma}
\end{figure}

\vspace{-0.05in}
\begin{small}
\begin{align}
\sigma_1^* = &\arg \max_{\sigma} \left(\Phi \left( n^+(y_1,x^*) \right) - \Phi \left( n^-(y_1,x^*) \right)  \right)
\nonumber \\
\approx& \arg \max_{\sigma}
\left( \left( \frac{n^-(y_1,x^*) \sigma}{(n^-(y_1,x^*))^2+\sigma^2} \right) \exp \left( - \frac{(n^-(y_1,x^*))^2}{2 \sigma^2} \right)  \right. \nonumber \\
& \left. -\left( \frac{n^+(y_1,x^*)\sigma}{(n^+(y_1,x^*))^2+\sigma^2} \right) \exp \left( - \frac{(n^+(y_1,x^*))^2}{2 \sigma^2} \right)\right)
\triangleq \tilde{g}(\sigma)
\label{eq:cdf_appr}
\end{align}
\end{small}\noindent
where $\Phi(x)$ is the \textit{cumulative distribution function} (CDF) of $n_1$, and in \eqref{eq:cdf_appr} we use the approximation of $\Phi(x)$ in \cite{pishro2016introduction}, \ie, 
\begin{align}
\Phi(x)&=\int_{-\infty}^x \frac{1}{ \sqrt{2 \pi}\sigma} \exp \left(-\frac{n_1^2}{2 \sigma^2} \right)  \;  d n_1 \nonumber \\
&\approx 1-\frac{1}{ \sqrt{2 \pi}} \frac{x\sigma}{x^2+\sigma^2}  \exp \left (-\frac{x^2}{2 \sigma^2} \right).
\end{align}
One can show that $\tilde{g}(\sigma)$ in \eqref{eq:cdf_appr} is a unimodal objective with a single local maximum at $\sigma_1^*$, as shown in Fig.\;\ref{fig:sigma}(a).
This means that $\sigma_1^*$ can be computed using numerical methods like the Newton's method \cite{press2007numerical} to locate $\sigma_1^*$ where $\tilde{g}'(\sigma_1^*) = 0$.

More generally, when there are $M > 1$ observations, the approximate objective $\tilde{g}(\sigma)$ becomes:

\begin{small}
\begin{align}
\tilde{g}(\sigma) &= 
\prod_{i=1}^M 
\left( \left( \frac{n^-(y_i,x^*) \sigma}{(n^-(y_i,x^*))^2+\sigma^2} \right) \exp \left( - \frac{(n^-(y_i,x^*))^2}{2 \sigma^2} \right) \right. \nonumber \\
&\left.-\left( \frac{n^+(y_i,x^*)\sigma}{(n^+(y_i,x^*))^2+\sigma^2} \right) \exp \left( - \frac{(n^+(y_i,x^*))^2}{2 \sigma^2} \right) \right).
\label{eq:cdf_appr2}
\end{align}
\end{small}\noindent
The optimal $\sigma^*$ in this general $M > 1$ case is lower- and upper-bounded as follows
\begin{align}
\min_{i \in \{1, \ldots, M\}} (\sigma_i^*) \leq \sigma^* \leq \max_{i \in \{1, \ldots, M\}} (\sigma_i^*).
\end{align}
Thus, we can first compute the lower / upper bounds, then use the \textit{Golden-section} (GS) search \cite{press2007numerical} to find $\sigma^*$.
GS finds the optimal value $\sigma^*$ that maximizes / minimizes $\tilde{g}(\sigma^*)$ via an iterative search, given $\tilde{g}(\sigma)$ is unimodal\footnote{Though in general the multiplication of unimodal functions is not necessarily unimodal, in our experiments we found this to be the case.}. 
See \cite{press2007numerical} for details.

Finally, given computed $\sigma^*$’s for different clusters with corresponding ground-truth depth $x^*$'s, we can parameterize $\alpha=1\times10^{-5}$, $\mu=-528$ and $\kappa=1.4$ in \eqref{eq:sigDependent} via nonlinear least-squares algorithm 
\cite{coleman1996interior} as shown in Fig.\;\ref{fig:sigma}(b).

\section{Approximation and Optimization}
\label{sec:formulate}
The image formation model in Section \ref{subsec:formation} is difficult to optimize directly;
we first present a practical approximation. 
We then introduce a mapping function from left-view pixels to right-view pixels assuming FoV overlapped.
We next define a likelihood term and a signal prior for individual pixel rows in depth images.
Finally, we formulate a MAP optimization problem and derive a corresponding algorithm based on AGD to reconstruct a target pixel row.

\subsection{Approximation of Image Formation Model}
\label{subsec:layer}

Additive noise in our image formation model, as discussed in Section\;\ref{subsubsec:noise}, is signal-dependent, which is difficult to address directly.
Thus, as a pre-processing step, we first segment a depth image into non-overlapping \textit{layers}, each of approximately the same depth, and then assume the same fixed noise variance in each layer. 
This constitutes a reasonable approximation in practice, as a depth image is typically composed of a static indoor background plus one or more foreground objects, each with roughly the same distance to the depth sensor.

\begin{figure}
\begin{center}
\begin{tabular}{cc}
\includegraphics[width=1.5in]{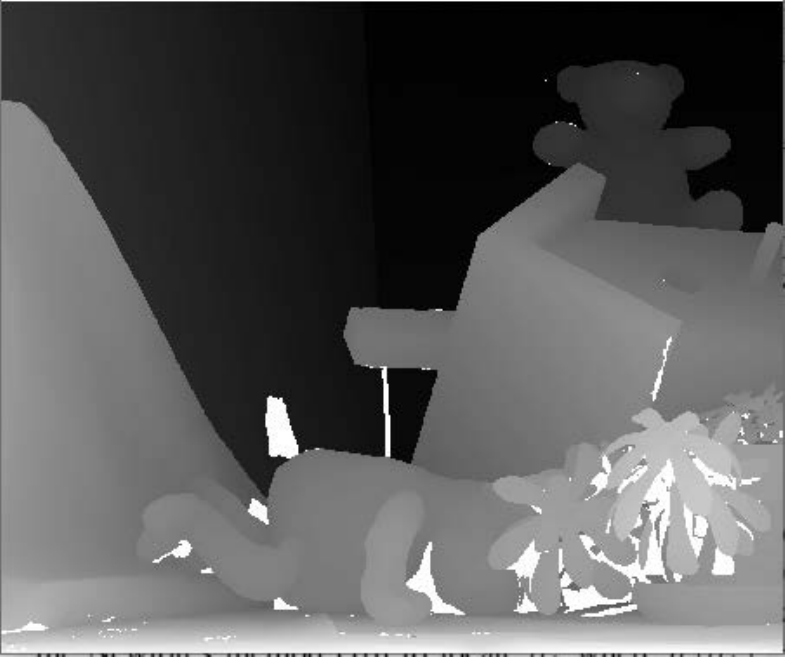}&\includegraphics[width=1.5in]{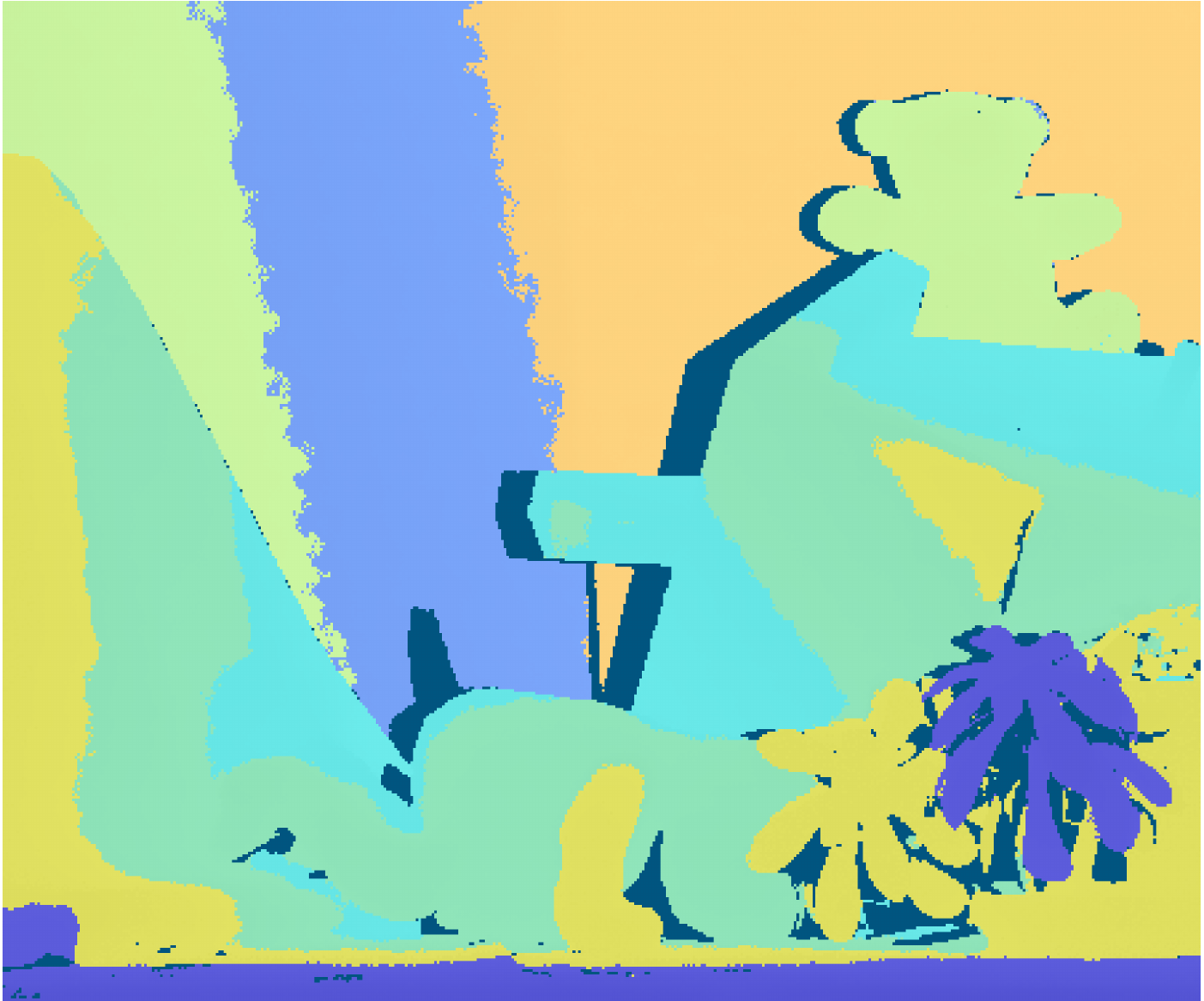}\\
\parbox{.45\linewidth}{\centering\small (a)} & \parbox{.45\linewidth}{\centering\small (b)}\\
\end{tabular}
\end{center}
\vspace{-0.1in}
\caption{An example of depth map segmentation: (a) the ground-truth disparity map, (b) the segmented eight layers of pre-filtered depth map, each of approximately the same depth with the same noise variance.}
\label{fig:seg}
\end{figure}

To achieve robust image segmentation, we first segment the corrupted depth image into layers using the k-means algorithm, where the optimal number of layers is determined by the Elbow method \cite{bholowalia2014ebk}.
We then pre-filter it using \textit{bilateral filter} (BF) \cite{tomasi98} layer by layer, where for each layer we compute the SD of the layer for the range filter parameter, while fixing the domain filter parameter at $3$ for all layers. 
Finally, we perform segmentation again on the pre-filtered depth map. 
See Fig.\;\ref{fig:seg} for an example.
 

After segmentation, for each layer $\zeta$ we compute an average depth $\bar{x}_{\zeta}$ given observed depth pixels of the layer, then use \eqref{eq:sigDependent} to compute a noise variance $\bar{\sigma}_{\zeta}^2$ for the layer. 
In the sequel, we will assume a constant and pre-computed noise variance $\bar{\sigma}_{\zeta}^2$ for each depth layer $\zeta$.

\subsection{View-to-view Mapping}
\label{subsec:V2VMapping}

Consider two depth images of adjacent viewpoints with FoV overlapped, as shown in Fig.\;\ref{fig:model}. 
Pixels in the left and right rectified images corresponding to overlapping FoV are projections of the same object surface onto two different camera planes, and thus are related.
We optimize a left pixel row of a layer exploiting this inter-view redundancy as follows. (Optimization of a right pixel row is similar and thus omitted.)
Specifically, denote by $\x_l$ a row with $N$ available pixels in a layer of the left view, and by $\x_r$ a corresponding \textit{sub-row} of $M$ pixels, where $M \leq N$, capturing the overlapping spatial region.  
For simplicity, we assume that there is no occlusion in the overlapping region between $\x_l$ and $\x_r$.

Given that the two depth images are rectified, we employ a known 1D warping procedure \cite{jin2016region} to relate $\x_l$ and $\x_r$.  
For the $i$-th pixel in the left view, $x_{l,i}$, its (non-integer) horizontal position $s(i,x_{l,i}) \in \mathbb{R}$ in the right view after projection is
\begin{align}
s(i, x_{l,i}) &= i-\delta(x_{l,i}), \nonumber \\
\delta(x_{l,i}) &= \frac{ f D}{x_{l,i}} 
\label{eq:disp}
\end{align}
where $\delta(x_{l,i}) \in \mathbb{R}$ is the \textit{disparity} of $x_{l,i}$, $f \in \mathbb{R}^+$ is the camera focal length, and $D \in \mathbb{R}^+$ is the baseline between the two capturing cameras.
Note that $s(\cdot)$ is a function of \textit{both} left pixel's integer horizontal position $i$ \textit{and} depth value $x_{l,i}$. 
Here, we use the pre-filtered depth values $\hat{x}_{l,i}$ to compute reliable disparities $\delta(\hat{x}_{l,i})$ for view-to-view mapping.



\begin{figure}
\begin{center}
\includegraphics[width=2.8in]{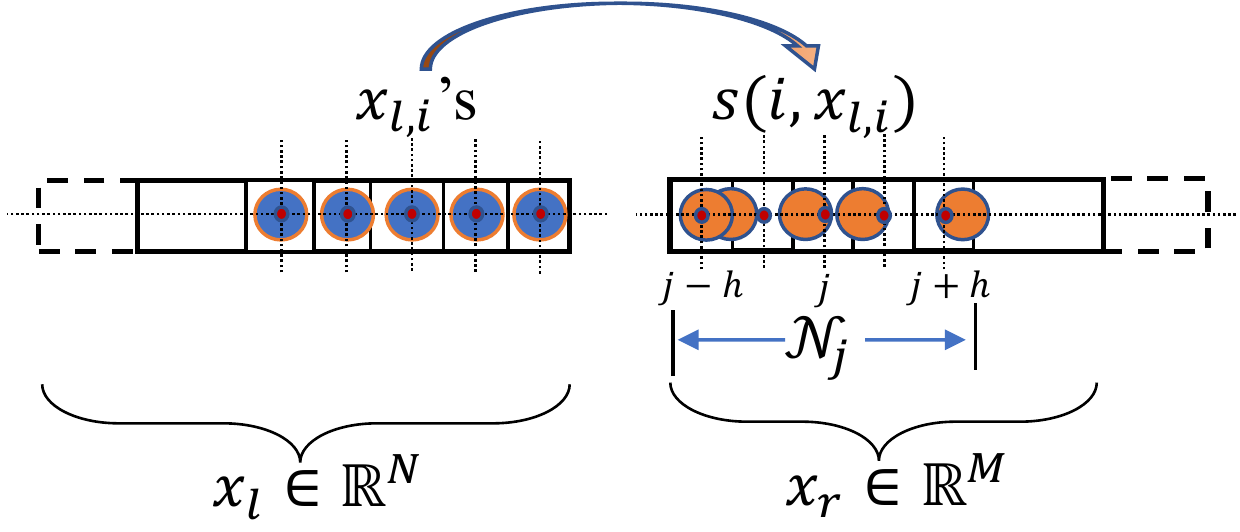}
\vspace{-0.1in}
\caption{\footnotesize Illustration of view-to-view mapping: using left pixels $x_{l,i}$'s with mapped (non-integer) positions $s(i,x_{l,i})$ falling within right pixel $x_{r,j}$'s \textit{neighborhood} $\cN_j \triangleq (j-h,j+h)$ to interpolate right pixel $x_{r,j}$.
}
\label{fig:mapping}
\end{center}
\end{figure}

\subsubsection{Sparse BF Interpolation}
Assuming that the object surface is smooth, we interpolate each right pixel $x_{r,j}$ in sub-row $\x_r$ as a \textit{sparse} weighted combination of left pixels $x_{l,i}$'s; 
\ie, each $x_{r,j}$ is linearly interpolated using $x_{l,i}$'s with mapped positions $s(i,x_{l,i})$ falling within $x_{r,j}$'s \textit{neighborhood} $\cN_j \triangleq (j-h,j+h)$, where $h \in \mathbb{Z}^+$ is a parameter that determines neighborhood size. 
See Fig.\;\ref{fig:mapping} for an illustration, where a left pixel $x_{l,i}$ maps to a location within neighborhood $\cN_j$ of right pixel $x_{r,j}$, and thus will be used for linear interpolation of $x_{r,j}$.

We use linear interpolation for two practical reasons.
First, it has been commonly used in the multiview-plus-depth imaging literature and shown to obtain high quality rendering \cite{merkle2008effect,motz2015re}.
Second, linear interpolation will ease computation of our optimization problem, to be formulated in Section\;\ref{subsec:MAP}.

In matrix form, we write the interpolation as
\begin{align}
\x_r=\W(\x_l) \, \x_l= \g(\x_l)
\label{eq:interpolate}
\end{align}
where $\W(\x_l) \in \mathbb{R}^{M \times N}$ is a sparse weight matrix.
We model the weight $w_{ij}$ between right pixel $x_{r,j}$ and left pixel $x_{l,i}$ using a Gaussian kernel based on the spatial proximity between the projected position $s(i,x_{l,i})$ of left pixel $i$ and the target right pixel position $j$ in $x_{r,j}$, \ie, 
\begin{align}
w_{ij}&=\frac{1}{\bar{w}_j} \exp \left(- \frac{(s(i,x_{l,i})-j)^2}{\sigma_s^2}\right),
\label{eq:weight} \\
\bar{w}_j &= \sum_{m | s(m,x_{l,m}) \in \cN_j} \exp \left(- \frac{(s(m,x_{l,m})-j)^2}{\sigma_s^2}\right)
\end{align}
where $\bar{w}_j$ is a normalization term so that $\sum_i w_{ij} = 1$. 
Gaussian kernel is commonly used in image filtering schemes such as BF~\cite{tomasi98}. 
In words, weight $w_{ij}$ is large if the distance between location $s(i,x_{l,i})$ of mapped left pixel $x_{l,i}$ and location $j$ of target right pixel $x_{r,j}$ is small.
To simplify \eqref{eq:weight}, we assume $\bar{w}_j$ is a constant. 
Combining with \eqref{eq:disp}, \eqref{eq:weight} is rewritten as
\begin{align}
w_{ij}= \bar{w}_j^{-1} \exp \left(-\frac{(i-f D      x_{l,i}^{-1}-j)^2}{\sigma_s^2} \right) .
\label{eq:nweight}
\end{align}

Given \eqref{eq:nweight}, the interpolation of $\g(\x_l)$ in \eqref{eq:interpolate} is differentiable w.r.t. $\x_l$. 
We use the first-order Taylor series expansion around $\hat{\x}_l$, the pre-filtered estimate of $\x_l$, to obtain a linear approximation.
Thus, 
\begin{align}
\g(\x_l) \approx ~ & \g(\hat{\x}_l) + \g'(\hat{\x}_l)(\x_l-\hat{\x}_l) \nonumber\\
= ~ & \g'(\hat{\x}_l)\x_l + \g(\hat{\x}_l)-\g'(\hat{\x}_l)\hat{\x}_l \nonumber\\
=~ & \H \x_l+ \e
\label{eq:approxG}
\end{align}
where $\H = \g'(\hat{\x}_l) = \left[ \frac{\partial g_j(\hat{\x}_l)}{\partial x_{l,i}} \right]_{j,i} \in \mathbb{R}^{M \times N} $ is the \textit{Jacobian matrix}
(first-order partial derivatives) of $\g(\x_l)$ at $\x_l=\hat{\x}_l$, and $\e=\g(\hat{\x}_l)-\g'(\hat{\x}_l)\hat{\x}_l$ is a constant vector in $\mathbb{R}^M$.
Note that row $j$ of $\H$ contains only non-zero entries $H_{j,i}$ corresponding to left pixels $x_{l,i}$ that map to positions $s(i,x_{l,i})$ within right pixel $x_{r,j}$'s neighborhood $\cN_j$. 
Thus, $\H$ is sparse, where sparsity depends on the number of left pixels $x_{l,i}$ mapping to $\cN_j$.


\subsection{Likelihood Term}

Since every entry of the zero-mean additive noise $\n_l \in \mathbb{R}^N$ follows an independent Gaussian distribution, the probability density function (pdf) of $\n_l$ is
\begin{align}
\rPr (\n_l) &=
\prod_{i=1}^N \rPr(n_{l,i}) = 
\prod_{i=1}^N \frac{1}{\sqrt{2\pi} \sigma_{l,i}} \exp \left( - \frac{n_{l,i}^2}{2\sigma_{l,i}^2} \right)
\label{eq:noiseModel}
\end{align}
where $\sigma_{l,i}^2$ is the noise variance for left pixel $i$ given that it belongs to a known layer $\zeta$ during image segmentation, as discussed in Section\;\ref{subsec:layer}. 
Recall that $\sigma_{l,i}$ is computed via \eqref{eq:sigDependent}, using the average depth $\bar{x}_{\zeta}$ of layer $\zeta$.

Given observation $\y_l$, the likelihood term $\rPr(\y_l | \x_l)$ is
\begin{align}
\rPr(\y_l | \x_l) 
= \prod_{i=1}^N \int_{\cR_{l,i}} \rPr(n_{l,i}) \; d\,n_{l,i}
\label{eq:likelihood}
\end{align}
where the region of integration $\cR_{l,i}$ is

\vspace{-0.1in}
\begin{small}
\begin{align}
\cR_{l,i} &= \left\{ n_{l,i} ~|~ n^-(y_{l,i},x_{l,i}) \leq n_{l,i} < n^+(y_{l,i},x_{l,i}) \right\}.
\label{eq:region}
\end{align}
\end{small}\noindent
The lower and upper bounds,  $n^-(y_{l,i},x_{l,i}) = z^-_{l,i} - x_{l,i}$ and $n^+(y_{l,i},x_{l,i}) = z^+_{l,i} - x_{l,i}$, are defined in \eqref{eq:noiseBound}. 

\begin{figure}
\begin{center}
\includegraphics[width=3in]{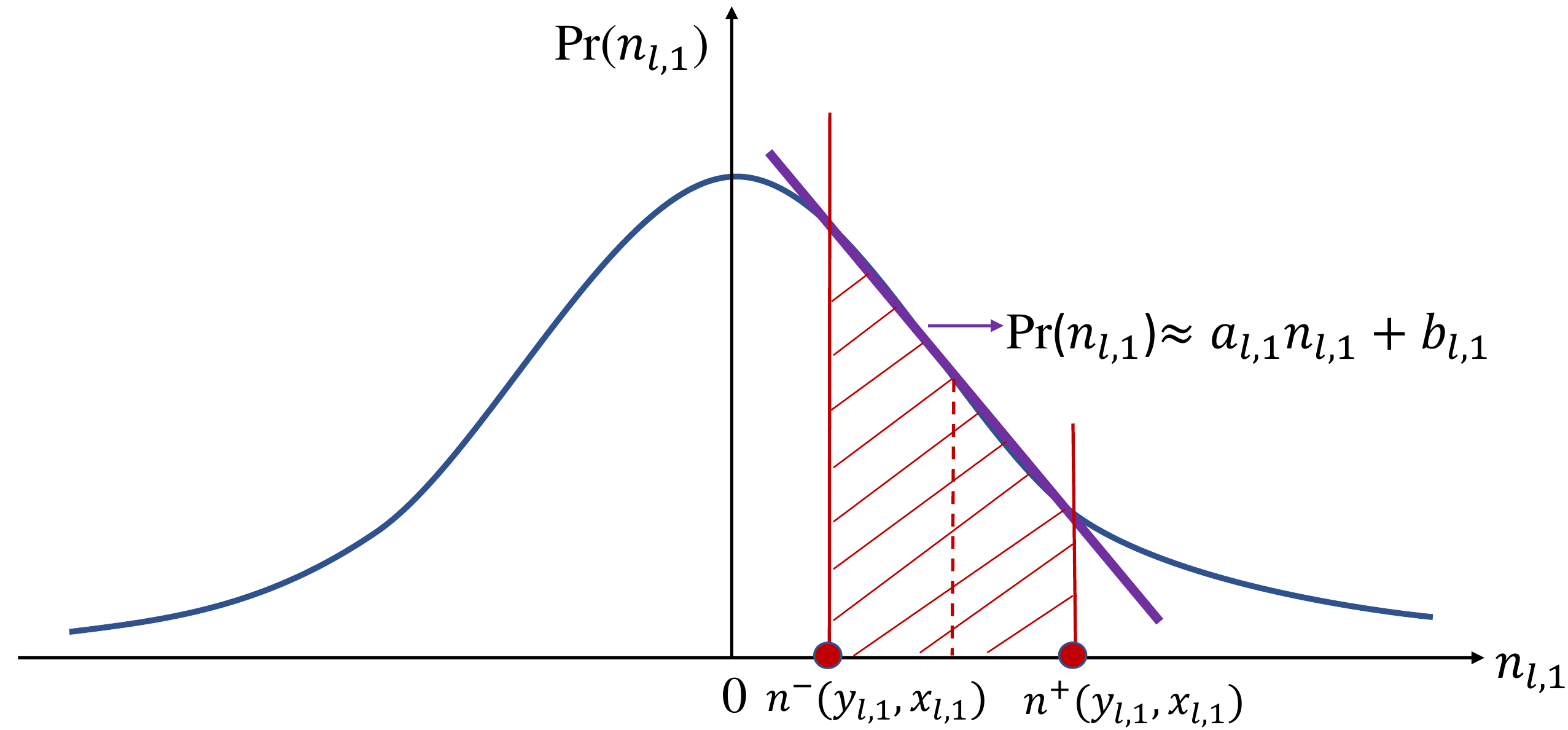}
\caption{\footnotesize Linear approximation of the Gaussian pdf.}
\label{fig:likelihood}
\end{center}
\vspace{-0.1in}
\end{figure}

Each integration in \eqref{eq:likelihood} is over a Gaussian pdf, which has no closed-form expression.
For ease of later optimization, we approximate each Gaussian function $\rPr(n_{l,i})$ as a linear function over the region of integration $\cR_{l,i} = \left[ n^-(y_{l,i},x_{l,i}), n^+(y_{l,i},x_{l,i}) \right]$, \ie, 
\begin{align}
\rPr(n_{l,i}) \approx 
a_{l,i} n_{l,i} + b_{l,i}, 
~~~ n_{l,i} \in \cR_{l,i}.
\label{eq:linearApprox}
\end{align}
$a_{l,i}$ and $b_{l,i}$ are constant scalars computed using Taylor series expansion at the initial estimate $n^0_{l,i}=y_{l,i}-\tilde{x}_{l,i}$, where the recovered $\tilde{x}_{l,i}$ is updated throughout the algorithm iterations (to be discussed in detail later).   
See Fig.\;\ref{fig:likelihood} for an illustration.

The linear approximation \eqref{eq:linearApprox} is good if i) the integration region $\cR_i$ is narrow, or ii) the Gaussian function $\text{Pr}(n_{l,i})$ is flat. 
When the captured depth pixel $x_{l,i}$ is near, the corresponding quantization bin is small, and hence $\cR_i$ is narrow.
On the other hand, when $x_{l,i}$ is far, the noise variance $\sigma_{l,i}^2$ is large, and hence $\text{Pr}(n_{l,i})$ is flat.
As we will demonstrate in Section\;\ref{sec:results}, in either case the linear approximation is sufficiently accurate.

Denote by $\1_i$ a suitably long \textit{canonical vector} of all zeros except entry $i$ is 1. 
We now rewrite \eqref{eq:likelihood} as
\begin{align}
\rPr(\y_l | \x_l) \approx & 
\prod_{i=1}^N \int_{\cR_{l,i}} 
\left( a_{l,i} n_{l,i} + b_{l,i} \right) d\, n_{l,i} \nonumber \\
=& \prod_{i=1}^N 
(\tilde{\a}_{l,i}^{\top} \x_{l} + \tilde{b}_{l,i})
\label{eq:multiInt} 
\end{align}
where $\tilde{\a}_{l,i}^{\top} = a_{l,i} (z^-_{l,i} - z^+_{l,i}) \1_i^{\top}$ and $\tilde{b}_{l,i} = \frac{a_{l,i}}{2} ((z^+_{l,i})^2 - (z^-_{l,i})^2) + b_{l,i} (z^+_{l,i} - z^-_{l,i})$.
\eqref{eq:multiInt} is proven in the Appendix.

Similarly, for noise $\n_r \in \mathbb{R}^M$, we can write
\begin{align}
\rPr(\y_r | \x_l) \approx & 
\prod_{j=1}^M \int_{\cR_{r,j}} 
\left( a_{r,j} n_{r,j} + b_{r,j} \right) d\, n_{r,j} \nonumber \\
\approx& \prod_{j=1}^M 
(\bar{\a}_{r,j}^{\top} \H \x_{l} +  \bar{b}_{r,j})
\label{eq:multiInt2} 
\end{align}
where $\bar{\a}_{r,j}^{\top} = a_{r,j}(z^-_{r,j} - z^+_{r,j}) \1_j^{\top}$ and $\bar{b}_{r,j} = \bar{\a}_{r,j}^{\top} \e + \frac{a_{r,j}}{2} ((z^+_{r,j})^2 - (z^-_{r,j})^2) + b_{r,j} (z^+_{r,j} - z^-_{r,j})$.
\eqref{eq:multiInt2} is proven in the Appendix.

\subsection{Signal Prior}

As done in graph-based image processing work \cite{kalofolias2016learn, egilmez2017graph, bai2018graph}, we model the similarities of pixel pairs in $\x_l$ using a graph Laplacian matrix $\L_l$, and thus prior $\rPr(\x_l)$ can be written as
\begin{align}
\rPr(\x_l) = \exp \left( - \frac{\x_l^{\top} \L_l \x_l}{\sigma_p^2} \right) 
\label{eq:prior}
\end{align}
where $\sigma_p > 0$ is a weight parameter that is a function of noise variance.
$\sigma_p$ determines the importance of prior $\text{Pr}(\x_l)$ in a MAP formulation. 
In particular, we define $\sigma_p$ as
\begin{align}
\sigma_p^2 = \frac{1}{g_1 \bar{\sigma}_{\zeta}^2 + g_2 }
\label{eq:sigma_p}
\end{align}
where $g_1>0$ and $g_2$ are empirically fitted constants for a dataset. 
Recall that $\bar{\sigma}_{\zeta}^2$ is the noise variance for the layer $\zeta$ in which pixels in $\x_l$ reside.
In words, \eqref{eq:sigma_p} states that a larger noise variance $\bar{\sigma}_{\zeta}^2$ leads to a smaller $\sigma_p^2$, making prior $\text{Pr}(\x_l)$ more important in a noisy setting.

We assume that the previous $K$ pixel rows in the left depth image have been enhanced first.
Assuming in addition that the next row $i$ follows a similar image structure, $\L_l$ can be learned from the previous $K$ rows. 
See Section\;\ref{sec:learn} for details.

\subsection{MAP Formulation}
\label{subsec:MAP}

We now formulate a MAP problem for $\widetilde{\x}_l$ as follows.
\begin{small}
\begin{align}
\max_{\x_l} &~ \rPr(\y_l,\y_r|\x_l,\x_r) \; \rPr(\x_l, \x_r) \\ 
=&~ \rPr(\y_l,\y_r|\x_l,\g(\x_l)) \; \rPr(\x_l, \g(\x_l)) 
\label{eq:sub} \\
=&~  \rPr(\y_l|\x_l) \rPr(\y_r|\g(\x_l)) \; \rPr(\x_l) \rPr(\g(\x_l)) 
\label{eq:ind} \\
\approx & \prod_{i=1}^N (\tilde{\a}_{l,i}^{\top} \x_{l} + \tilde{b}_{l,i}) 
 \prod_{j=1}^M (\bar{\a}_{r,j}^{\top} \H \x_{l} + \bar{b}_{r,j})
\nonumber \\
&\exp \left( - \frac{\x_l^{\top} \L_l \x_l}{\sigma_p^2} \right) \exp \left( - \frac{\g(\x_l)^{\top} \L_r \g(\x_l)}{\sigma_p^2} \right)
\label{eq:map}
\end{align}
\end{small}\noindent
where in \eqref{eq:sub} we substituted $\g(\x_l)$ for $\x_r$, and in \eqref{eq:ind} we split up the first term since left and right noise, $\n_l$ and $\n_r$, are independent.
Note that parameters $\tilde{\a}_{l,i}$, $\tilde{b}_{l,i}$, $\bar{\a}_{r,j}$ and $\bar{b}_{r,j}$ are all computed from Taylor series expansion using the updated $\tilde{\x}_l$ in each algorithm iteration.

To ease optimization, we minimize the negative log of \eqref{eq:map}:  

\vspace{-0.1in}
\begin{small}
\begin{align}
\min_{\x_l} & - \ln \prod_{i=1}^N (\tilde{\a}_{l,i}^{\top} \x_{l} + \tilde{b}_{l,i}) 
- \ln \prod_{j=1}^M (\bar{\a}_{r,j}^{\top} \H \x_{l} + \bar{b}_{r,j})  \nonumber \\
&  +\x_l^{\top} \L_l \x_l+ \g(\x_l)^{\top} \L_r \g(\x_l) \nonumber \\
= & - \sum_{i=1}^N \ln (\tilde{\a}_{l,i}^{\top} \x_{l} + \tilde{b}_{l,i}) 
- \sum_{j=1}^M \ln (\bar{\a}_{r,j}^{\top} \H \x_{l} + \bar{b}_{l,i})  \nonumber \\
&  + \frac{1}{\sigma_p^2} \left( \x_l^{\top} \L_l \x_l + \x_l^{\top} \H^{\top} \L_r \H \x_l+2\;\e^{\top} \L_r \H \x_l \right)  .
\label{eq:fmap}
\end{align}
\end{small}\noindent
\eqref{eq:fmap} is an unconstrained convex and differentiable objective, but has no closed-form solution. 
Thus, we solve for its minimum efficiently using AGD \cite{nesterov1983method, bubeck2014convex}. 
AGD is an extension of gradient descent (GD) that provably achieves a convergence rate of $1/t^2$ after $t$ steps in the convex scenario. 
To execute AGD efficiently, selecting an appropriate step size is important. 
We next overview AGD and our choice of step size in our optimization.


\subsection{Optimization of $\x_l$}

For  notation  simplicity, we rewrite the objective in \eqref{eq:fmap} as
\begin{align}
\min_{\x_l} f(\x_l)=& - \sum_{i=1}^N \ln (\tilde{\a}_{l,i}^{\top} \x_{l} + \tilde{b}_{l,i})  - \sum_{j=1}^M \ln (\bar{\a}_{r,j}^{\top} \H \x_{l} + \bar{b}_{r,j})   \nonumber\\
& + \frac{1}{\sigma_p^2} \left( \x_l^{\top} \cL \x_l + 2 \h^{\top} \x_l \right)
\label{eq:fsim}
\end{align}
where $\cL = \L_l + \H^{\top} \L_r \H$ and $\h^{\top} = \e^{\top} \L_r \H$. 

We summarize AGD in Algorithm \ref{alg:AGM}.
Suppose objective $f(\x_l)$ is \textit{$\beta$-smooth} (\ie,  $\beta$-Lipschitz)\footnote{Though log functions are used in \eqref{eq:fsim}, each argument is an approximation of Gaussian density integral over a quantization bin in \eqref{eq:multiInt}, and thus is sufficiently larger than $0$. 
Hence, the log function slopes are upper-bounded.}, with gradient denoted by $\nabla f(\x_l)$.
From \eqref{eq:fsim}, $\nabla f(\x_l)$ is

\vspace{-0.1in}
\begin{footnotesize}
\begin{align}
\nabla f(\x_l) =&-\sum_{i=1}^N  \frac{\tilde{\a}_{l,i}}{\tilde{\a}_{l,i}^{\top} \x_{l} + \tilde{b}_{l,i}}-\sum_{j=1}^M \frac{\H^{\top} \bar{\a}_{r,j}}{\bar{\a}_{r,j}^{\top} \H \x_{l} + \bar{b}_{r,j}} + \frac{2}{\sigma_p^2} \left( \cL \x_l + \h \right). 
\label{eq:gradient}
\end{align}
\end{footnotesize}\noindent

At each iteration $t$, AGD first takes a greedy step in the negative gradient direction $-\nabla f(\x_l^t)$ from previous solution $\x_l^t$ to $\c_l^{t+1}$ with step size $1/\beta$.
Then, new solution $\x_l^{t+1}$ is a convex combination of $\c_l^{t+1}$ and previously computed $\c_l^{t}$ in iteration $t-1$, given parameter $\gamma^t$ determined by AGD.

The only remaining question is how to best estimate smoothness $\beta$ of $f(\x_l)$ for step size $1/\beta$. 
As discussed in \cite{bubeck2014convex}, $\beta$ is the upper bound of the largest eigenvalue of the Hessian matrix $\bsPsi$ of $f(\x_l)$. 
Thus, we first write the Hessian matrix $\nabla^2 f(\x_l)$ of $f(\x_l)$ as

\vspace{-0.05in}
\begin{small}
\begin{align}
\nabla^2 f(\x_l) =& \sum_{i=1}^N \frac{\tilde{\a}_{l,i} \tilde{\a}_{l,i}^{\top}}{(\tilde{\a}_{l,i}^{\top} \x_{l} + \tilde{b}_{l,i})^2} + \sum_{j=1}^M \frac{\H^{\top} \bar{\a}_{r,j} \bar{\a}_{r,j}^{\top} \H}{(\bar{\a}_{r,j}^{\top} \H \x_{l} + \bar{b}_{r,j})^2} + \frac{2}{\sigma_p^2} \cL. 
\label{eq:hessian}
\end{align}
\end{small}\noindent 


\renewcommand{\algorithmicrequire}{\textbf{Input:}} 
\renewcommand{\algorithmicensure}{\textbf{Output:}} 
\begin{algorithm}[t]
\caption{Accelerated Gradient Descent} 
\label{alg:AGM} 
\begin{algorithmic}[1] 
\Require Convergence parameter $\epsilon$, smooth parameter $\beta$. 
\State Initialize $\c_l^1 \leftarrow \x_l^1$, $t \leftarrow 1$ and $\eta^0 \leftarrow 0$;
\State \textbf{while} $\| \nabla f(\x_l^t) \|^2 \geq\epsilon$ \textbf{do}
\State \qquad \ \ $\c_l^{t+1} \leftarrow \x_l^t-\frac{1}{\beta} \nabla f(\x_l^t)$,
\State \qquad \ \ $\eta^t \leftarrow \frac{1+\sqrt{1+4(\eta^{t-1}})^2}{2} $
\State \qquad \ \ $\gamma^t \leftarrow \frac{1-\eta^t}{\eta^{t+1}} $
\State \qquad \ \ $\x_l^{t+1} \leftarrow \left(1-\gamma^{t} \right) \c_l^{t+1} + \gamma^{t} \c_l^t$
\State \qquad \ \ $t \leftarrow t+1$.
\State \textbf{endwhile}
\Ensure $\x_l^{t+1}$.
\end{algorithmic} 
\end{algorithm}

Recall that $\tilde{\a}_{l,i}^{\top} \x_l + \tilde{b}_{l,i}, \forall i$ are linear functions in \eqref{eq:multiInt} that approximate Gaussian integrals over quantization bins. 
In contrast, matrix $\tilde{\a}_{l,i} \tilde{\a}_{l,i}^{\top}$ in the numerator of the first summation has a single non-zero entry proportional to $(z^+_{l,i} - z^-_{l,i})^2$, which is small when compared to $( \tilde{a}^{\top}_{l,i} \x_l + \tilde{b}_{l,i})^2$. 
(A similar argument can be made for matrix $\tilde{\a}_{r,j} \tilde{\a}_{r,j}^{\top}$ of the second summation.) 
Hence, the two summations in \eqref{eq:hessian} are small compared to $(2/\sigma_p^2) \cL$, and we can approximate $\nabla^2 f(\x_l) \approx (2/\sigma_p^2) \cL$. 


Instead of using computation-expensive eigen-decomposition, we consider \textit{Gershgorin Circle Theorem} (GCT) \cite{horn2012matrix} to compute an upper bound of the largest eigenvalue $\lambda_{\max}$ of $\cL$. 
By GCT, each eigenvalue $\lambda$ of $\bsPsi$ resides in at least one \textit{Gershgorin disc} corresponding to row $i$ of $\cL$, with center $o_i=\cL_{i,i}$ and radius $r_i=\sum_{j \neq i} |\cL_{i,j}|$. 
Thus, $\lambda_{\max}$ must satisfy
\begin{align}
\lambda_{\max} \leq & \max_i \; (o_i+r_i)   \nonumber\\
= & \max_i \; \left(\cL_{i,i} + \sum_{j \neq i} |\cL_{i,j}|\right) \triangleq \lambda^+_{\max} (\cL) .
\label{eq:maxL}
\end{align}
$\lambda^+_{\max}(\cL)$ can be computed efficiently given $\cL$ is a sparse matrix.
Thus, we can finally compute $\tilde{\beta} \triangleq (2/\sigma_p^2) \lambda^+_{\max}(\cL)$.

\subsection{Computation Complexity Analysis}
We analyze the computation complexity of our algorithm.
Assuming metric matrix $\M$ is optimized via feature graph learning (see Section\;\ref{sec:learn} for details), we compute sparse symmetric graph Laplacian matrices $\L_l \in \mathbb{R}^{N \times N}$ and $\L_r \in \mathbb{R}^{M \times M}$, where $M \leq N$, with non-zero entries $(i,j)$ iff $i-2 \leq j \leq i+2$.
From Section\;\ref{subsec:V2VMapping}, we know $\H \in \mathbb{R}^{M \times N}$ is also sparse, with non-zero entries $(j,i)$ iff $s(i,x_{l,i}) \in \cN_j$. 
Thus, assuming $h \geq 2$, matrix product $\L_r \H$ has non-zero entries $(i,j)$ iff $s(i,x_{l,i}) \in \bigcup_{k=j-2}^{j+2} \cN_{k} = \cN_j^+$.
Since each $i$ of $N$ left pixels is mapped to $\cO(1)$ neighborhoods $\cH_j^+$ of right pixels $j$, $\L_r \H$ is sparse and has $\cO(N)$ non-zero entries.
Similarly, we can conclude $\H^\top \L_r \H$ is also sparse and contains $\cO(N)$ non-zero entries.

Thus, $\cL  = \L_l + \H^{\top} \L_r \H$ is also sparse and contains $\cO(N)$ non-zero entries, computed in $\cO(N)$.
Similarly, $\h^{\top} = \e^{\top} \L_r \H$ contains $\cO(N)$ non-zero entries.
To compute $\lambda^+_{\max}(\cL)$ in \eqref{eq:maxL}, we compute all candidates $i$, which requires accessing each non-zero entry in $\cL$ exactly once, and thus complexity is $\cO(N)$. 
Thus, computing $\tilde{\beta}$ is also $\cO(N)$.

Given $\cL$, $\h$ and $\tilde{\beta}$, we execute AGD iteratively. 
In each iteration, we calculate gradient $\nabla f(\x_l)$ in \eqref{eq:gradient}.
Recalling that $\bar{\a}_{r,j}$ has only one single non-zero entry and $\H$ is sparse, the cost of computing $\H^{\top} \bar{\a}_{r,j}$ and $\bar{\a}_{r,j}^{\top} \H \x_{l}$ in the second summation can be omitted. 
Computing $\cL \x_{l}$ has complexity $\cO(N)$ since $\cL$ has $\cO(N)$ non-zero entries. 
Thus, computing $\nabla f(\x_l)$ has complexity $\cO(N+M+N) = \cO(N)$. 
Since the number of iterations of AGD is $\cO(1/\sqrt\epsilon)$ \cite{bubeck2014convex}, executing AGD has $\cO \left(\frac{N}{\sqrt\epsilon} \right)$ computation.
To summarize, combining the computation cost of $\cL$, $\h$, $\beta$ and AGD, we get $\cO \left(N + N + N + \frac{N}{\sqrt\epsilon} \right) = \cO(\frac{N}{\sqrt{\epsilon}})$, which is linear time.
Linear-time complexity essentially means accessing each datum once, which constitutes a complexity lower bound for serial data processing.
Thus, our depth enhancement algorithm is practical and can potentially be implemented in real-time.

\section{Feature Graph Learning}
\label{sec:learn}
\subsection{Learning Metric for Graph Construction}

When pixel row $i$ of the left view is optimized, we assume that the previous $K$ rows, $i-1, \ldots i-K$, have already been enhanced into $\widetilde{\x}_l^{i-1}, \ldots \widetilde{\x}_l^{i-K}$.
Using these $K$ enhanced rows, we compute graph Laplacian $\L_l$ to define prior $\rPr(\x)$ in \eqref{eq:prior}.
Given $K < N$ in practice, estimating $\L_l \in \mathbb{R}^{N \times N}$ reliably using only $K$ signal observations is difficult.
In particular, statistical graph learning algorithms such as \textit{graphical lasso} (GLASSO) \cite{friedman08} 
that compute a sparse precision matrix using as input a reliable empirical covariance matrix estimated from many observations do not work in our scenario.

Instead, we construct an appropriate similarity graph via \textit{metric learning} \cite{hu2020feature}.
We first assume that associated with each pixel (graph node) $i$ in $\x_l$ is a length-$F$ relevant \textit{feature vector} $\f_i \in \mathbb{R}^F$ (to be discussed). 
The \textit{feature distance} $d_{ij}$ between two nodes $i$ and $j$ is computed using a real, symmetric and PD \textit{metric matrix} $\M \in \mathbb{R}^{F \times F}$ as
\begin{align}
d_{ij} = (\f_i-\f_j)^{\top} \M (\f_i-\f_j).
\label{eq:featureDist}
\end{align}
\eqref{eq:featureDist} is also called the \textit{Mahalanobis distance} in the machine learning literature \cite{mahalanobis1936generalized}.
Since $\M$ is PD, $d_{ij} > 0$ for $\f_i - \f_j \neq \0$. The edge weight $u_{ij}$ between nodes $i$ and $j$ is then computed using a Gaussian kernel:
\vspace{-0.01in}
\begin{align}
u_{ij} = \exp \left( -d_{ij} \right) .
\label{eq:edgeWeight}
\end{align}
Note that, to reduce computation complexity, we construct a sparse graph where each pixel $i$ is only connected to its four closest neighbors $i\pm 1$ and $i\pm 2$. 
Thus, Laplacian $\L$ is sparse with $\cO(1)$ non-zero entries per row / column.

To optimize $\M$, we minimize the \textit{graph Laplacian regularizer} (GLR) evaluated using $K$ previous pixel rows, \ie, 
\begin{align} 
\min_{\M \succ 0} ~~ & \sum_{k=1}^K 
\left( \widetilde{\x}_l^k \right)^{\top} \L^k_l(\M) \widetilde{\x}_l^k 
\label{eq:metricLearn} \\
&= \sum_{k=1}^K \sum_{i,j} u^k_{ij} \left( \widetilde{x}_{l,i}^k - \widetilde{x}_{l,j}^k \right)^2
\end{align}
where edge weights $u_{ij}^k$ in Laplacian $\L^k_l(\M)$ is computed using features $\f_i^k$ and $\f_j^k$ of the $k$-th observation $\widetilde{\x}_l^k$ via \eqref{eq:featureDist} and \eqref{eq:edgeWeight}.
To optimize $\M$ in \eqref{eq:metricLearn}, \cite{hu2020feature} proposed a fast algorithm to optimize the diagonal and off-diagonal entries of $\M$ alternately. 
See \cite{hu2020feature} for details.

\subsection{Feature Selection for Metric Learning}

To construct a feature vector $\f_i$ for each pixel $i$ in $\x_l$, we first compute the pixel's corresponding \textit{surface normal} $\n_i \in \mathbb{R}^3$ by projecting it to 3D space and computing it using its neighboring points via a method in \cite{avron2010l1}.
Then, together with depth value $x_i$ and location $\l_i \in \mathbb{R}^2$ in the 2D grid, we construct $\f_i \in \mathbb{R}^6$.
Because $\M$ is symmetric, the number of matrix entries we need to estimate is only $21$.

\section{Experimentation}
\label{sec:results}
\subsection{Experimental Setup}

We conducted simulations with three types of datasets:

\begin{enumerate}[(i)]

\item Public synthetic dataset: in\cite{mayer2016large}, three synthetic datasets, \texttt{FlyingThings}, \texttt{Monkaa} and \texttt{Driving}, were introduced. They are generated using Blender\footnote{https://www.blender.org/}, and the ground-truth depth images of both views are provided.
As a pre-processing, we projected the depth images to 3D space, leading to PCs with around 1 million points;
\item Public real dataset: the Middlebury dataset is collected with structured-light sensor \cite{scharstein2014high}, which also provides the ground-truth depth images of both views.
We adopted five representative cases for our test, {\it i.e.}, \texttt{Adir}, \texttt{Playtable}, \texttt{Recycle}, \texttt{Teddy}, and \texttt{ArtL}.
After projecting the depth images into 3D space, the PCs of \texttt{Adir}, \texttt{Playtable} and \texttt{Recycle} have roughly 60K points, while \texttt{Teddy} and \texttt{ArtL} have roughly 32K and 17K points, respectively;
\item In-house real dataset: we collected three pairs of raw depth images using the \textit{Intel RealSense\textsuperscript{TM} D435} for three different scenes, \texttt{Scene0}, \texttt{Scene1} and \texttt{Scene2}. The associated projected PCs have roughly 17k points. This dataset has no ground-truths.
\end{enumerate}
For the datasets (i) and (ii), we first added signal-dependent \textit{Gaussian} noise (SDGN) to both views according to \eqref{eq:sigDependent}, then they were quantized using \eqref{eq:nonUniform}.
In addition, instead of Gaussian noise, we added signal-dependent \textit{Laplacian} noise (SDLN) \cite{dinesh2020point} to both views according to \eqref{eq:sigDependent} in order to evaluate the robustness of our approach.
For dataset (iii), the collected data was already corrupted by real noise.

When learning the metric for graph construction, we considered the previous $K=30$ pixel rows. 
To reduce computation complexity, the same optimized $\M$ was used for the right view when enhancing the left view, and vice versa. 
Given feature vector in $\y^i$ of the current row $i$, we computed the corresponding Laplacian $\L^i(\M)$.  

We compared our 3D PC enhancement method against three model-based image denoising schemes, BF \cite{tomasi98}, BM3D \cite{dabov2007image} and the early version of this work, SINUQ --- short for Signal-Independent Noise Uniform Quantization \cite{zhang20203d}.
Similar to our workflow, we applied these three schemes on the noise-corrupted and quantized depth image pairs before the PC synthesis steps.
We also compared our work with six representative PC denoising algorithms, APSS \cite{guennebaud2007algebraic}, RIMLS \cite{oztireli2009feature}, MRPCA \cite{mattei2017point}, GLR \cite{zeng20193d}, PCN \cite{rakotosaona2020pointcleannet} and DMR \cite{luo2020differentiable}.
This set of methods was applied to the PCs projected from the corrupted left and right views.
Among these six methods, PCN and DMR are based on deep learning, where we directly applied their released trained models to our test datasets.  

To make different PCs comparable, we re-centered each PC to the origin then scaled it inside the unit sphere. 
For datasets (i) and (ii) with ground-truth PCs, we adopted two commonly used PC evaluation metrics for objective evaluation, {\it i.e.}, the parameter-free point-to-point (C2C) error \cite{girardeau2005change} and point-to-plane (C2P) error \cite{tian2017geometric}, where plane normal in C2P was computed using six neighboring points (the only parameter). 
For dataset (iii) without ground-truths, we employed a no-reference metric (\ie, Pseudo MOS) using sparse convolutional neural network, designed specifically for quality assessment of 3D PCs \cite{liu2021LSPCQA}.
We include also visual comparisons.

\subsection{Experimental Results}

\subsubsection{Linear Approximation of the Gaussian pdf}

\begin{figure}
\begin{center}
\begin{tabular}{cc}
\includegraphics[width=1.56in]{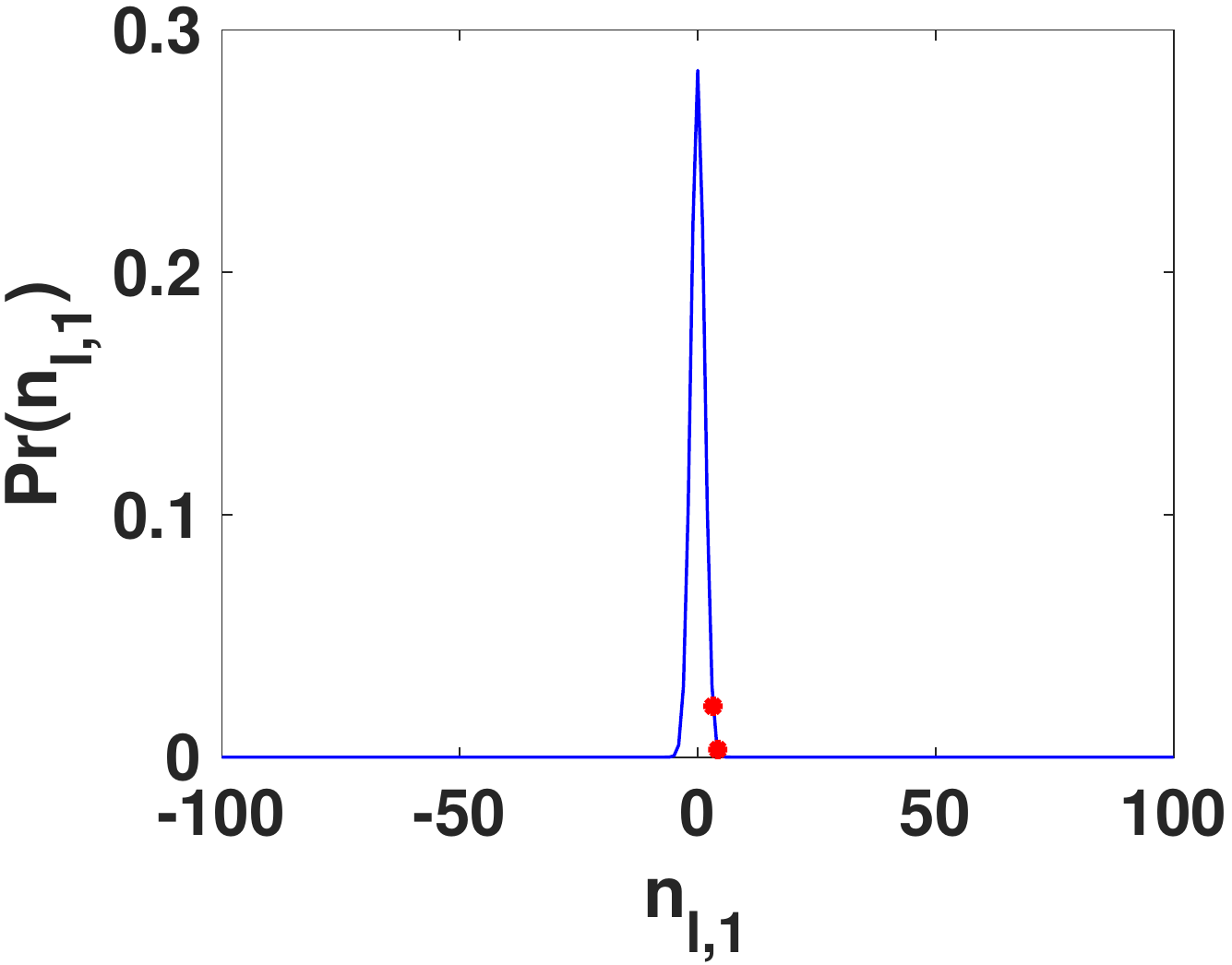}&\includegraphics[width=1.6in]{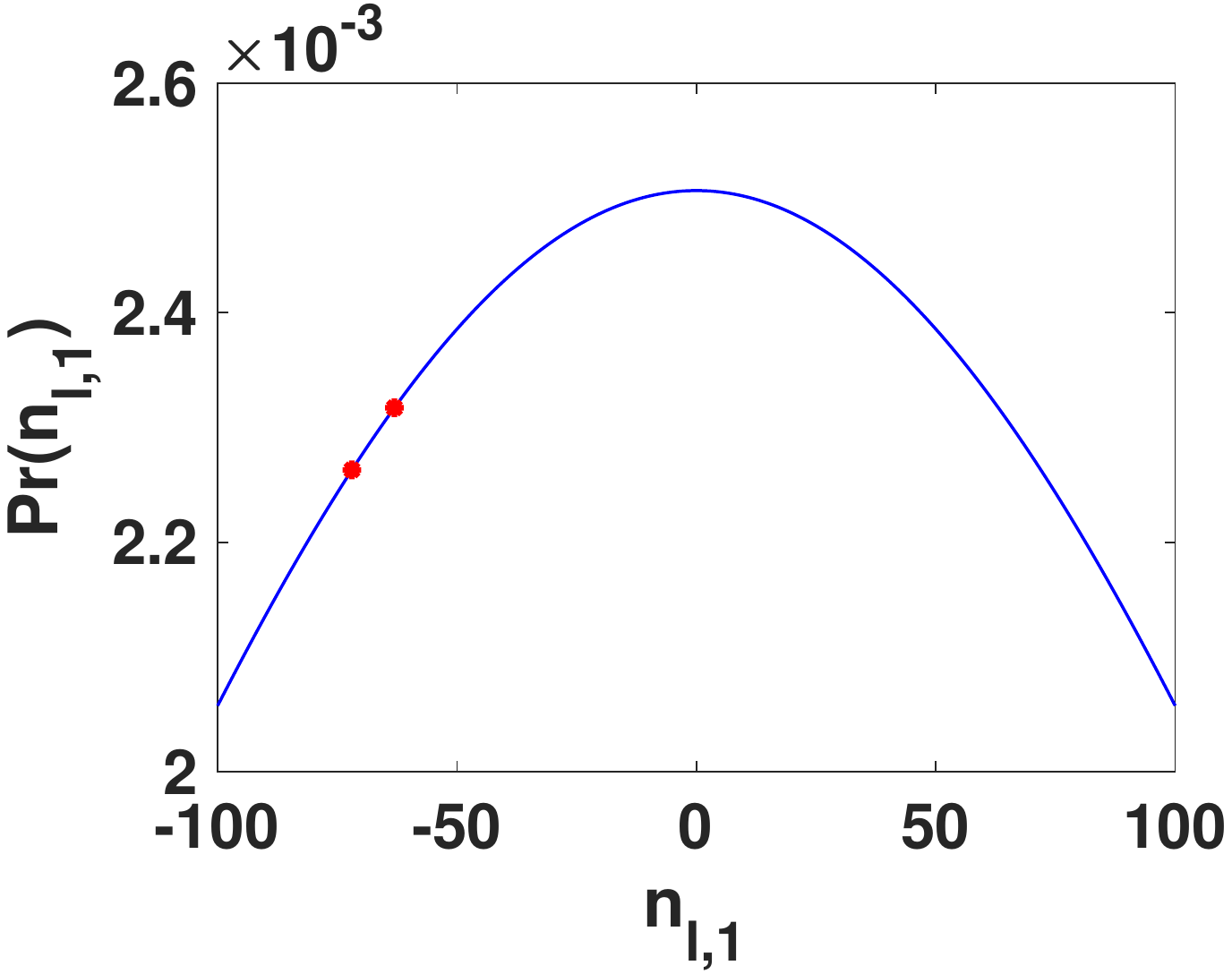}\\
\parbox{.45\linewidth}{\centering\small (a)} & \parbox{.45\linewidth}{\centering\small (b)}\\
\end{tabular}
\end{center}
\vspace{-0.1in}
\caption{Gaussian pdf of (a) $x_{l,1}=0.5$m and (b) $x_{l,1}=4.5$m, where the red dots are pdf of the lower and upper bounds $n^-(y_{l,1},x_{l,1})$ and $n^+(y_{l,1},x_{l,1})$ in the region of integration $\cR_{l,1}$, respectively, which is an empirical demonstration corresponding to Fig.\;\ref{fig:likelihood}. }
\label{fig:linApprox}
\end{figure}

To verify the accuracy of the linear approximation \eqref{eq:linearApprox} of the Gaussian pdf, we constructed two extreme cases, as shown in Fig.\;\ref{fig:linApprox}. 
We first consider a close depth pixel $x_{l,1}=0.5$m and with a small quantization bin, which results in a very narrow region of integration $\cR_1$, as shown in Fig.\;\ref{fig:linApprox}(a). 
We observe that the pdf $\rPr(n_{l,1})$ within $\cR_1$ is roughly linear.
Specifically, the likelihood term $\rPr(y_{l,1}|x_{l,1})$ in \eqref{eq:likelihood} is $0.0098$, and its linear approximation in \eqref{eq:linearApprox} is $0.0083$.
In contrast, for a far depth pixel $x_{l,1}=4.5$m, and a large noise variance $\sigma_{l,1}^2$, the pdf $\rPr(n_{l,1})$ becomes relatively flat, see Fig.\;\ref{fig:linApprox}(b). 
We observe that $\rPr(n_{l,1})$ within $\cR_1$ is roughly linear as well. 
In this case, the likelihood term $\rPr(y_{l,1}|x_{l,1})$ in \eqref{eq:likelihood} is $0.0209$, and its linear approximation in \eqref{eq:linearApprox} is $0.0209$.
Thus, we can conclude that, in both cases, the linear approximation was sufficiently accurate.

\subsubsection{Comparison for Public Dataset}
\begin{table}
	\centering
	\caption{\footnotesize C2C$(\times10^{-3})$ and C2P$(\times10^{-5})$ results of competing methods for the PCs of the public synthetic dataset corrupted by SDGN.} \label{tab:Syn}
	\scriptsize
	\begin{tabular}{c||c|c|c|c}
	    \hline
		\tiny \diagbox{C2P}{C2C} & FlyingThings & Monkaa & Driving & Average\\
		 \hline
	    \hline
	     \multirow{2}{*}{BF\cite{tomasi98}} & 1.83 & 0.88 & 2.06 & 1.59   \\
		\cdashline{2-5} 
	        & 2.3 & 0.2 & 6.3 & 2.9\\
	    \hline 
	    \multirow{2}{*}{BM3D\cite{dabov2007image}} & 4.77 & 1.15 & 4.11 & 3.34 \\
		\cdashline{2-5} 
	        & 8.3 & 0.3 & 16.8 & 8.5  \\
	   \hline 
	   \multirow{2}{*}{SINUQ\cite{zhang20203d}} & 1.44 & 1.03 & 2.09 & 1.52 \\
		\cdashline{2-5} 
	        & 2.9 & 0.3 & 10.2 & 4.5  \\
	    \hline
	    \hline
		 \multirow{2}{*}{APSS\cite{guennebaud2007algebraic}} & 2.67 & 1.00 & 2.94 & 2.20  \\
		\cdashline{2-5} 
	        & 2.1 & 0.2 & 5.9 & 2.7 \\
	   \hline 
	    	\multirow{2}{*}{RIMLS\cite{oztireli2009feature}} & 2.65 & 0.99 & 2.95 & 2.20 \\
	   \cdashline{2-5} 
	        & 2.1  & 0.2 & 5.9 & 2.7\\
	    \hline 
	     	\multirow{2}{*}{MRPCA\cite{mattei2017point}} & 2.69 & 0.95 & 2.93 & 2.19  \\
	   \cdashline{2-5} 
	      & 2.1 & 0.2 & 5.2 & 2.5  \\
	     \hline
	     	\multirow{2}{*}{GLR\cite{zeng20193d}} & 2.77 & 1.03 & 3.02 & 2.27   \\
	   \cdashline{2-5} 
	      & 2.2 & 0.2 & 5.5 & 2.6  \\
	   \hline 
	     	\multirow{2}{*}{PCN\cite{rakotosaona2020pointcleannet}} & 2.13 & 0.98 & 1.62 & 1.58\\
	   \cdashline{2-5} 
	      & 1.1 & 0.2 & 2.2 & 1.2 \\
	    \hline
	     	\multirow{2}{*}{DMR\cite{luo2020differentiable}} & 2.70 & 0.96 & 2.98 & 2.21 \\
	   \cdashline{2-5} 
	      & 2.0 & 0.2  & 4.9  & 2.4 \\
	     \hline
	    \hline
	    \multirow{2}{*}{Proposed} & \textbf{1.12}  & \textbf{0.59} & \textbf{1.22} & \textbf{0.98} \\
	    \cdashline{2-5} 
	     & \textbf{0.5} & \textbf{0.1} & \textbf{1.1} & \textbf{0.5} \\
        \hline
    	\end{tabular}
    	\vspace{-0.4cm}
\end{table}

\begin{table}
	\centering
	\caption{\footnotesize C2C$(\times10^{-3})$ and C2P$(\times10^{-5})$ results of competing methods for the PCs of the public real dataset corrupted by SDGN.} \label{tab:Mid}
	\scriptsize
	\begin{tabular}{c||c|c|c|c|c|c}
		\hline
		 \tiny \diagbox{C2P}{C2C} & Adir & ArtL & Teddy & Recycle & Playtable & Average\\
		\hline
		\hline
	     \multirow{2}{*}{BF\cite{tomasi98}} & 4.88 & 7.76 & 4.81 & 2.75 & 3.11 &  4.66\\
		\cdashline{2-7} 
	        & 4.2 & 8.7 & 2.9 & 1.0 & 1.8 & 3.7\\
	    \hline 
	    \multirow{2}{*}{BM3D\cite{dabov2007image}} & 2.92 & 8.19 & 5.22 & 2.37 & 3.80 & 4.50 \\
		\cdashline{2-7} 
	        & 2.9 & 11.1 & 8.0 & 2.9  & 6.2 & 6.2\\
	   \hline 
	    \multirow{2}{*}{SINUQ\cite{zhang20203d}} & 3.12 & 7.96 & 4.16 & 2.68 & 3.44 & 4.27 \\
		\cdashline{2-7} 
	        & 2.4 & 10.8 & 2.4 & 1.3  & 2.5 & 3.9\\
	    \hline
	    \hline
		 \multirow{2}{*}{APSS\cite{guennebaud2007algebraic}} & 10.64 & 12.05 & 9.13 & 6.91 & 6.23 & 8.99 \\
		\cdashline{2-7} 
	        & 13.8 & 12.5 & 7.8 & 6.3  & 5.8 & 9.2\\
	   \hline 
	    	\multirow{2}{*}{RIMLS\cite{oztireli2009feature}} & 10.41 & 11.74 & 8.86 & 6.69 & 5.98 & 8.74\\
	   \cdashline{2-7} 
	        & 13.7  & 12.3 & 7.0 & 6.1 & 5.6 & 8.9\\
	    \hline 
	     	\multirow{2}{*}{MRPCA\cite{mattei2017point}} & 10.18 & 11.26 & 8.56 & 6.30 & 5.48 & 8.36 \\
	   \cdashline{2-7} 
	      & 15.9 & 13.5 & 8.8 & 5.2 & 4.4 & 9.6\\
	     \hline 
	     	\multirow{2}{*}{GLR\cite{zeng20193d}} & 10.02 & 11.11 & 8.39 & 6.03 & 5.00 & 8.11\\
	   \cdashline{2-7} 
	      & 11.9 & 9.5 & 6.7 & 4.6 & 3.7 & 7.3 \\
	    \hline
	     	\multirow{2}{*}{PCN\cite{rakotosaona2020pointcleannet}} & 7.03 & 9.03 & 5.51 & 4.00 & 4.23 &  5.96 \\
	   \cdashline{2-7} 
	      & 5.5 & 6.8 & 2.7 & 1.8 & 2.4 & 3.8 \\
	    \hline
	     	\multirow{2}{*}{DMR\cite{luo2020differentiable}} & 9.60 & 9.90 & 7.86 & 5.63 & 4.26 &  7.45\\
	   \cdashline{2-7} 
	      & 11.1 & 8.0 & 5.7 & 3.9 & 2.9  &  6.3\\
	    \hline
	    \hline
	    \multirow{2}{*}{Proposed} & \textbf{2.60}  & \textbf{4.11} & \textbf{3.71} & \textbf{2.18} &  \textbf{2.58} & \textbf{3.04} \\
	    \cdashline{2-7} 
	     & \textbf{0.9} & \textbf{1.7} & \textbf{1.4} & \textbf{0.7} & \textbf{1.2} & \textbf{1.2} \\
       \hline
    	\end{tabular}
    	\vspace{-0.2cm}
\end{table}

\begin{table}
	\centering
	\caption{\footnotesize C2C$(\times10^{-3})$ and C2P$(\times10^{-5})$ results of the PCs of the public synthetic dataset and real dataset corrupted by SDLN.} \label{tab:lap}
	\scriptsize
	\begin{tabular}{c||c|c|c|c|c|c}
		\hline
		 \tiny \diagbox{C2P}{C2C} & FlyingThings & Driving &  Adir & ArtL & Teddy & Average\\
		\hline
		\hline
	     \multirow{2}{*}{BF\cite{tomasi98}} & 1.83 & 2.03 & 4.12 & 7.71 & 4.19 &  3.98\\
		\cdashline{2-7} 
	        & 2.3 & 6.0 & 2.8 & 8.7 & 2.3 & 4.4\\
	   \hline 
	    \multirow{2}{*}{BM3D\cite{dabov2007image}} & 4.78 & 4.10 & 2.92 & 8.20 & 5.23 & 5.05 \\
		\cdashline{2-7} 
	        & 8.4 & 16.9 & 3.0 & 11.0  & 8.6 & 9.6\\
	   \hline
	    \multirow{2}{*}{SINUQ\cite{zhang20203d}} & 1.37 & 2.01 & 3.12 & 6.63 & 4.15 & 3.46 \\
		\cdashline{2-7} 
	        & 2.3 & 9.4 & 2.4 & 9.2  & 2.3 & 5.1\\
	    \hline
	    \hline
		 \multirow{2}{*}{APSS\cite{guennebaud2007algebraic}} & 2.44 & 2.70 & 9.44 & 10.72 & 8.15 & 6.69 \\
		\cdashline{2-7} 
	        & 1.7 & 6.1 & 13.5 & 12.1  & 7.5 & 8.2\\
	    \hline
	    	\multirow{2}{*}{RIMLS\cite{oztireli2009feature}} & 2.39 & 2.62 & 9.19 & 10.39 & 7.89 & 6.50\\
	   \cdashline{2-7} 
	        & 1.7  & 5.8 & 13.4 & 11.8 & 7.2 & 7.9\\
	    \hline
	     	\multirow{2}{*}{MRPCA\cite{mattei2017point}} & 2.29 & 2.53 & 8.62 & 9.47 & 7.33 & 6.05 \\
	   \cdashline{2-7} 
	      & 1.8 & 5.1 & 13.8 & 10.9 & 7.6 & 7.8\\
	   \hline 
	     	\multirow{2}{*}{GLR\cite{zeng20193d}} & 2.32 & 2.45 & 8.67 & 9.61 & 7.42 & 6.09\\
	   \cdashline{2-7} 
	      & 1.8 & 5.2 & 11.3 & 8.7 & 6.2 & 6.6 \\
	    \hline
	     	\multirow{2}{*}{PCN\cite{rakotosaona2020pointcleannet}} & 1.86 & 1.26 & 5.57 & 7.45 & 4.71 &  4.17 \\
	   \cdashline{2-7} 
	      & 0.9 & 1.1 & 4.7 & 6.3 & 2.3 & 3.1 \\
	   \hline 
	     	\multirow{2}{*}{DMR\cite{luo2020differentiable}} & 2.23 & 2.82 & 7.31 & 7.22 & 6.39 &  5.19\\
	   \cdashline{2-7} 
	      & 1.6 & 5.0 & 9.5 & 5.8 & 5.0 &  5.4\\
	    \hline
	    \hline
	    \multirow{2}{*}{Proposed} & \textbf{1.12}  & \textbf{1.21} & \textbf{2.66} & \textbf{4.10} &  \textbf{3.67} & \textbf{2.55} \\
	    \cdashline{2-7} 
	     & \textbf{0.4} & \textbf{1.1} & \textbf{1.0} & \textbf{1.7} & \textbf{1.4} & \textbf{1.1} \\
       \hline
    	\end{tabular}
    	\vspace{-0.3cm}
\end{table}

In the case of the depth views corrupted by SDGN, quantitative results of different methods in terms of C2C and C2P errors are shown in Table~\ref{tab:Syn} and \ref{tab:Mid}, where the former one provides results for the PCs of the synthetic datasets, and the latter one is for the PCs of the Middlebury dataset.
Overall, our method achieved by far the best performance under both metrics for both datasets. 
For synthetic datasets in Table~\ref{tab:Syn}, our proposal outperformed the second-best algorithm SINUQ by $0.54$ in terms of C2C\,$(\times10^{-3})$ and
PCN by $0.7$ in terms of C2P\,$(\times10^{-5})$ on average, respectively.
For the Middlebury dataset in Table~\ref{tab:Mid}, in terms of C2C\,$(\times10^{-3})$, our method outperformed the second-best algorithm SINUQ by $1.23$, while in terms of C2P\,$(\times10^{-5})$, our method outperformed the second-best algorithm BF by $2.5$ on average.

The synthetic datasets had relatively small depth values, meaning that the signal-dependent additive noise was small. 
Thus, the performance of the six PC denoising algorithms were comparable to that of the three image denoising schemes. 
However, for the Middlebury dataset capturing real-world scenes, they had relatively large depth values, and thus the additive noise was large.

Table~\ref{tab:lap} shows C2C and C2P results of the PCs from both the public synthetic dataset and real dataset, corrupted by SDLN. 
Similarly, our proposal outperformed nine competitors in both metrics, with C2C\,$(\times10^{-3})$ error reduced by $0.91$ compared with the second-best algorithm SINUQ, and C2P\,$(\times10^{-5})$ error reduced by $2.0$ compared with the second-best algorithm PCN, respectively.

\begin{figure*}[htbp]
\begin{center}
\begin{tabular}{m{0.2cm}m{0.85in}m{0.85in}m{0.85in}m{0.85in}m{0.85in}m{0.85in}m{0.4in}}
\small (a) & \includegraphics[width=0.92in]{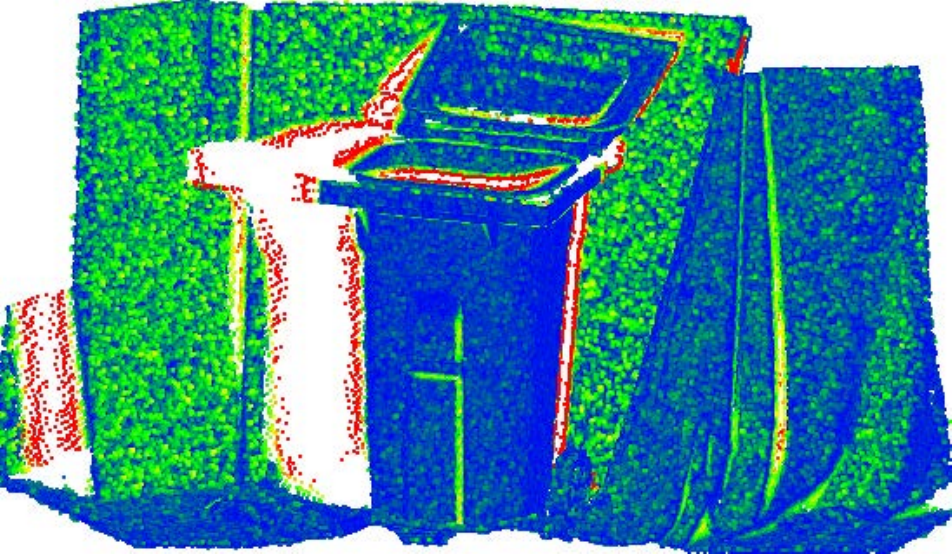}&\includegraphics[width=0.92in]{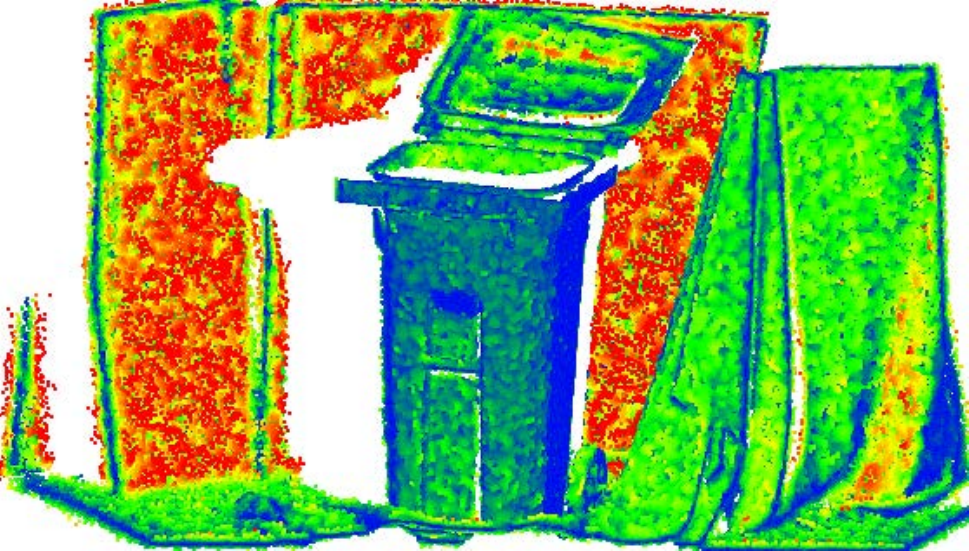}&\includegraphics[width=0.92in]{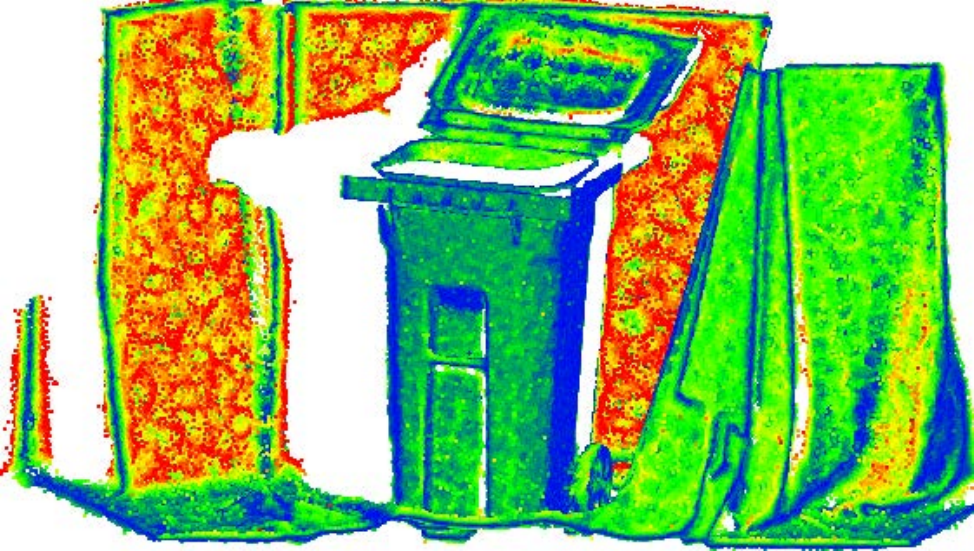}&\includegraphics[width=0.92in]{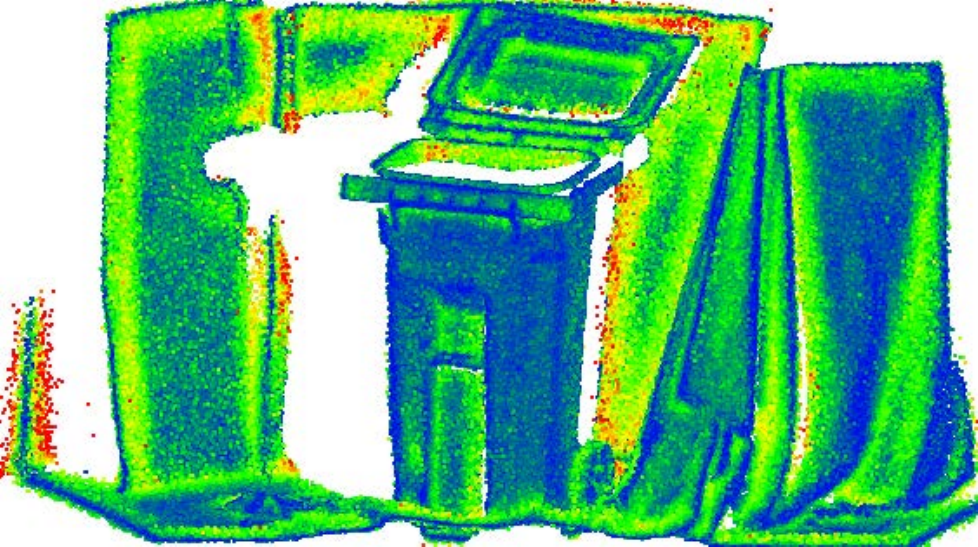}&\includegraphics[width=0.92in]{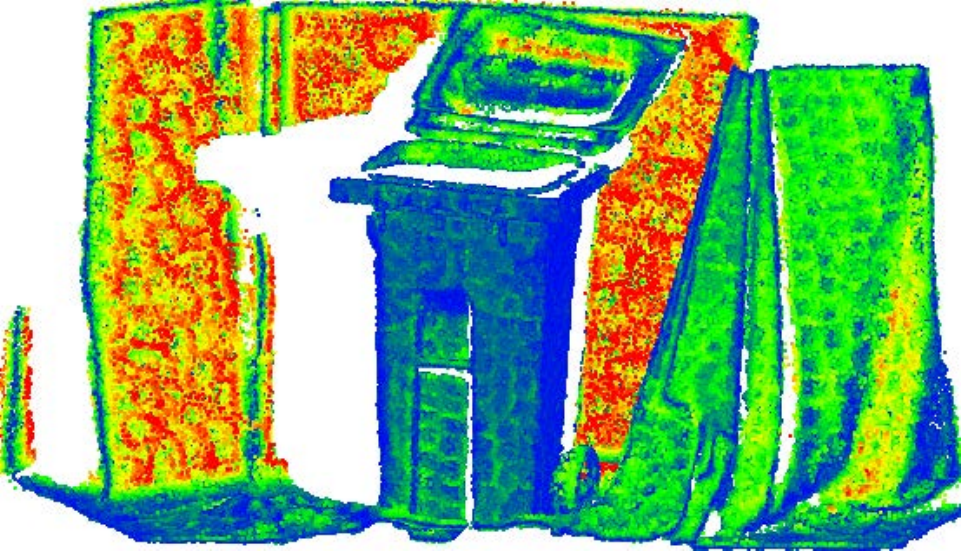}&\includegraphics[width=0.92in]{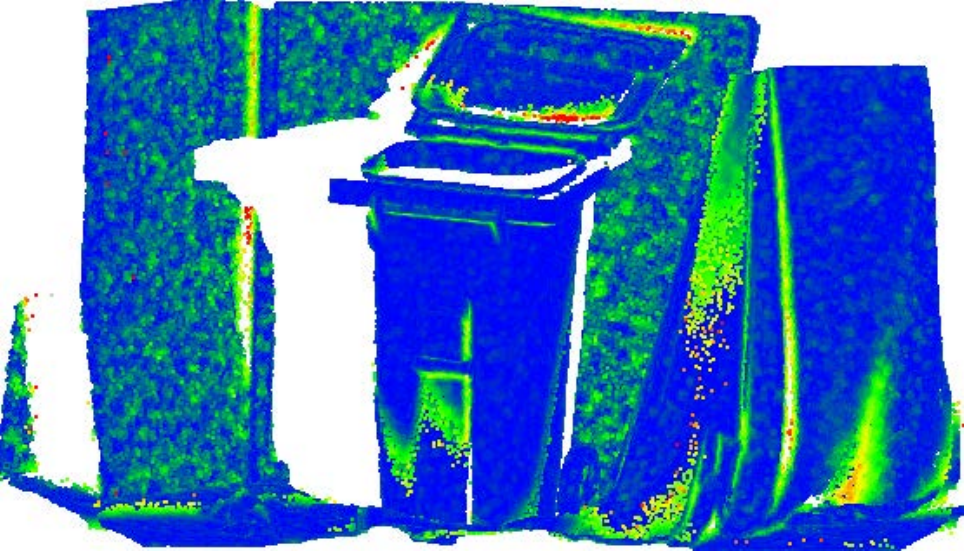} & \multirow{1}{*}{\includegraphics[width=0.55in]{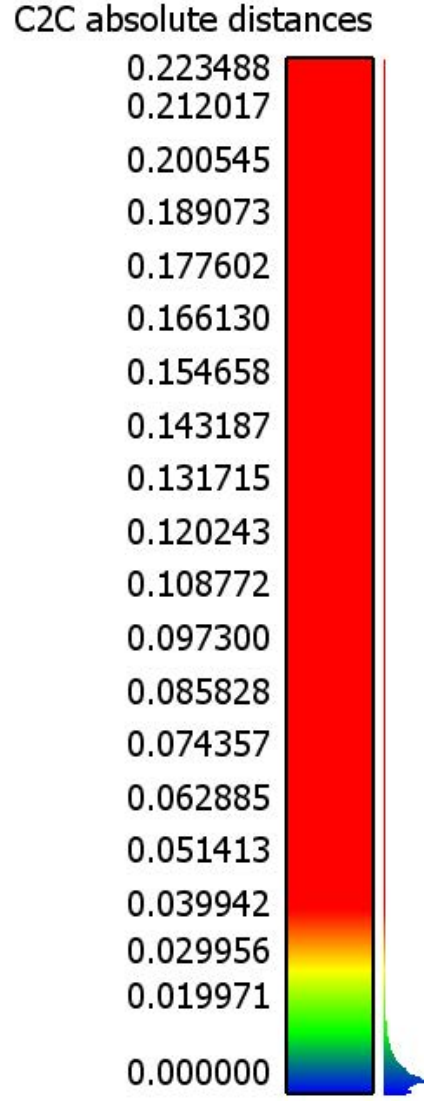}} 
\\ 
 \small (b)& \includegraphics[width=0.92in]{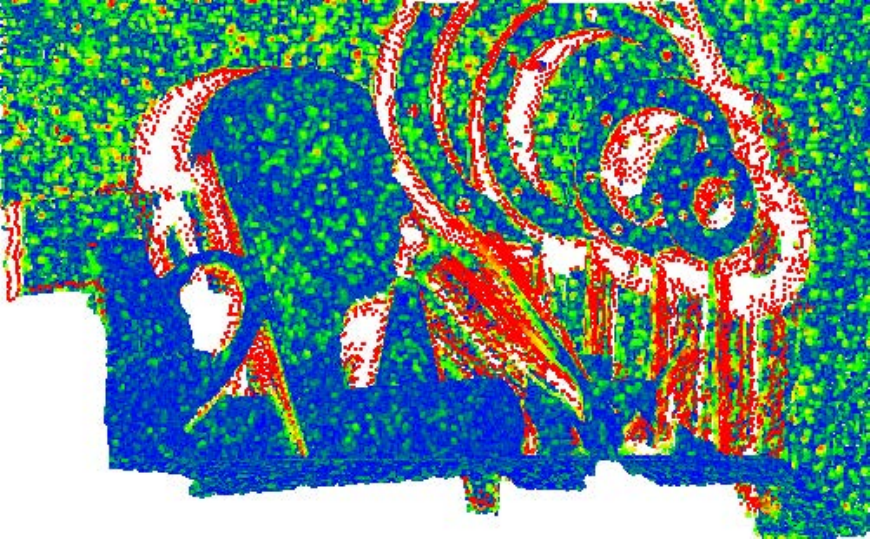} &\includegraphics[width=0.92in]{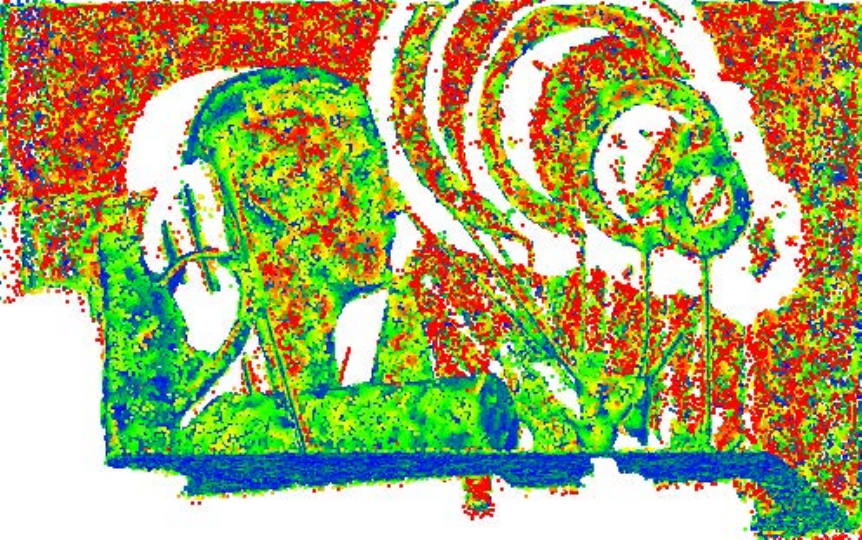}&\includegraphics[width=0.92in]{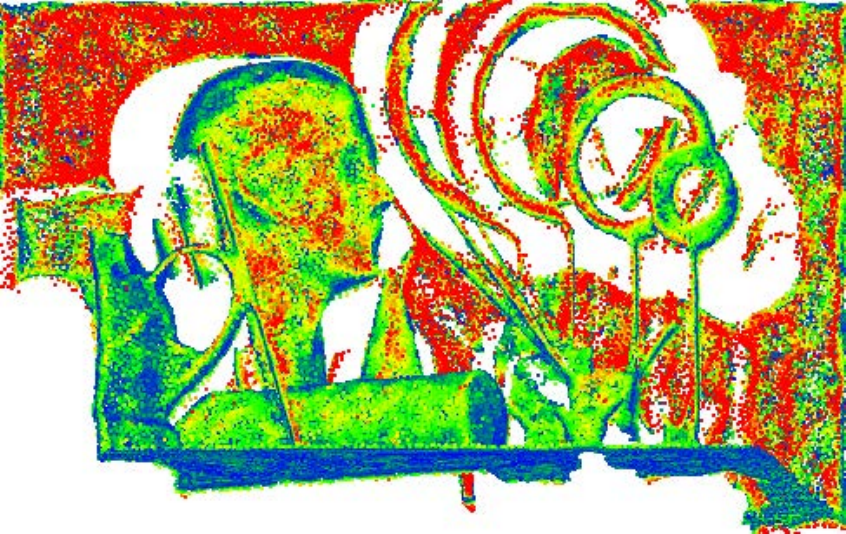}&\includegraphics[width=0.92in]{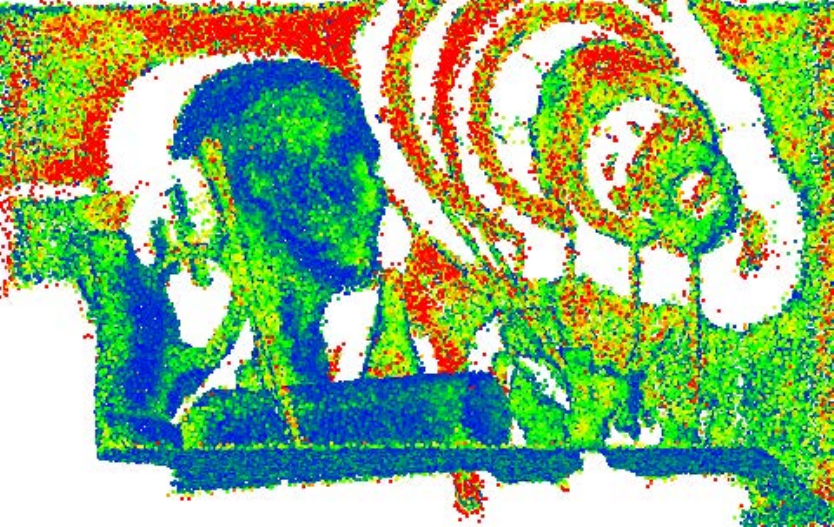}&\includegraphics[width=0.92in]{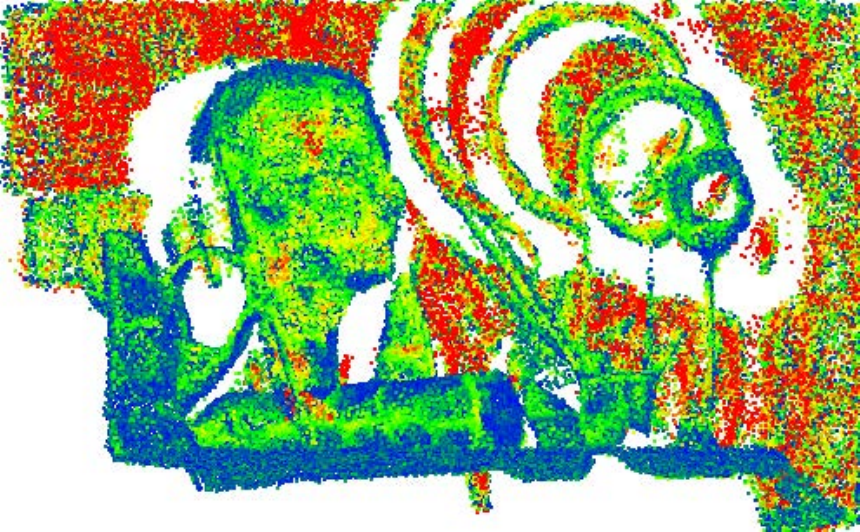}&\includegraphics[width=0.92in]{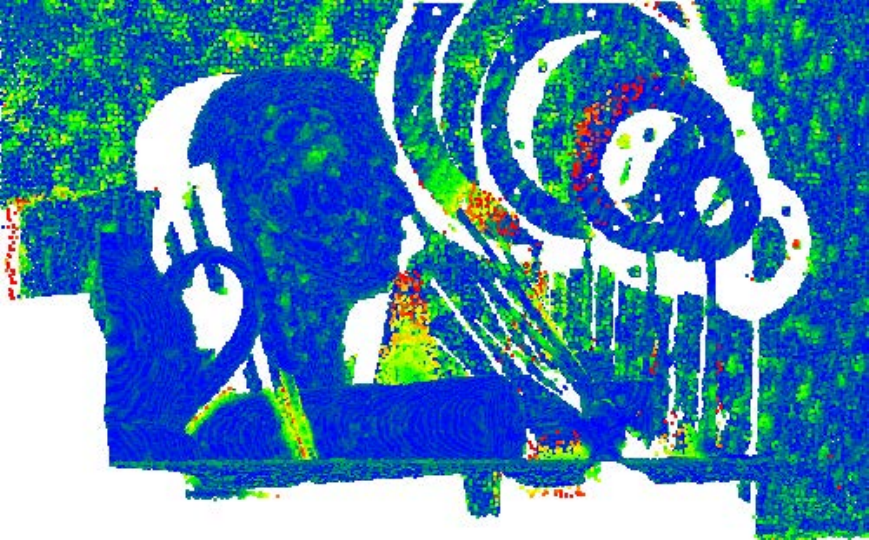} &  \\ 
 \small (c)& \includegraphics[width=0.92in]{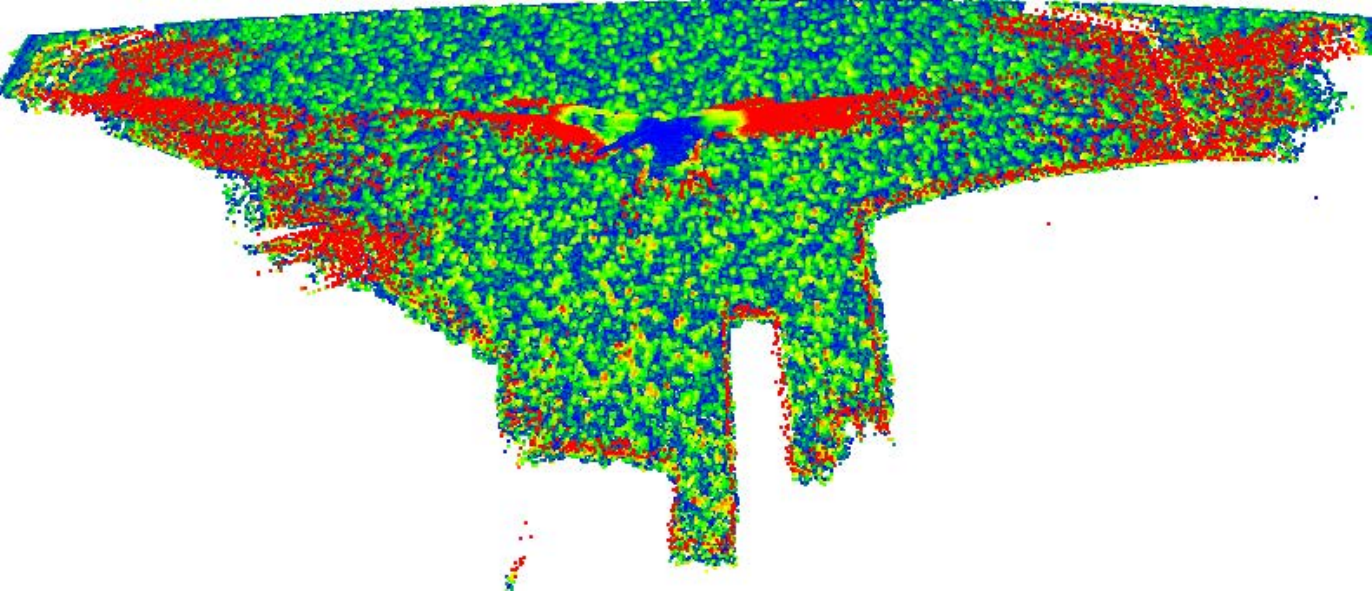} &\includegraphics[width=0.92in]{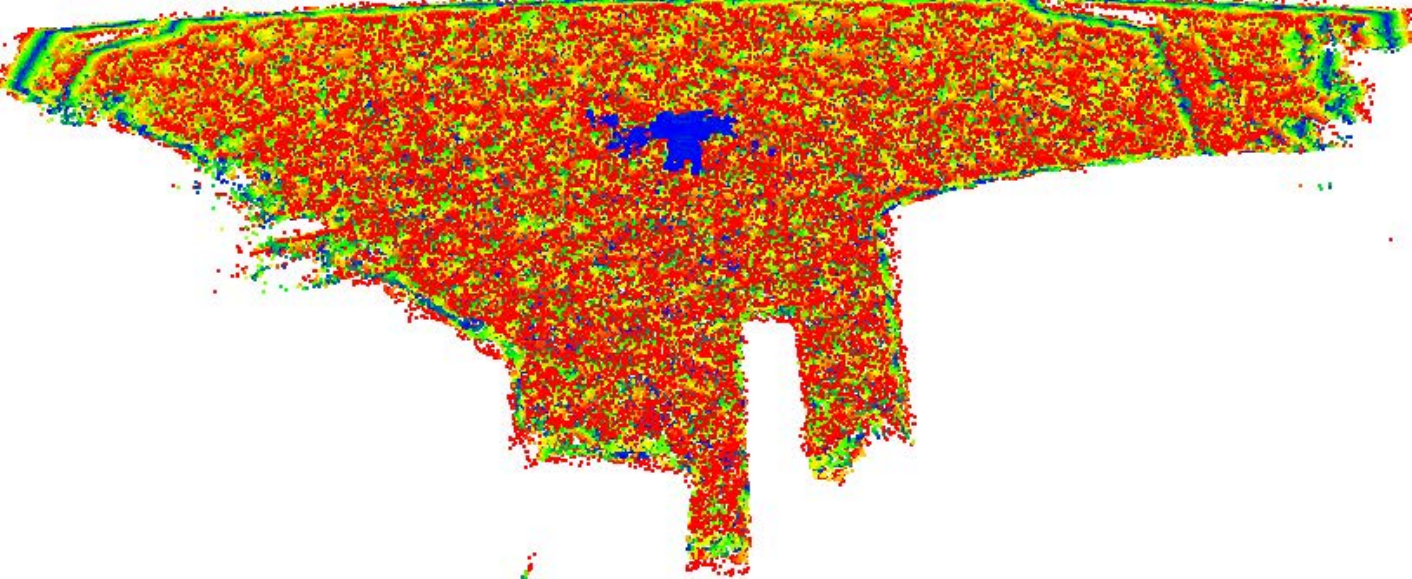}&\includegraphics[width=0.92in]{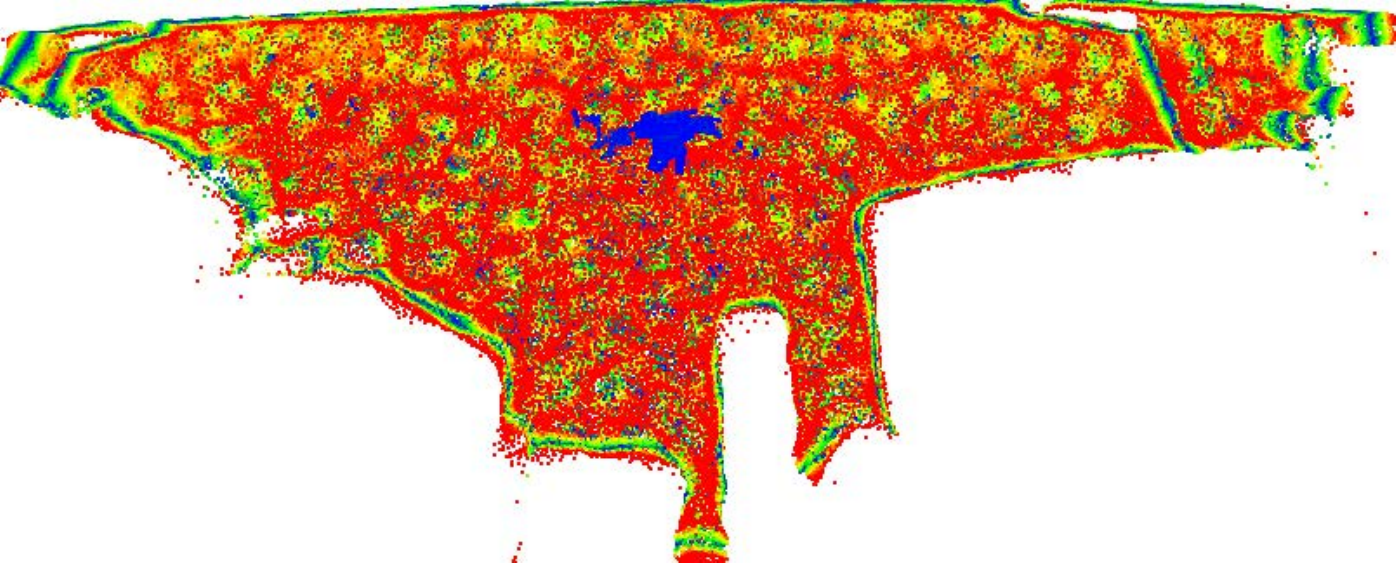}&\includegraphics[width=0.92in]{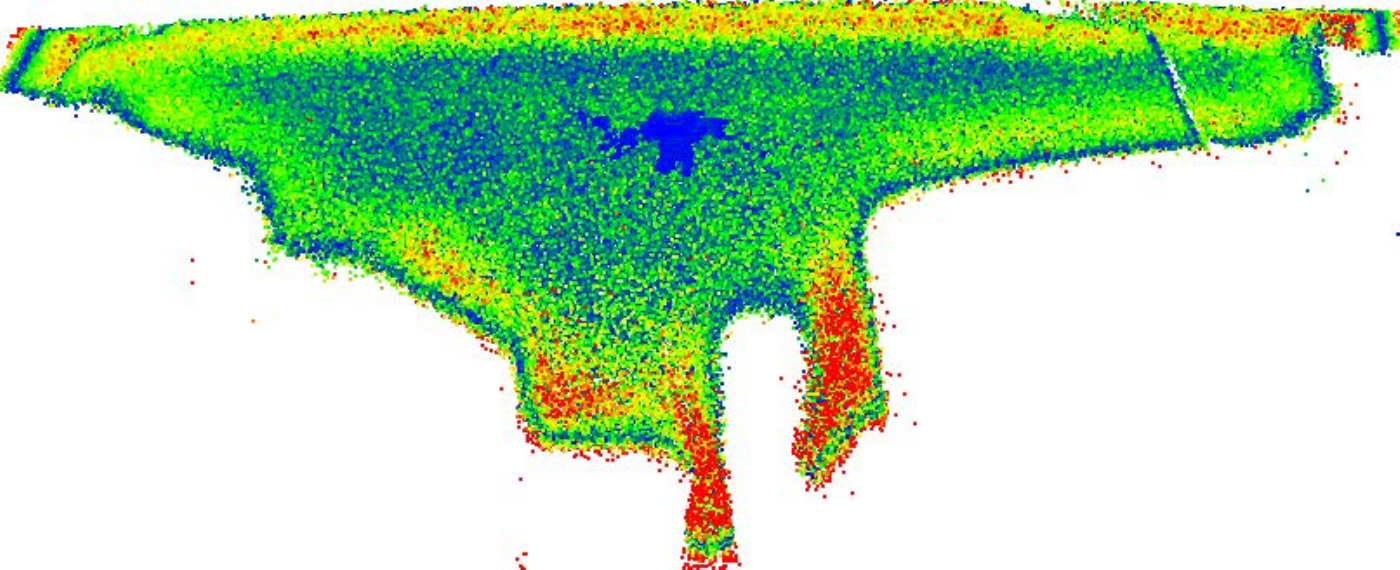}&\includegraphics[width=0.92in]{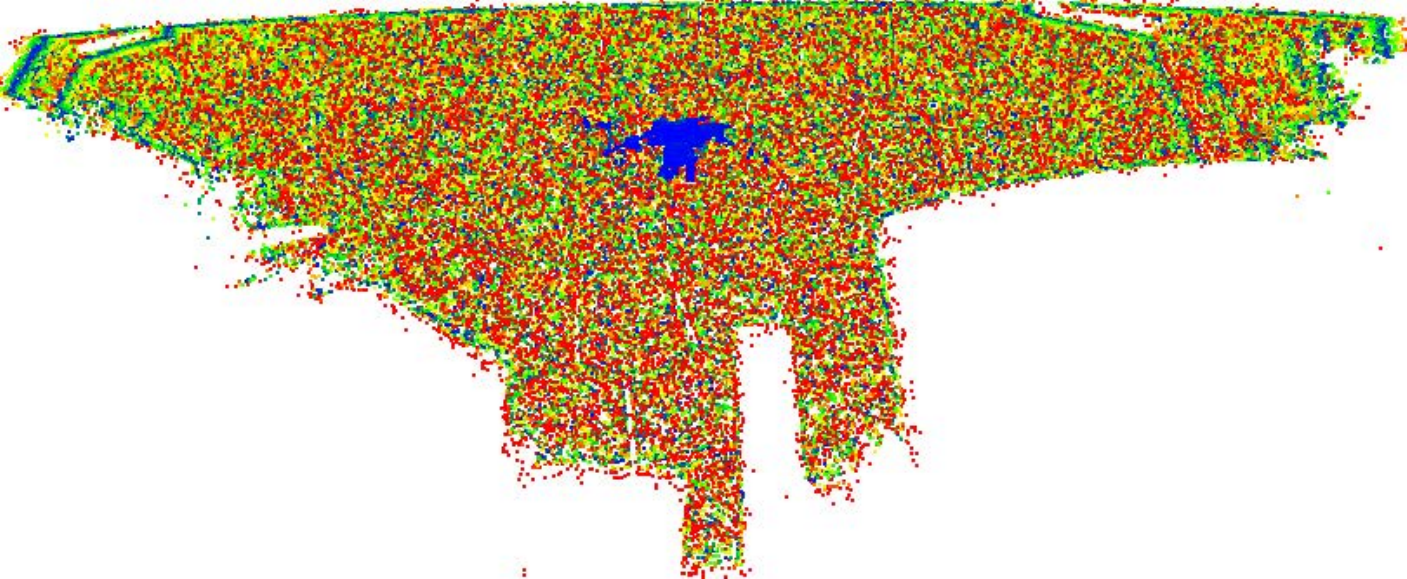}&\includegraphics[width=0.92in]{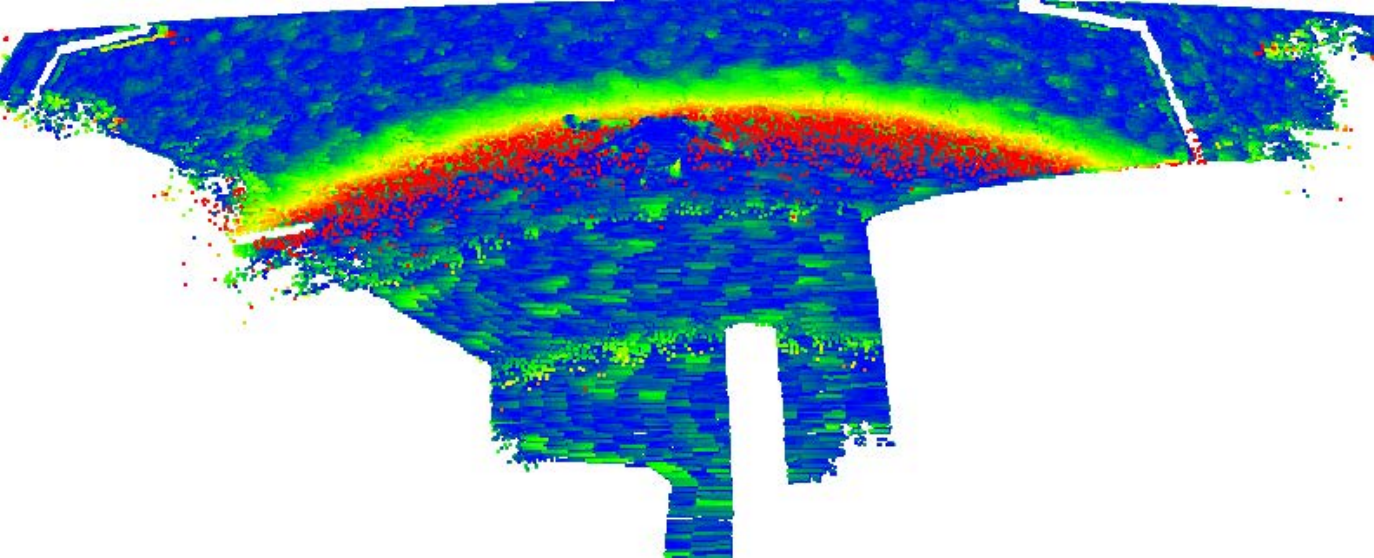} & \\ 
 & \centering\small BF\cite{tomasi98} & \centering\small MRPCA\cite{mattei2017point} & \centering\small GLR\cite{zeng20193d} & \centering\small PCN\cite{rakotosaona2020pointcleannet} & \centering\small DMR\cite{luo2020differentiable}  & \centering\small Proposed & 
\end{tabular}
\caption{\footnotesize Visual comparisons using (a) \texttt{Recycle} and (b) \texttt{ArtL} in the Middlebury dataset \cite{scharstein2014high} and (c) \texttt{Driving} in the synthetic dataset \cite{mayer2016large} in terms of C2C errors, where the depth views were corrupted by SDGN. From blue to red, C2C absolute errors gradually become larger. More blue points are noticeably included in the proposed method. }
\label{fig:visual}
\end{center}
\end{figure*}

Visual results of the denoised PCs \texttt{Recycle} and \texttt{ArtL} from the Middlebury dataset, and a PC \texttt{Driving} from the synthetic dataset for the additive SDGN case are shown in Fig.\,\ref{fig:visual}, where we colored the point clouds according to the C2C absolute distances between the ground truth points and their closest denoised points.
For the result of \texttt{Driving}, we only show its background for better visualization.
From Fig.\,\ref{fig:visual}, it is obvious that our proposal achieved smaller C2C errors compared to the competitors.

Overall, one can observe that the six PC denoising algorithms were worse than the image denoising schemes.
The reason is that these PC denoising algorithms assume that i.i.d. Gaussian noise are added to the point coordinates in 3D space. 
This assumption is inaccurate in practice, because a corrupted depth pixel causes errors only in the depth dimension perpendicular to the image plane, which is quite different from i.i.d. noise in 3D space.
Thus, enhancing depth measurements before synthesizing a PC is more sensible.
In contrast, the three selected image denoising schemes performed poorly under signal-dependent noise and non-uniform quantization. 
Further, it was difficult for BF \cite{tomasi98} and BM3D \cite{dabov2007image} to handle existing holes throughout the images, leading to locally poor performance.

In contrast, our proposal enhances depth measurements before projecting to 3D space to synthesize a PC.
It allows us to tailor the optimization specifically for our depth formation model which complies with the physical acquisition process.
Particularly, we model the combination of signal-dependent noise addition and non-uniform log-based quantization. 
Further, by employing feature graph learning, we can flexibly enhance only the available pixels around missing ones in a depth image.
For these reasons, our scheme achieved noticeable performance gains.

\subsubsection{Comparison for In-house Dataset}

All the above experiments were conducted using artificial noise. 
To see the effectiveness of our method on real sensor noise, we computed Pseudo MOS scores of no-reference metric \cite{liu2021LSPCQA} for different methods using PCs from our in-house dataset. 
When MOS score equals $5$, it means that no discernible distortion is perceived in the PC. 
MOS score equaling $3$ implies that distortion slightly obstructs viewing. 
Note that this no-reference metric considers both color and geometry attributes, while our optimization restores only geometric information, and thus generally lowering the scores. 
However, we observe that our method still outperformed three model-based image denoising schemes and six PC denoising algorithms.

\begin{table} 
	\centering
	\caption{\footnotesize Pseudo MOS results of competing methods for the PCs of the in-house real dataset.} \label{tab:real}
	\scriptsize
	\begin{tabular}{c||c|c|c|c}
	    \hline
		Pseudo MOS & Scene0 & Scene1 & Scene2 & Average\\
		 \hline
	    \hline
	     BF\cite{tomasi98} & 2.96 & 2.73 & 2.55 & 2.75   \\
	    \hline
	   BM3D\cite{dabov2007image} & 2.72 & 2.31 & 2.15 & 2.39 \\
	    \hline
	   SINUQ\cite{zhang20203d} & 2.88 & 2.47  & 2.47 & 2.61 \\
	    \hline
	    \hline
	APSS\cite{guennebaud2007algebraic} & 2.94 &  2.88 & 3.06 &  2.96 \\
	     \hline
	    RIMLS\cite{oztireli2009feature} & 2.99 & 2.80  & 2.94 & 2.91 \\
	    \hline
	     MRPCA\cite{mattei2017point} & 3.11 & 2.82 & 2.84 & 2.92  \\
	     \hline
	     GLR\cite{zeng20193d} & 2.84 & 2.65 & 2.97 & 2.82   \\
	      \hline 
	     PCN\cite{rakotosaona2020pointcleannet} & 3.10 & 3.02 & 3.16 & 3.09\\
	     \hline 
	     DMR\cite{luo2020differentiable} & 3.05 & 2.59 & 2.84 & 2.83 \\
	     \hline
	    \hline
   	  Proposed & \textbf{3.13}  & \textbf{3.15} & \textbf{3.35} & \textbf{3.21} \\
        \hline
    	\end{tabular}
    	\vspace{-0.4cm}
\end{table}

\begin{figure*}[htbp]
\begin{center}
\begin{tabular}{m{0.1cm}ccccc}
\small (a)&\includegraphics[width=1.35in]{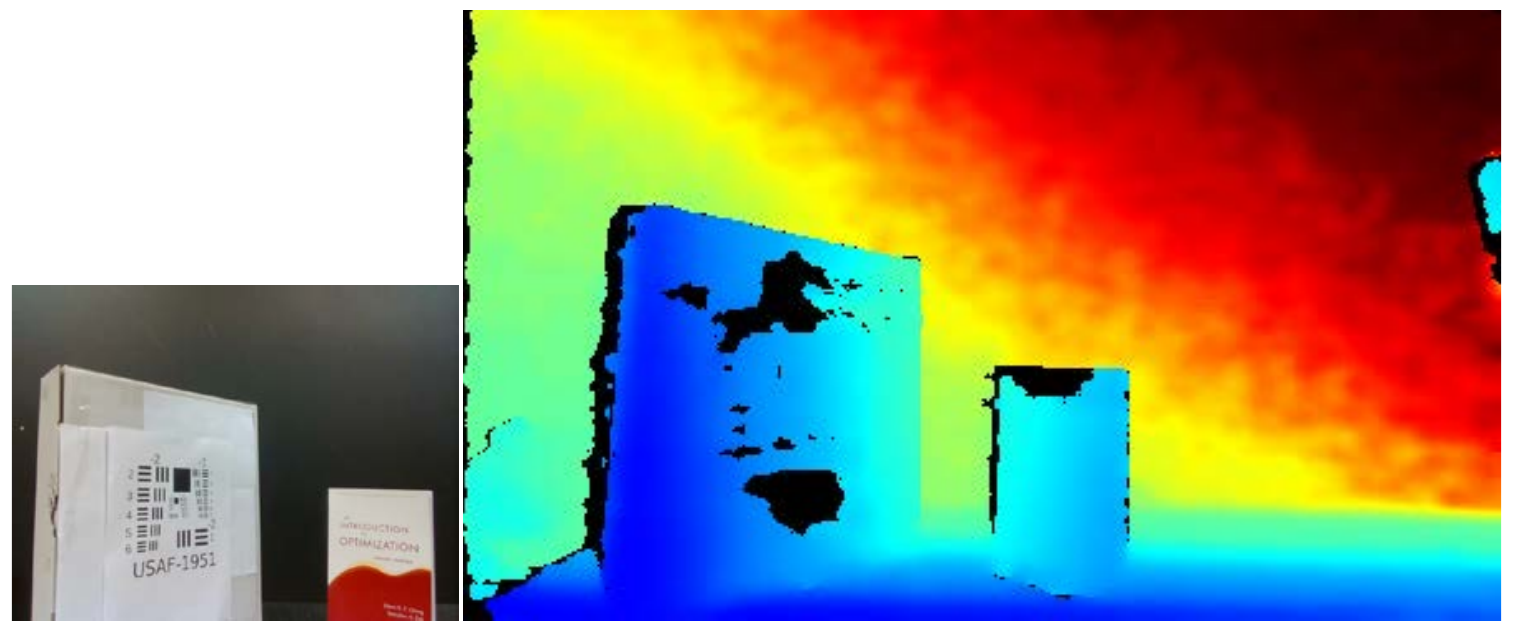}&\includegraphics[width=1.15in]{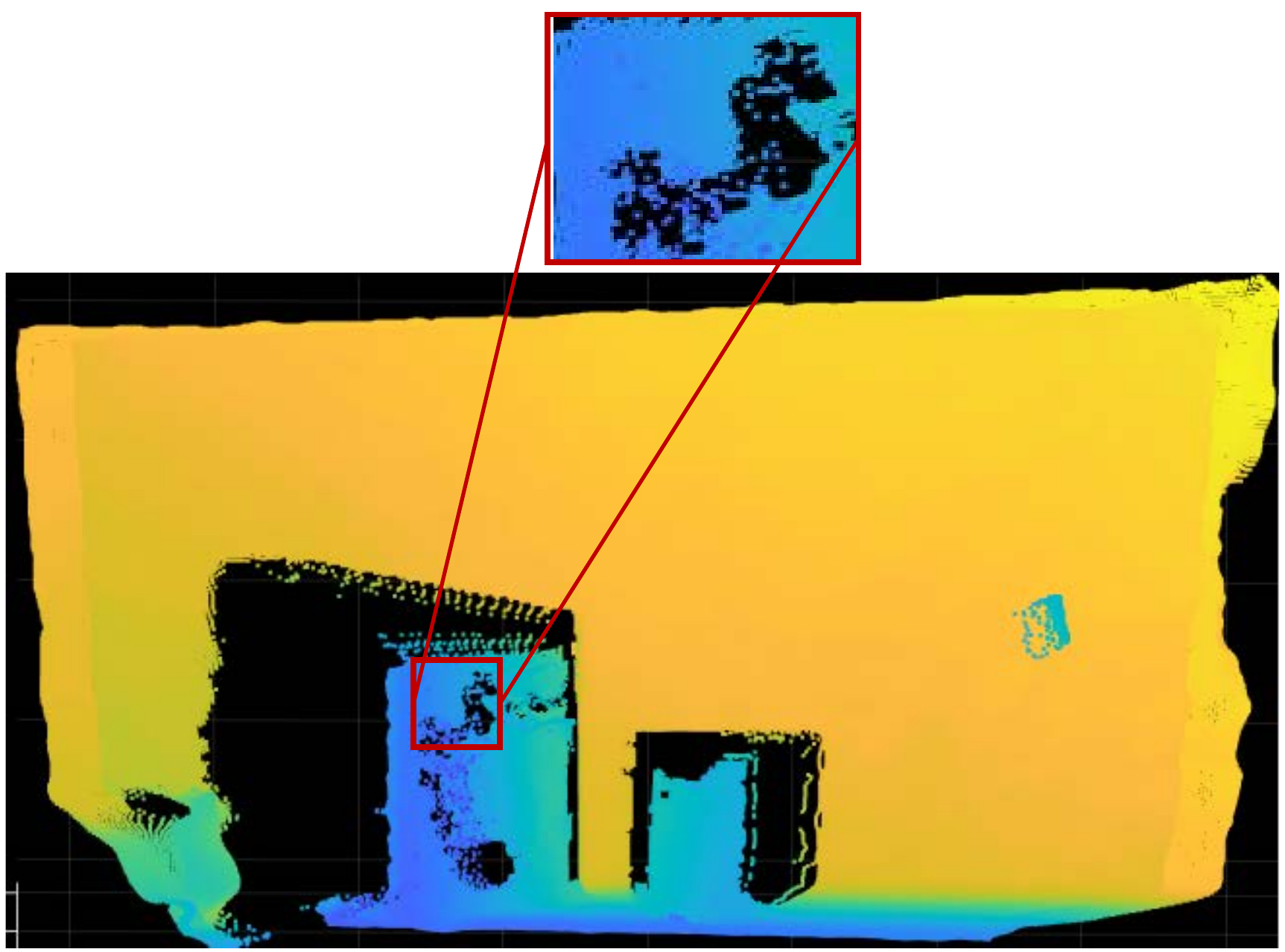}&\includegraphics[width=1.15in]{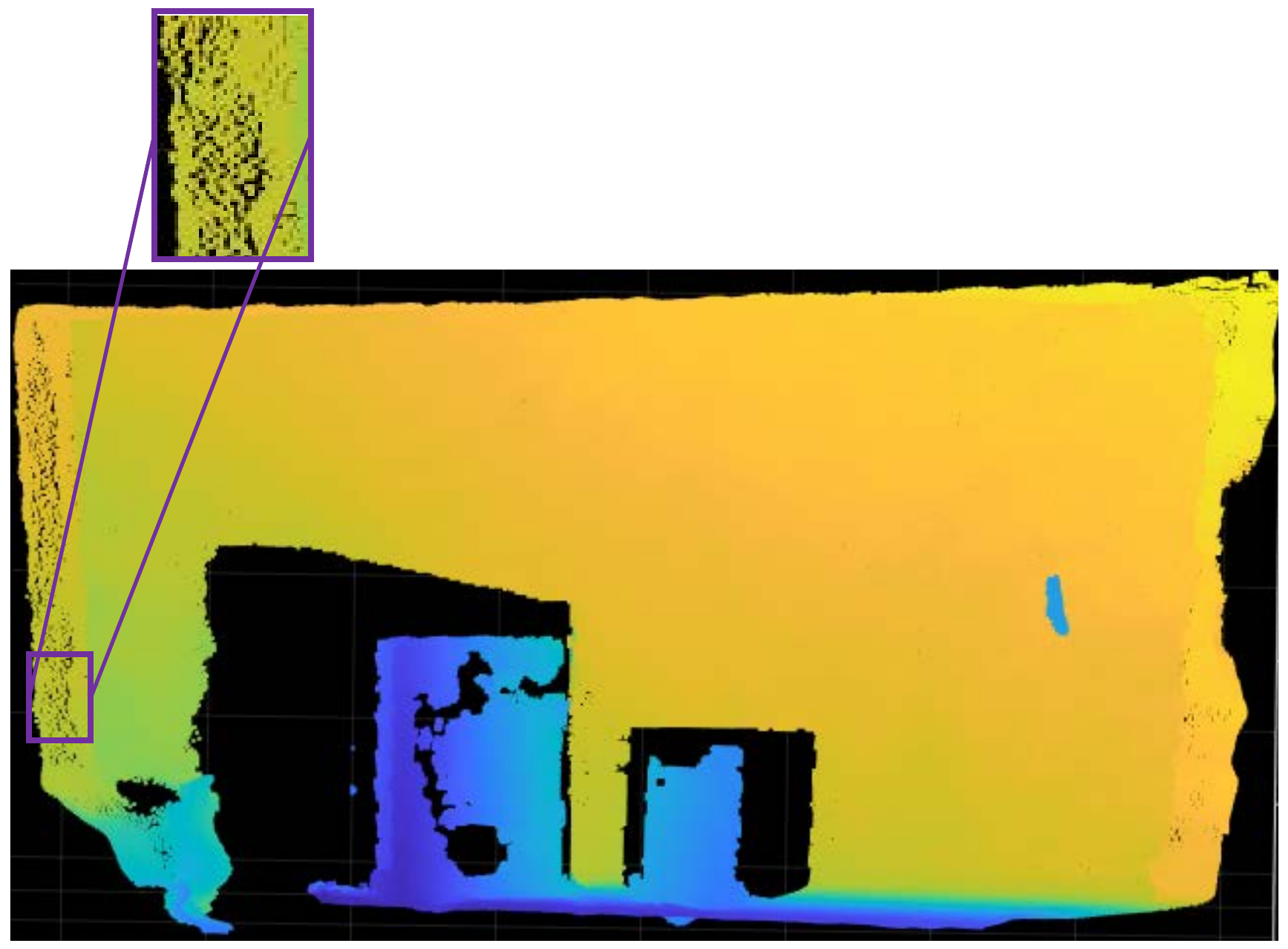}&\includegraphics[width=1.15in]{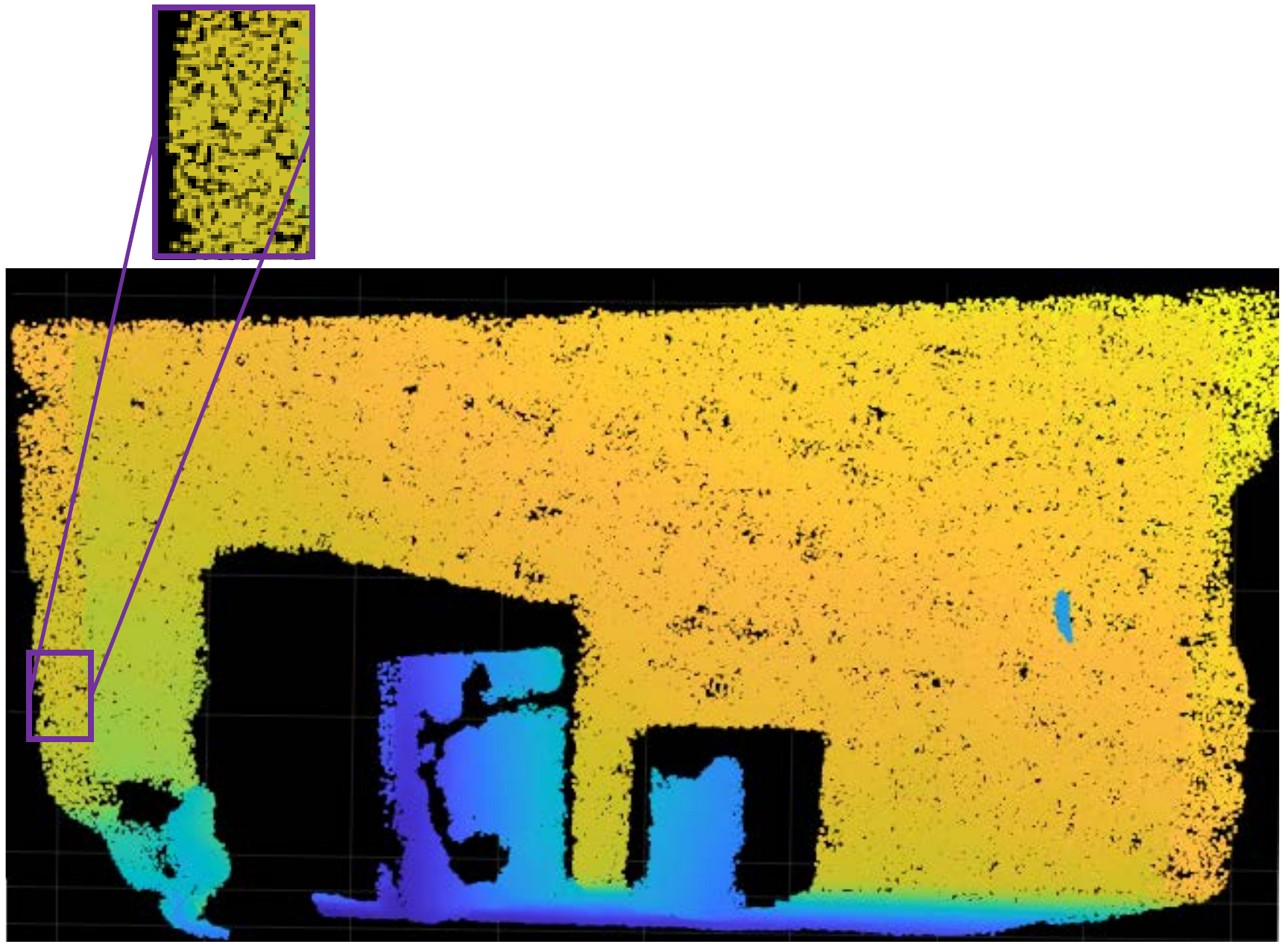}&\includegraphics[width=1.15in]{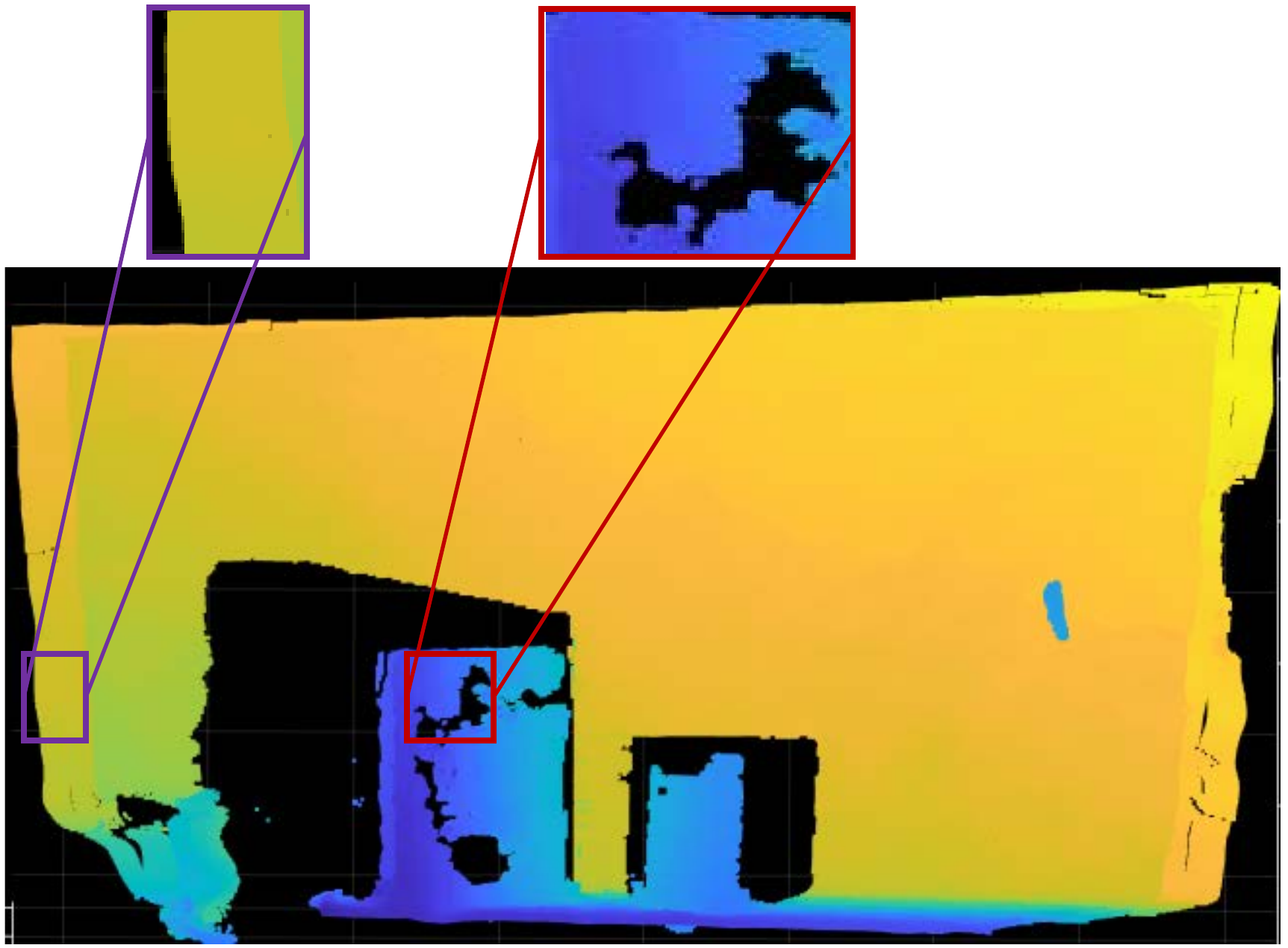} 
\\
\small (b)&\includegraphics[width=1.35in]{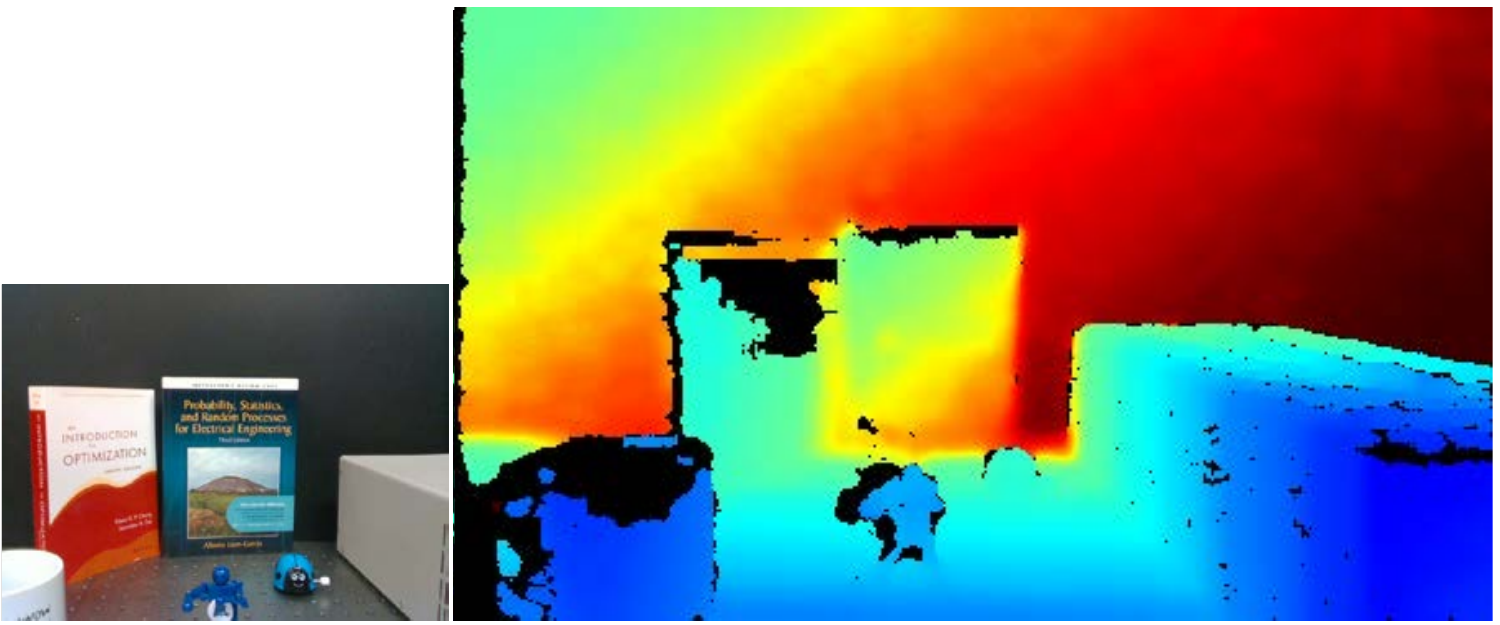}&\includegraphics[width=1.15in]{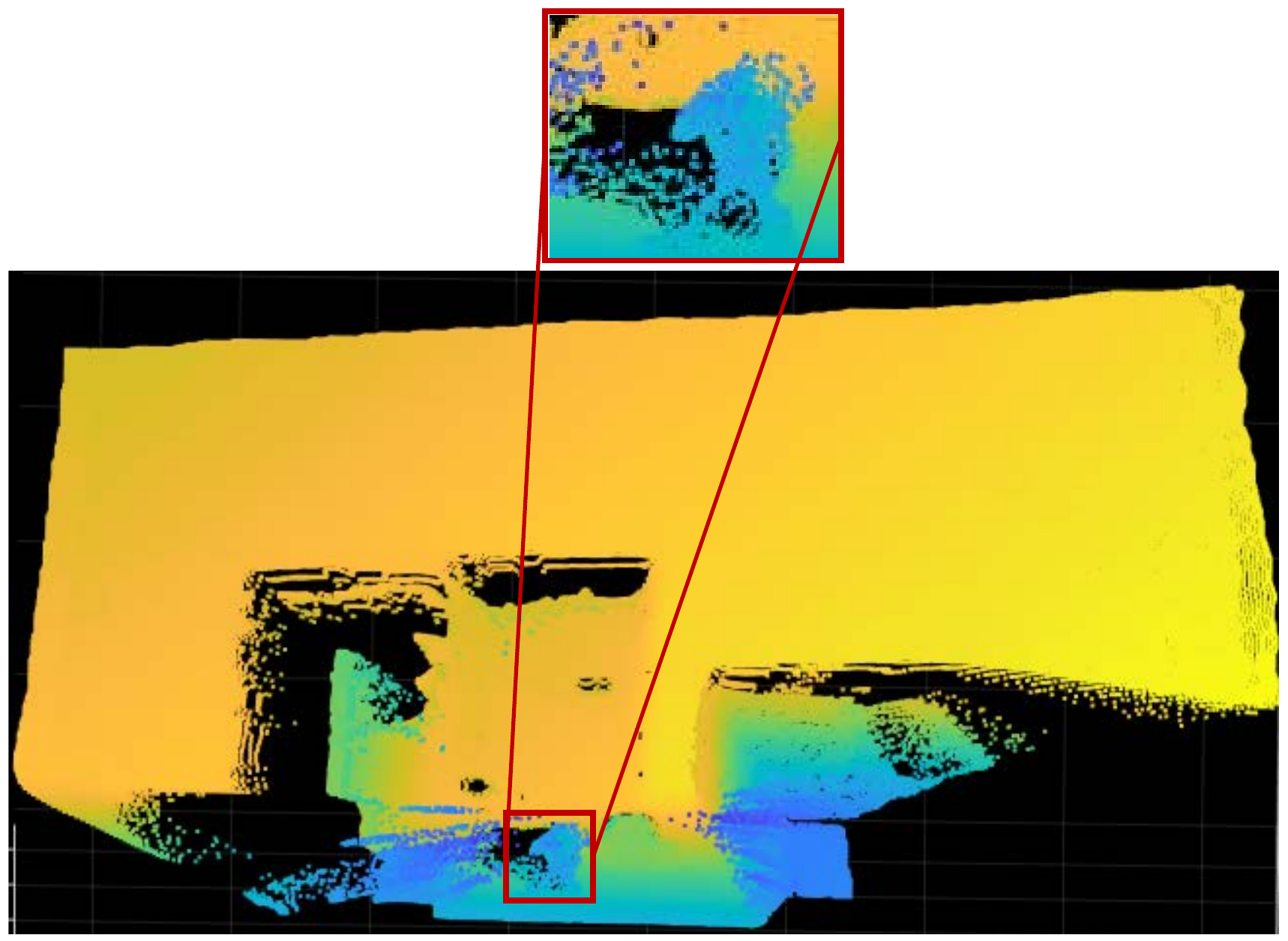}&\includegraphics[width=1.15in]{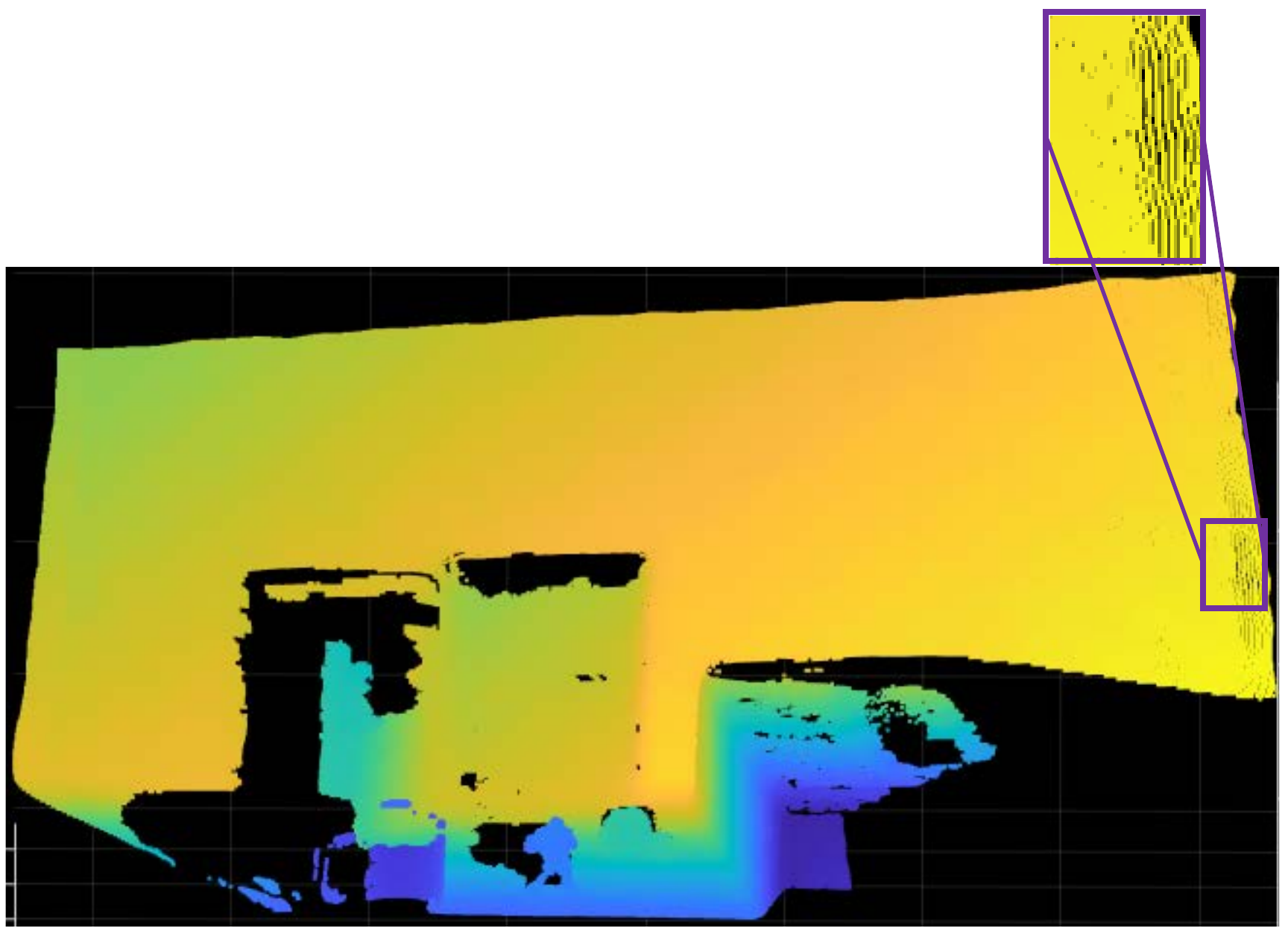}&\includegraphics[width=1.15in]{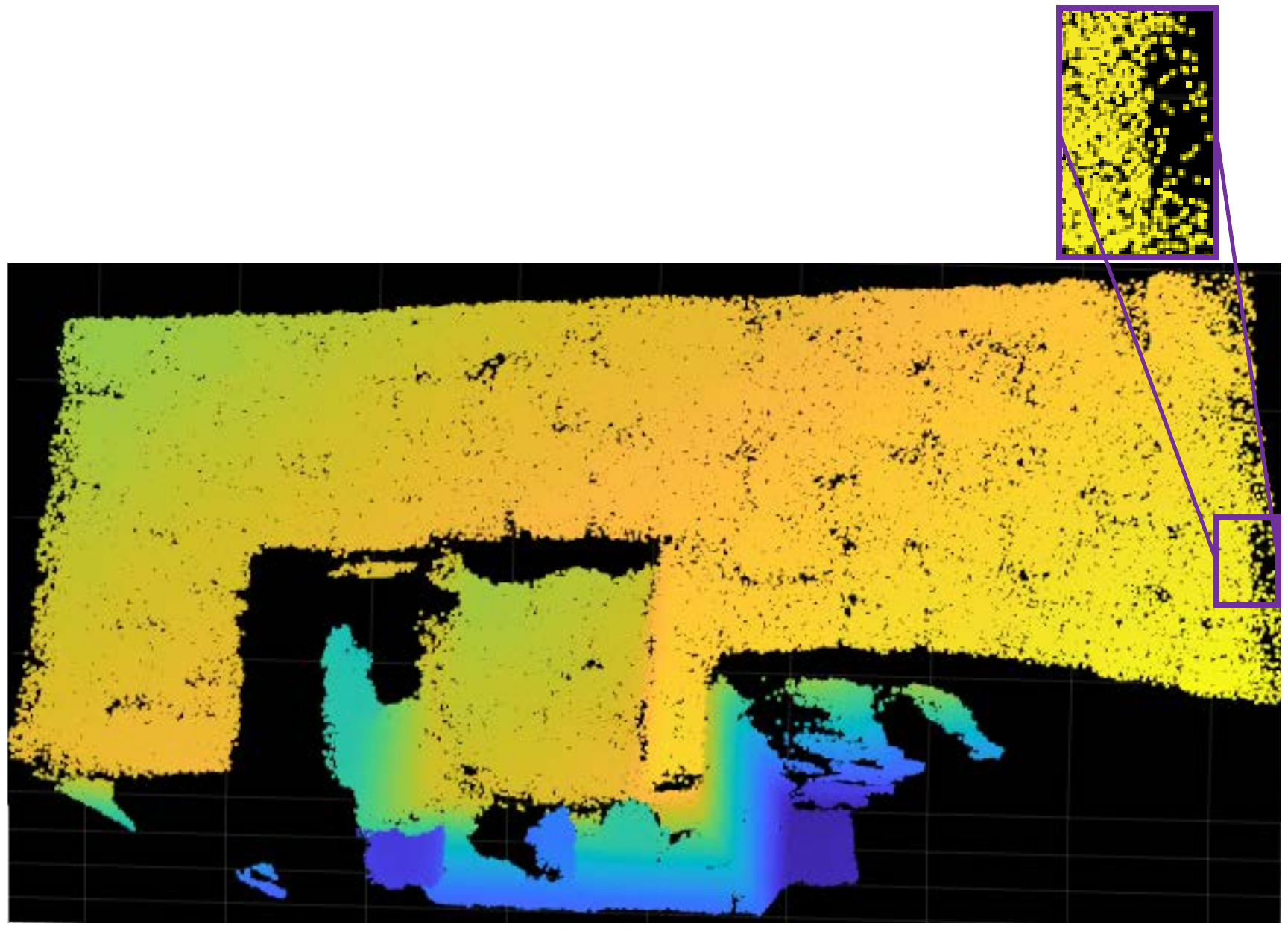}&\includegraphics[width=1.15in]{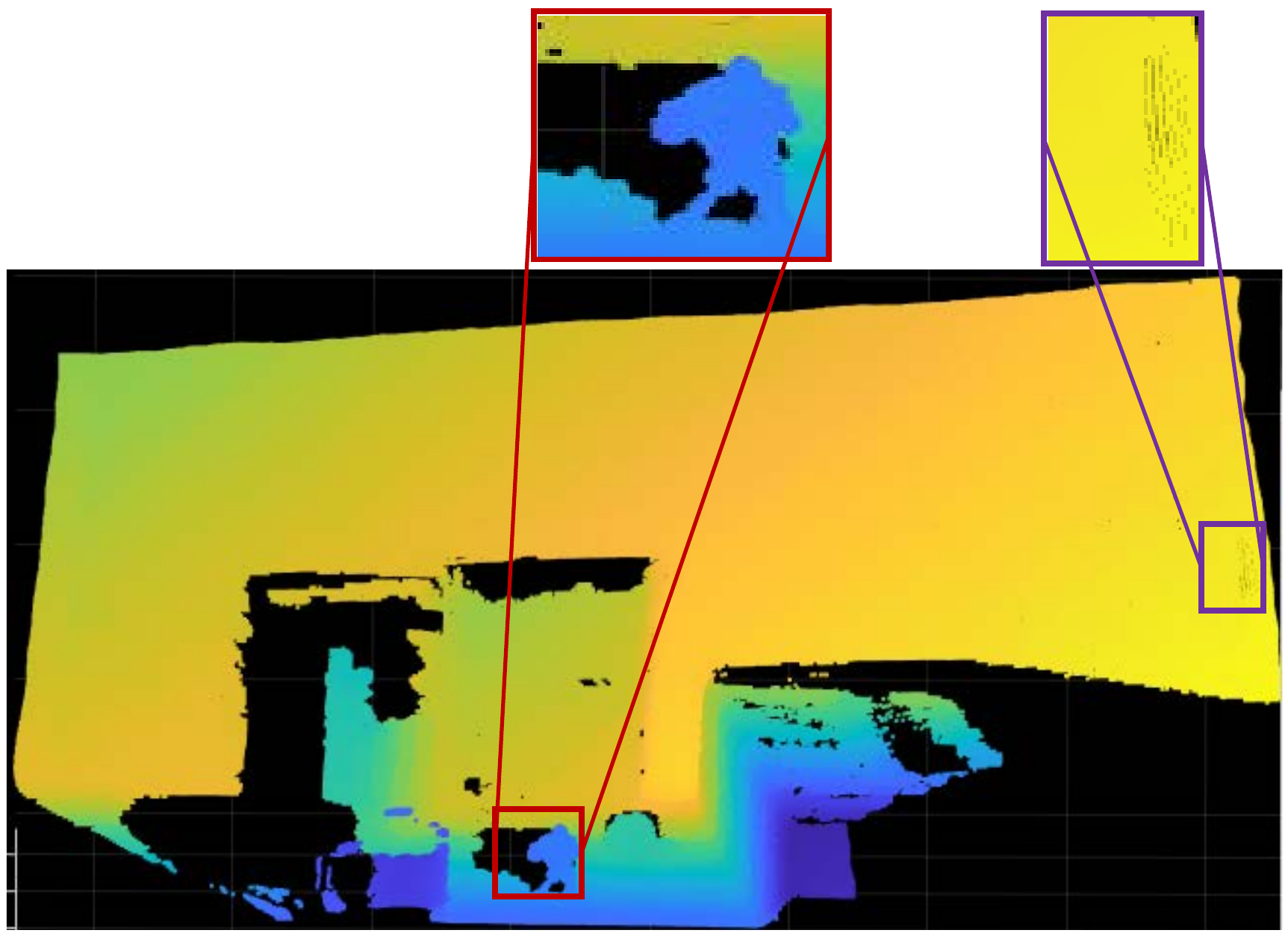} 
\\
\small (c)&\includegraphics[width=1.35in]{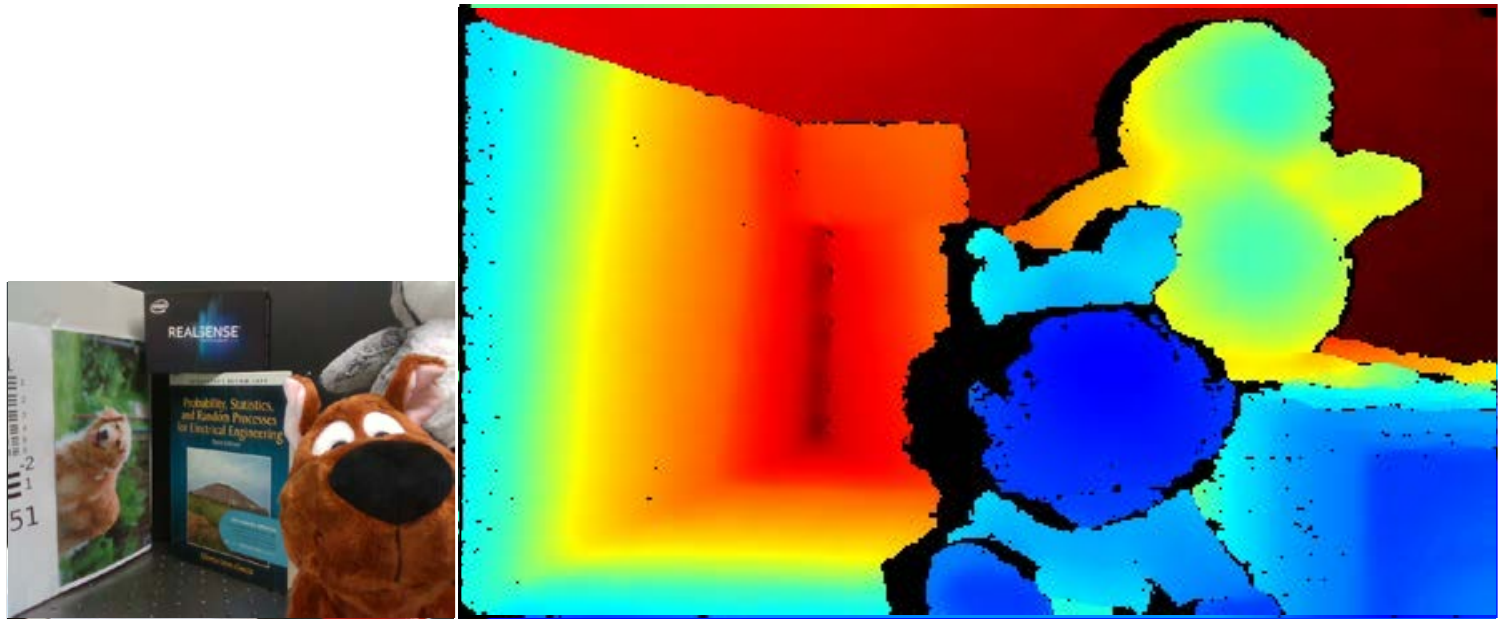}&\includegraphics[width=1.15in]{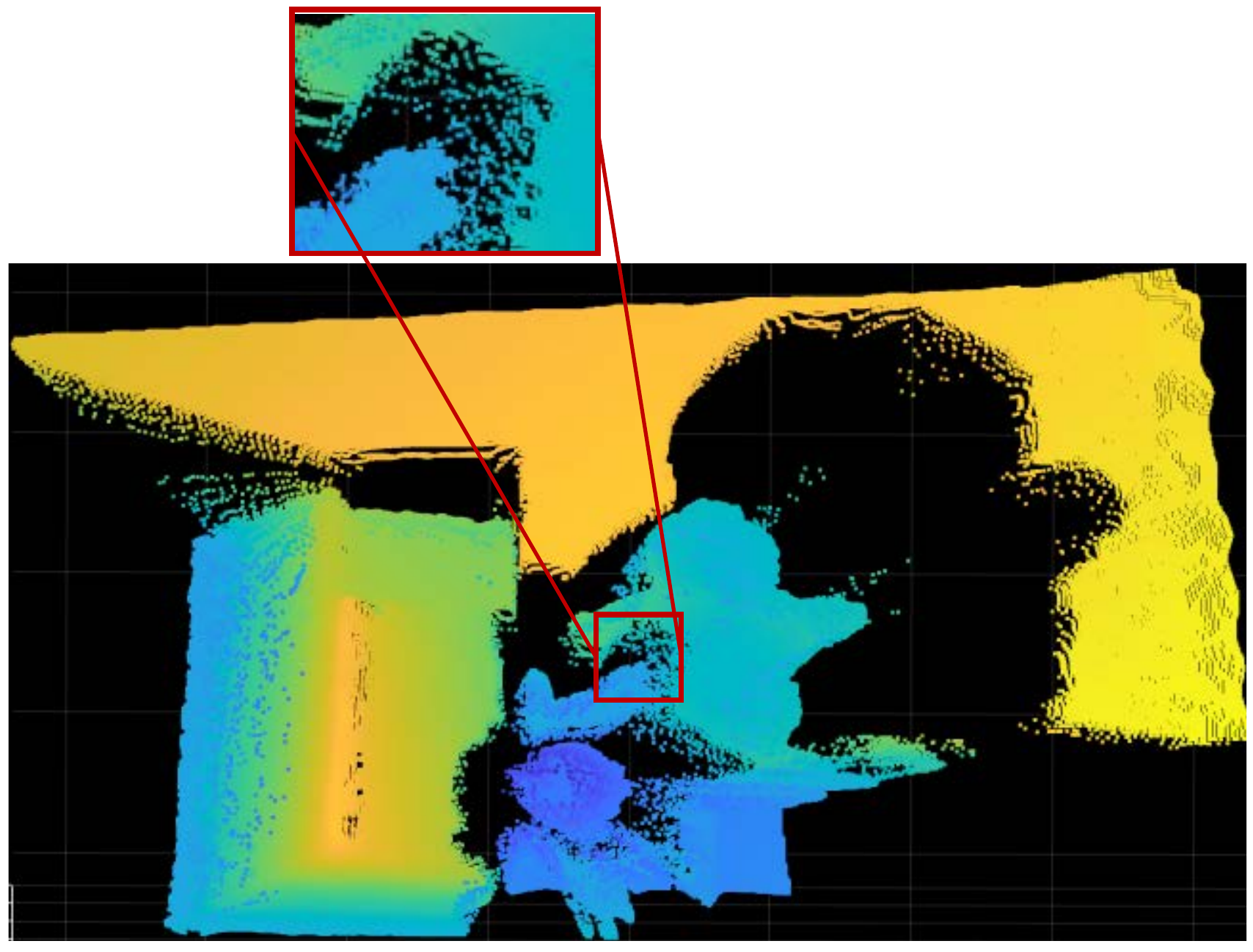}&\includegraphics[width=1.15in]{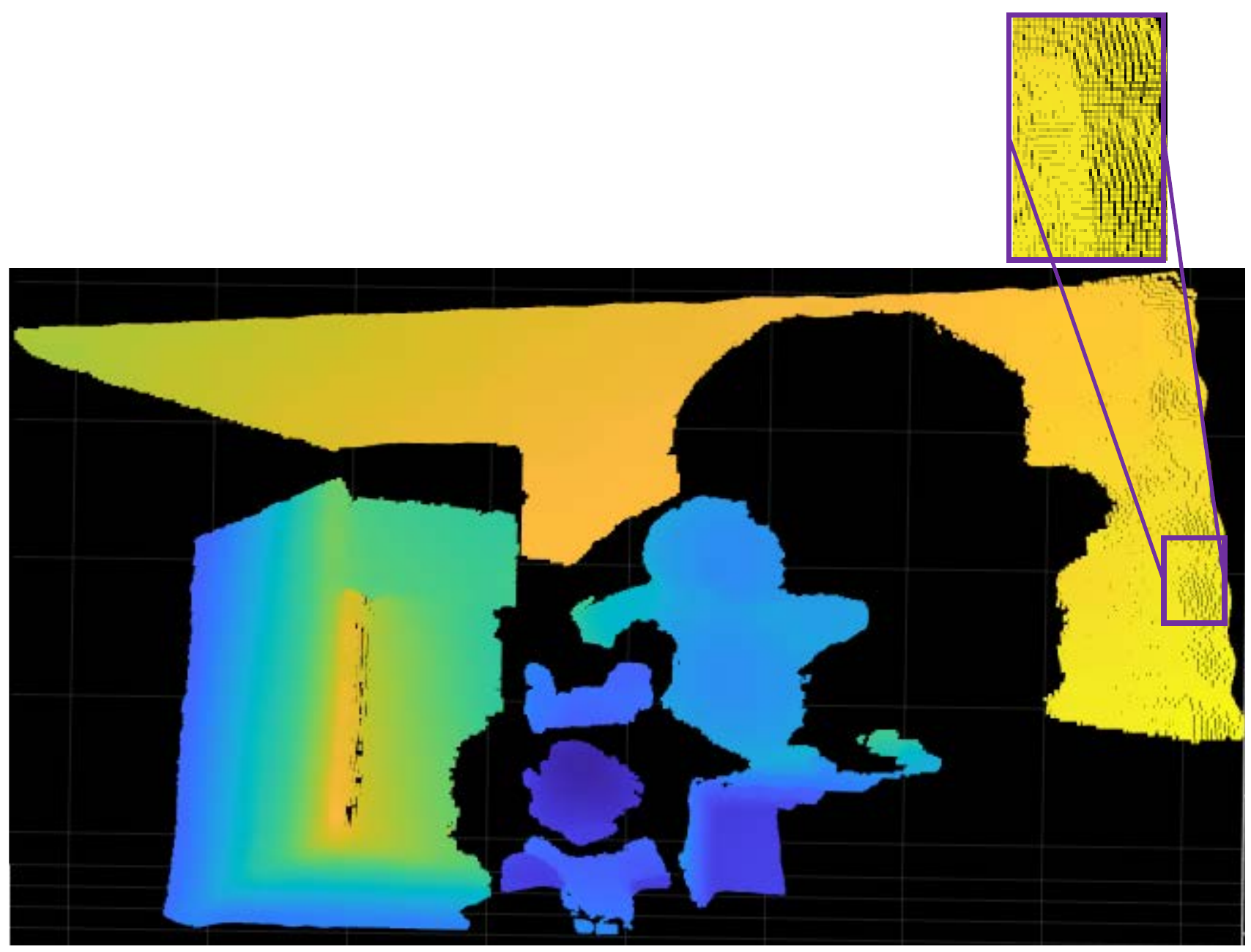}&\includegraphics[width=1.15in]{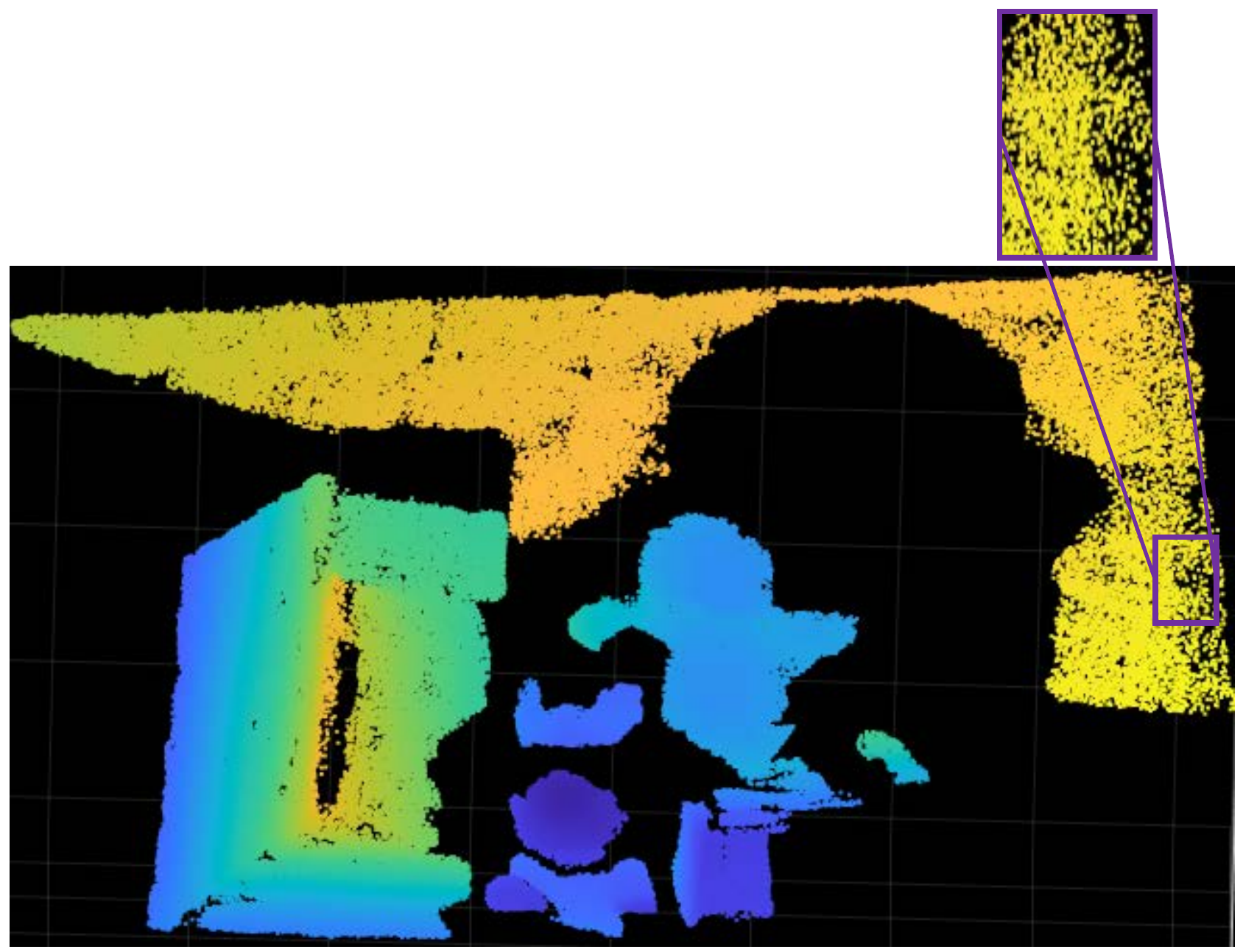}&\includegraphics[width=1.15in]{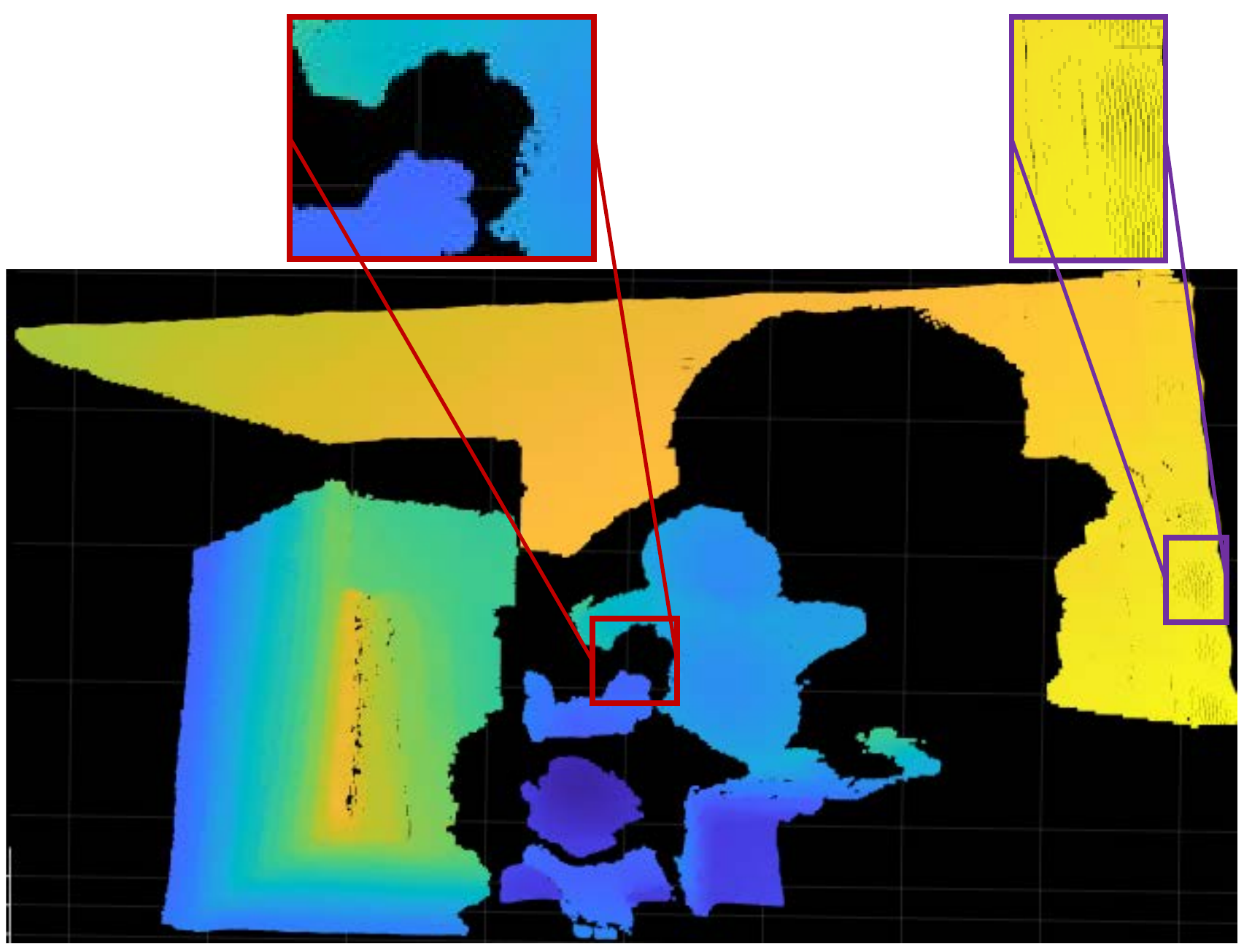} 
\\
 ~ & \centering\small RGB and depth images & \centering\small BF\cite{tomasi98} & \centering\small PCN\cite{rakotosaona2020pointcleannet} & \centering\small DMR\cite{luo2020differentiable}  & \centering\small Proposed
\end{tabular}
\caption{\footnotesize Visual comparisons for the denoised PCs (a) \texttt{Scene0}, (b) \texttt{Scene1} and (c) \texttt{Scene2}, projected from our in-house raw depth image pairs collected using \textit{Intel RealSense\textsuperscript{TM} D435}. Smoother surface silhouette implies a higher denoising quality. 
Note that the collected RGB images and depth images are for reference without alignment, and thus are different view images captured by slightly offset cameras. The resulting PCs are shown from a particular chosen viewpoint.}
\label{fig:visual2}
\end{center}
\vspace{-0.5cm}
\end{figure*}

We next show visual comparisons for our in-house collected PCs as perceived from the right depth image in Fig.\,\ref{fig:visual2}.
For the denoised PCs, a smoother surface silhouette basically implies a higher restoration quality.
Note that BF reconstructed the intensity of each pixel with a weighted average of intensity values from nearby pixels. 
Consequently, given an image with many missing pixels, this method diffused the available pixels to the missing pixels and propagated errors from missing pixels to the target pixels, as shown in the enlarged regions in the second column.
Due to the i.i.d. Gaussian noise assumption on PCs, distortion in 3D points stemming from the real formation process were challenging for the two deep learning based methods, PCN and DMR, leading to poor restoration results.
In contrast, our proposal provided restoration with noticeably better visual quality, which has smoother surface and fewer noisy points.

\section{Conclusion and Discussion}
\label{sec:conclude}
Point clouds are typically synthesized from finite-precision depth measurements that are noise-corrupted.
In this paper, we improve the quality of a synthesized point cloud by jointly enhancing multiview depth images---the ``rawest" signal we can acquire from an off-the-shelf sensor---prior to steps in a typical point cloud synthesis pipeline that obscure acquisition noise. 
We formulate a graph-based MAP optimization that specifically targets an image formation model accounting for both signal-dependent noise addition and non-uniform log-based quantization. We validate the designed model using collected empirical data from an actual depth sensor.
We optimize the objective efficiently using AGD with GCT-aided optimal step size determination. 
Simulation results show that our proposed scheme outperforms competing schemes that denoise point clouds after the synthesis pipeline and representative image denoising schemes.

\appendices
\section{Proof of Multiple Integral}
We first prove \eqref{eq:multiInt} by induction. 
\begin{small}
\begin{align}
&\rPr(\y_l | \x_l)  \approx 
\prod_{i=1}^N \int_{\cR_{l,i}} 
\left( a_{l,i} n_{l,i} + b_{l,i} \right) d\, n_{l,i} \nonumber \\
&=  \prod_{i=1}^N \int_{ z^-_{l,i} - x_{l,i}}^{ z^+_{l,i} - x_{l,i}} 
\left( a_{l,i} n_{l,i} + b_{l,i} \right) d\, n_{l,i} \nonumber \\
&= \prod_{i=1}^N \left. \frac{a_{l,i}}{2} n_{l,i}^2 + b_{l,i} n_{l,i} \right|_{z^-_{l,i} - x_{l,i}}^{z^+_{l,i} - x_{l,i}} 
\nonumber \\
&=\prod_{i=1}^N  a_{l,i}(z^-_{l,i}-z^+_{l,i})x_{l,i}+\frac{a_{l,i}}{2}((z^+_{l,i})^2 - (z^-_{l,i})^2) +  b_{l,i}(z^+_{l,i}-z^-_{l,i}).
\label{eq:prove1} 
\end{align}
\end{small}\noindent
Denote by $\1_i$ an one-hot vector, which is a suitably long \textit{canonical vector} of all zeros except entry $i$ is $1$, $\tilde{\a}_{l,i}^{\top} = a_{l,i} (z^-_{l,i} - z^+_{l,i}) \1_i^{\top}$ and $\tilde{b}_{l,i} = \frac{a_{l,i}}{2} ((z^+_{l,i})^2 - (z^-_{l,i})^2) + b_{l,i} (z^+_{l,i} - z^-_{l,i})$. We now rewrite \eqref{eq:prove1} as
\begin{small}
\begin{align}
\rPr(\y_l | \x_l) \approx & \prod_{i=1}^N 
(\tilde{\a}_{l,i}^{\top} \x_{l} + \tilde{b}_{l,i}).
\label{eq:prove2} 
\end{align}
\end{small}\noindent

Given \eqref{eq:prove1}, \eqref{eq:prove2} and \eqref{eq:approxG}, we now prove \eqref{eq:multiInt2} for noise $\n_r$.
\begin{small}
\begin{align}
\rPr(\y_r | \x_l) \approx & 
\prod_{j=1}^M \int_{\cR_{r,j}} 
\left( a_{r,j} n_{r,j} + b_{r,j} \right) d\, n_{r,j} \nonumber \\
= & \prod_{j=1}^M 
(\tilde{\a}_{r,j}^{\top} \x_{r} + \tilde{b}_{r,j}) 
\approx \prod_{j=1}^M 
(\tilde{\a}_{r,j}^{\top} (\H \x_{l}+\e) + \tilde{b}_{r,j}).
\label{eq:prove3} 
\end{align}
\end{small}
Denote by $\bar{\a}_{r,j}^{\top}=\tilde{\a}_{r,j}^{\top} = a_{r,j} (z^-_{r,j} - z^+_{r,j}) \1_i^{\top}$ and $\bar{b}_{r,j} = \tilde{\a}_{r,j}^{\top}\e+\tilde{b}_{r,j} =\bar{\a}_{r,j}^{\top}\e+\frac{a_{r,j}}{2} ((z^+_{r,j})^2 - (z^-_{r,j})^2) + b_{r,j} (z^+_{r,j} - z^-_{r,j})$. We now rewrite \eqref{eq:prove3} as
\begin{small}
\begin{align}
\rPr(\y_r | \x_l) 
\approx& \prod_{j=1}^M 
(\bar{\a}_{r,j}^{\top} \H \x_{l} +  \bar{b}_{r,j}).
\end{align}
\end{small}

$\qed$

\bibliographystyle{IEEEtran}
\bibliography{ref2}

\vspace{-0.7cm}
\begin{IEEEbiography}[{\includegraphics[width=1in,height=1.25in,clip,keepaspectratio]{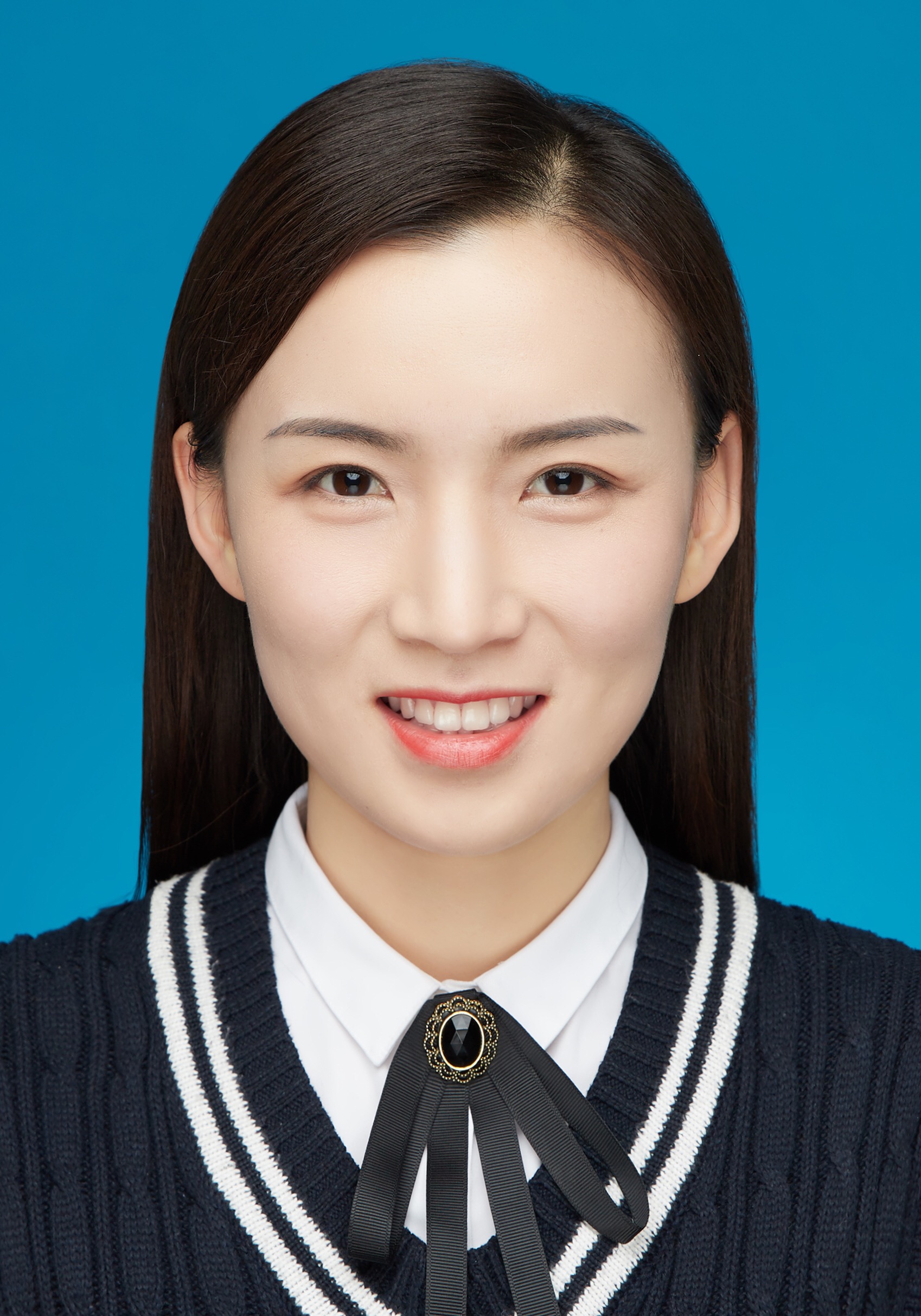}}]{Xue Zhang} (Member, IEEE) received the Ph.D. degree in signal and information processing from Beijing Jiaotong Unversity (BJTU), Beijing, China, in 2019. From 2015 to 2017, she was a Visiting Ph.D. student with the Signal Processing Laboratory (LTS4), Swiss Federal Institute of Technology (EPFL), Lausanne, Switzerland. In 2018, she was a Visiting Ph.D. student with National Institute of Informatics (NII), Tokyo, Japan. She was a Post-Doctoral Fellow with York University, Toronto, Canada, from 2019 till 2022. She is now an Assistant Professor in Shandong University of Science and Technology (SDUST), Qingdao, China. 
Her research interests include 3D image/video processing, interactive media navigation, and graph signal processing.
\end{IEEEbiography}

\begin{IEEEbiography}[{\includegraphics[width=1in,height=1.25in]{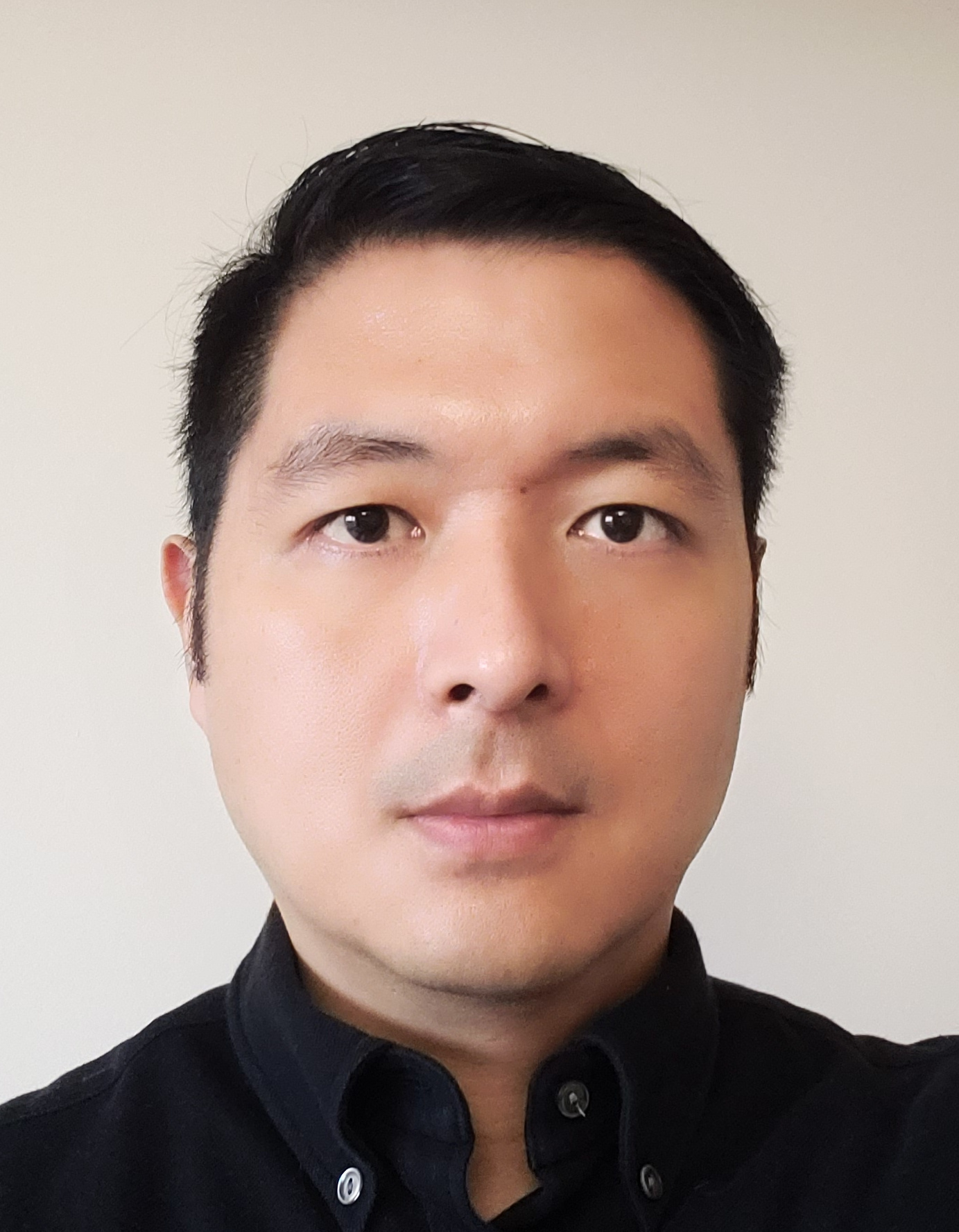}}]{Gene Cheung} (M'00--SM'07--F'21)
received the B.S. degree in electrical engineering from Cornell University in 1995, and the M.S. and Ph.D. degrees in electrical engineering and computer science from the University of California, Berkeley, in 1998 and 2000, respectively. 

He was a senior researcher in Hewlett-Packard Laboratories Japan, Tokyo, from 2000 till 2009. 
He was an assistant then associate professor in National Institute of Informatics (NII) in Tokyo, Japan, from 2009 till 2018. 
He is now a professor in York University, Toronto, Canada.

His research interests include 3D imaging and graph signal processing. 
He has served as associate editor for multiple journals, including IEEE Transactions on Multimedia (2007--2011), IEEE Transactions on Circuits and Systems for Video Technology (2016--2017) and IEEE Transactions on Image Processing (2015--2019).
He currently serves as senior associate editor for IEEE Signal Processing Letters (2021--present).
He served as a member of the Multimedia Signal Processing Technical Committee (MMSP-TC) in IEEE Signal Processing Society (2012--2014), and a member of the Image, Video, and Multidimensional Signal Processing Technical Committee (IVMSP-TC) (2015--2017, 2018--2020). 
He is a co-author of several paper awards and nominations, including the best student paper finalist in ICASSP 2021, best student paper award in ICIP 2013, ICIP 2017 and IVMSP 2016, best paper runner-up award in ICME 2012, and IEEE Signal Processing Society (SPS) Japan best paper award 2016.
He is a recipient of the Canadian NSERC Discovery Accelerator Supplement (DAS) 2019.
He is a fellow of IEEE.
\end{IEEEbiography}

\begin{IEEEbiography}[{\includegraphics[width=1in,height=1.25in]{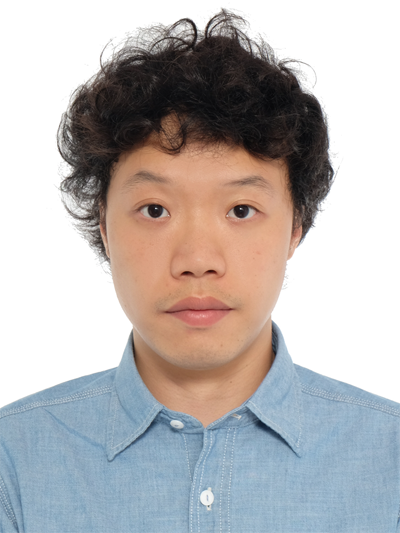}}]{Jiahao Pang}(Member, IEEE) received the B.Eng. degree from South China University of Technology, Guangzhou, China, in 2010, and the M.Sc. and Ph.D. degrees from the Hong Kong University of Science and Technology, Hong Kong, China, in 2011 and 2016, respectively. 
He was a Senior Researcher with SenseTime Group Limited, Hong Kong, China, from 2016 to 2019.
He is currently a Senior Staff Engineer with InterDigital, New York, NY, USA.
His research interests include 3D computer vision, image processing, graph signal processing, and deep learning.
He has over 40 publications in international journals, conferences, and book chapters.
He also serves as a regular reviewer for IEEE TIP, IEEE CVPR, NeurIPS, {\it etc}.
\end{IEEEbiography}

\begin{IEEEbiography}[{\includegraphics[width=1in,height=1.25in,clip,keepaspectratio]{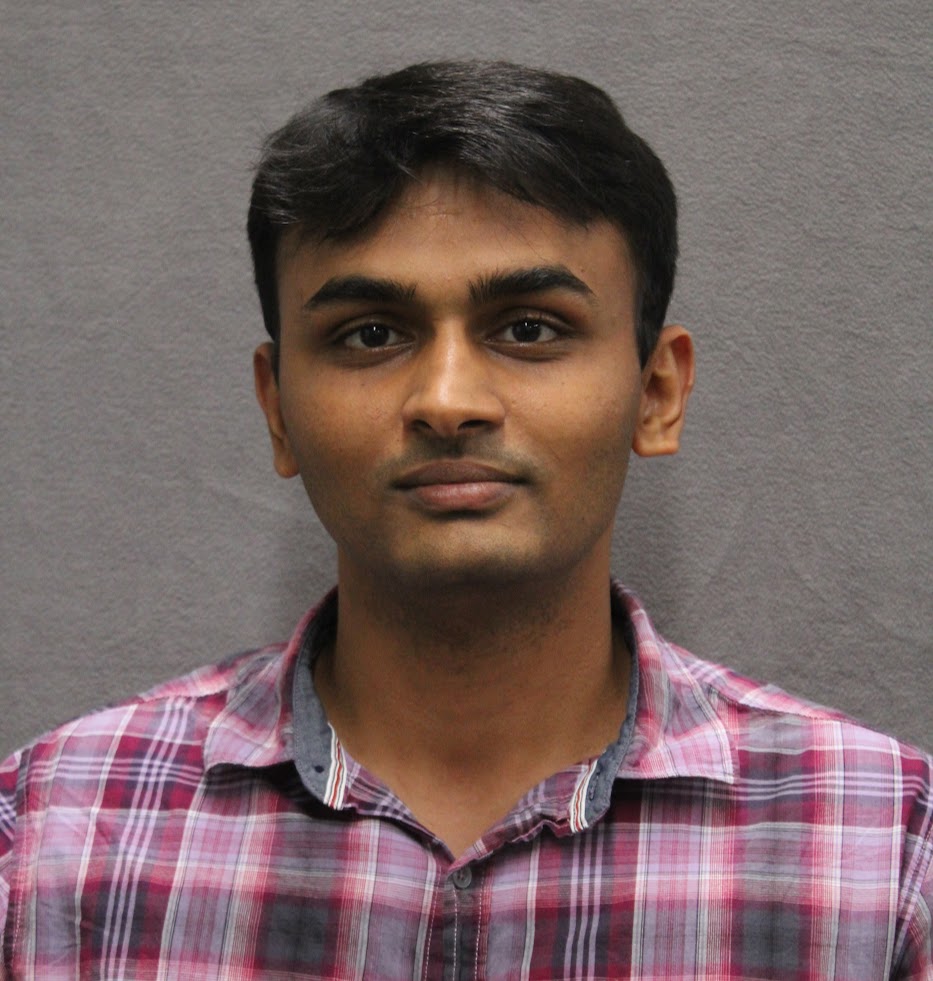}}]{Yash Sanghvi}
(Student Member, IEEE) received the B.Tech. and M.Tech degree in electrical engineering from Indian Institute of Technology, Bombay in 2018. He joined Purdue University in 2019, where he currently is a graduate research assistant in the Intelligent Imaging Lab at Electrical and Computer Engineering, Purdue University, West Lafayette and pursuing his PhD. From 2018 to 2019, he was working on the inverse scattering problem as a project scientist at Indian Institute of Technology Madras. His research interests include inverse problems, computational imaging and deep learning. \end{IEEEbiography}

\begin{IEEEbiography}[{\includegraphics[width=1in,height=1.25in,clip,keepaspectratio]{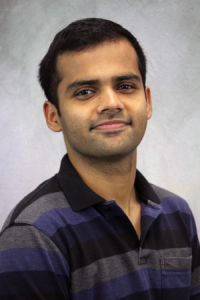}}]
{Abhiram Gnanasambandam} received his B.Tech. Degree in Electrical Engineering from the Indian Institute of Technology Madras in 2017, and his Ph.D. from the School of Electrical and Computer Engineering, Purdue University, in 2022. His Ph.D. dissertation dealt with computational imaging, image processing, deep learning, and computer vision. He is currently a Senior Engineer at Samsung Research America.
\end{IEEEbiography}

\begin{IEEEbiography}[{\includegraphics[width=1in,height=1.25in,clip,keepaspectratio]{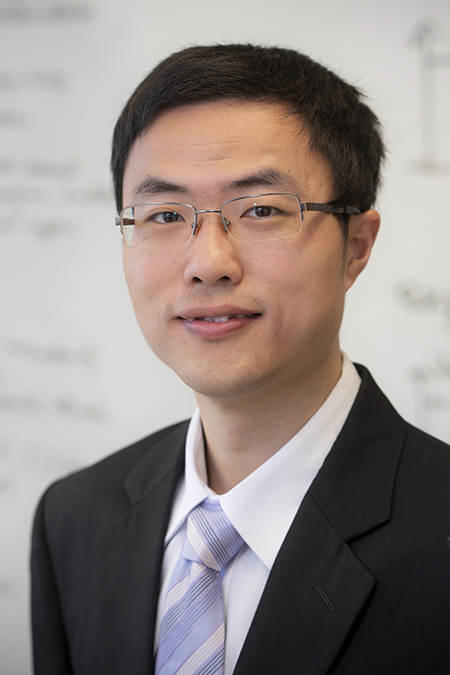}}]
{Stanley H. Chan}
(S'06--M'12--SM'17) received the B.Eng. degree (with first-class honor) in Electrical Engineering from the University of Hong Kong in 2007, the M.A. degree in Mathematics, and the Ph.D. degree in Electrical Engineering from the University of California at San Diego in 2009 and 2011, respectively. From 2012 to 2014, he was a postdoctoral research fellow at Harvard University. He joined Purdue University, West Lafayette, IN in 2014, where he is currently an Elmore Associate Professor of Electrical and Computer Engineering, with a joint appointment in the Department of Statistics.
Dr. Chan is a recipient of the Best Paper Award of IEEE International Conference on Image Processing 2016, IEEE Signal Processing Cup 2016 Second Prize, Purdue College of Engineering Exceptional Early Career Teaching Award 2019, Purdue College of Engineering Outstanding Graduate Mentor Award 2016, and Eta Kappa Nu (Beta Chapter) Outstanding Teaching Award 2015. His research interests include single-photon imaging, imaging through atmospheric turbulence, and computational photography. Dr. Chan is the author of a popular undergraduate textbook \emph{Introduction to Probability for Data Science}, Michigan Publishing 2021. The book is part of the University of Michigan's free textbook initiative to disseminate high-quality educational materials to students and families around the world. Since the launch of the book in Fall 2021, the book has saved more than \$3 million USD for families globally.
He is an Associate Editor of the IEEE Transactions on Computational Imaging since 2018 where he is recognized as an outstanding editorial board member in 2021. He also served as an Associate Editor of the OSA Optics Express in 2016-2018, and an Elected Member of the IEEE Signal Processing Society Technical Committee in Computational Imaging in 2015-2020.
\end{IEEEbiography}

\end{document}